\documentclass[runningheads]{llncs}
\usepackage{graphicx}
\usepackage{amsmath,amssymb} % define this before the line numbering.
\usepackage{color}
\usepackage{times}
\usepackage{epsfig}
\usepackage{graphicx}
\usepackage{amsmath}
\usepackage{amssymb}
\usepackage{rotating}
\usepackage{color}  % for rebuttal
\usepackage{soul} % for rebuttal

\begin{document}
\pagestyle{headings}
\mainmatter

\def\ACCV16SubNumber{119}  % Insert your submission number here

%===========================================================
\title{Semantic-Aware Depth Super-Resolution\\ in Outdoor Scenes} % Replace with your title
\titlerunning{Semantic-Aware Depth Super-Resolution in Outdoor Scenes}
\authorrunning{Miaomiao Liu, Mathieu Salzmann, Xuming He}

\author{Miaomiao Liu${}^1$, Mathieu Salzmann${}^2$, Xuming He${}^1$}
\institute{${}^1$NICTA, ${}^2$EPFL}

\maketitle

\newcommand{\Perp}{\perp\!\!\! \perp}

\newcommand{\comment}[1]{}

\newcommand{\eg}{e.g.}
\newcommand{\ie}{i.e.}
\newcommand{\bK}{\mathbf{K}}
\newcommand{\bX}{\mathbf{X}}
\newcommand{\bY}{\mathbf{Y}}
\newcommand{\bk}{\mathbf{k}}
\newcommand{\bx}{\mathbf{x}}
\newcommand{\bbx}{\bar{\mathbf{x}}}
\newcommand{\by}{\mathbf{y}}
\newcommand{\bhy}{\hat{\mathbf{y}}}
\newcommand{\bty}{\tilde{\mathbf{y}}}
\newcommand{\bG}{\mathbf{G}}
\newcommand{\bI}{\mathbf{I}}
\newcommand{\bg}{\mathbf{g}}
\newcommand{\bS}{\mathbf{S}}
\newcommand{\bs}{\mathbf{s}}
\newcommand{\bM}{\mathbf{M}}
\newcommand{\bw}{\mathbf{w}}
\newcommand{\eye}{\mathbf{I}}
\newcommand{\bU}{\mathbf{U}}
\newcommand{\bV}{\mathbf{V}}
\newcommand{\bW}{\mathbf{W}}
\newcommand{\bn}{\mathbf{n}}
\newcommand{\bv}{\mathbf{v}}
\newcommand{\bq}{\mathbf{q}}
\newcommand{\bR}{\mathbf{R}}
\newcommand{\bi}{\mathbf{i}}
\newcommand{\bj}{\mathbf{j}}
\newcommand{\bp}{\mathbf{p}}
\newcommand{\bt}{\mathbf{t}}
\newcommand{\bJ}{\mathbf{J}}
\newcommand{\bu}{\mathbf{u}}
\newcommand{\bB}{\mathbf{B}}
\newcommand{\bD}{\mathbf{D}}
\newcommand{\bz}{\mathbf{z}}
\newcommand{\bP}{\mathbf{P}}
\newcommand{\bC}{\mathbf{C}}
\newcommand{\bA}{\mathbf{A}}
\newcommand{\bZ}{\mathbf{Z}}
\newcommand{\bff}{\mathbf{f}}
\newcommand{\bF}{\mathbf{F}}
\newcommand{\bo}{\mathbf{o}}
\newcommand{\bc}{\mathbf{c}}
\newcommand{\bm}{\mathbf{m}}
\newcommand{\bT}{\mathbf{T}}
\newcommand{\bQ}{\mathbf{Q}}
\newcommand{\bL}{\mathbf{L}}
\newcommand{\bl}{\mathbf{l}}
\newcommand{\ba}{\mathbf{a}}
\newcommand{\bE}{\mathbf{E}}
\newcommand{\bH}{\mathbf{H}}
\newcommand{\bd}{\mathbf{d}}
\newcommand{\br}{\mathbf{r}}
\newcommand{\be}{\mathbf{e}}
\newcommand{\bhe}{\hat{\mathbf{e}}}
\newcommand{\bb}{\mathbf{b}}
\newcommand{\bh}{\mathbf{h}}
\newcommand{\bhh}{\hat{\mathbf{h}}}
\newcommand{\beps}{\boldsymbol{\epsilon}}

\newcommand{\lambdahat}{\hat{\lambda}}
\newcommand{\lambdae}{{\lambda}^E}
\newcommand{\lambdaev}{{\boldsymbol{\lambda}}^E}
\newcommand{\lambdai}{{\lambda}^I}
\newcommand{\lambdaiv}{{\boldsymbol{\lambda}}^I}

\newcommand{\btheta}{\boldsymbol{\theta}}
\newcommand{\bpi}{\boldsymbol{\pi}}
\newcommand{\bphi}{\boldsymbol{\phi}}
\newcommand{\bPhi}{\boldsymbol{\Phi}}
\newcommand{\bmu}{\boldsymbol{\mu}}
\newcommand{\bSigma}{\boldsymbol{\Sigma}}
\newcommand{\bGamma}{\boldsymbol{\Gamma}}
\newcommand{\bbeta}{\boldsymbol{\beta}}
\newcommand{\bomega}{\boldsymbol{\omega}}
\newcommand{\blambda}{\boldsymbol{\lambda}}
\newcommand{\bkappa}{\boldsymbol{\kappa}}
\newcommand{\btau}{\boldsymbol{\tau}}
\newcommand{\balpha}{\boldsymbol{\alpha}}
\def\bgamma{\boldsymbol\gamma}

\newcommand{\argmin}{\operatornamewithlimits{argmin}}
\newcommand{\minimize}{\operatornamewithlimits{min}}
\newcommand{\minimizem}{\operatornamewithlimits{minimize}}
\newcommand{\maximize}{\operatornamewithlimits{max}}

 \newcommand{\ikron}[1] {\bI\otimes #1}
  \newcommand{\val}{\bar{\bx}}
    \newcommand{\train}[1]{{\phi(\bx_{#1})}}
    \newcommand{\ikronval}[1]{(\ikron{\phi(\val_{#1}))}}
\newcommand{\ikronvalT}[1]{(\ikron{\phi(\val_{#1})^T)}}
\newcommand{\ikrontrainT}{(\ikron{\train{i}^T)}}
\newcommand{\ikrontrain}[1]{(\ikron{\train{#1})}}
\newcommand{\ikrontrainAT}{(\ikron{\phi(\bx)^T)}}
\newcommand{\ikrontrainA}{(\ikron{\phi(\bx))}}
  \newcommand{\half}{\frac{1}{2}}
  \newcommand{\con}{C^{(c,u)}}
  \newcommand{\ineqcon}{D^{(d,v)}}
\newcommand{\conv}{C^{(u)}}
  \newcommand{\ineqconv}{D^{(v)}}
    \newcommand{\ig}{\frac{1}{\gamma}}
      \newcommand{\Bi}{\bB^{-1}}
 \newcommand{\kernel}{\bK}    
 \newcommand{\ikrontestT}{(\ikron{\test^T)}}
   \newcommand{\test}{\phi(\bx_*)}

% partial derivatives
 \newcommand{\pardev}[2]{\frac{\partial #1}{\partial #2}}
  \newcommand{\dw}{\delta\bw}
  \newcommand{\dW}{\delta\bW}
    \newcommand{\deps}{\delta \beps}
  
    \newcommand{\lab}{\mathcal{L}}
 	     \newcommand{\unlab}{\mathcal{U}}

\newcommand{\argmax}[1]{\underset{#1}{\mathrm{argmax}} \:}
\def\eop {{\noindent\framebox[0.5em]{\rule[0.25ex]{0em}{0.75ex}}}}

\newcommand{\todo}[1]{{\bf \textcolor{blue}{[TODO: #1]}}}

\begin{abstract}
\label{sec:abstract}

While depth sensors are becoming increasingly popular, their spatial resolution often remains limited. Depth super-resolution therefore emerged as a solution to this problem. Despite much progress, state-of-the-art techniques suffer from two drawbacks: (i) they rely on the assumption that intensity edges coincide with depth discontinuities, which, unfortunately, is only true in controlled environments; and (ii) they typically exploit the availability of high-resolution training depth maps, which can often not be acquired in practice due to the sensors' limitations. By contrast, here, we introduce an approach to performing depth super-resolution in more challenging conditions, such as in outdoor scenes. To this end, we first propose to exploit semantic information to better constrain the super-resolution process. In particular, we design a co-sparse analysis model that learns filters from joint intensity, depth and semantic information. Furthermore, we show how low-resolution training depth maps can be employed in our learning strategy. We demonstrate the benefits of our approach over state-of-the-art depth super-resolution methods on two outdoor scene datasets.

%This paper addresses the problem of depth super-resolution via the study of a multi-mode co-sparse analysis model, namely exploiting the co-sparse patterns for the registered image, depth and semantics by applying a triplet of learned analysis operators. Literature work ~\cite{kiechle2013} has demonstrated that depth super-resolution can be achieved by the co-sparse analysis between image and depth. In this work, we first show that the depth super-resolution result can be improved by exploiting the co-sparsity property among image, depth and semantics. We then tackle the challenging and realistic scenario in which no high resolution depth available for training the operators and show that our model can jointly estimate the super-resolution depth map and improve the noisy semantics results.

\end{abstract}
\section{Introduction}
\label{sec:introduction}

Depth sensors are becoming increasingly popular in many applications, such as virtual reality and autonomous navigation. While huge progress has been made in the development of such sensors, a typical example of which is the Kinect, for outdoor scenes, existing sensors remain limited in the spatial resolution they provide. As a consequence, depth super-resolution has emerged as a way to compute high-resolution depth maps from low-resolution ones.

\begin{figure}[t!]
\label{fig:introduction}
\vspace{-0.0cm}
\begin{small}
\begin{tabular}{cc}
\multicolumn{2}{c}{\hspace{-0.3cm}\includegraphics[width=0.8\linewidth]{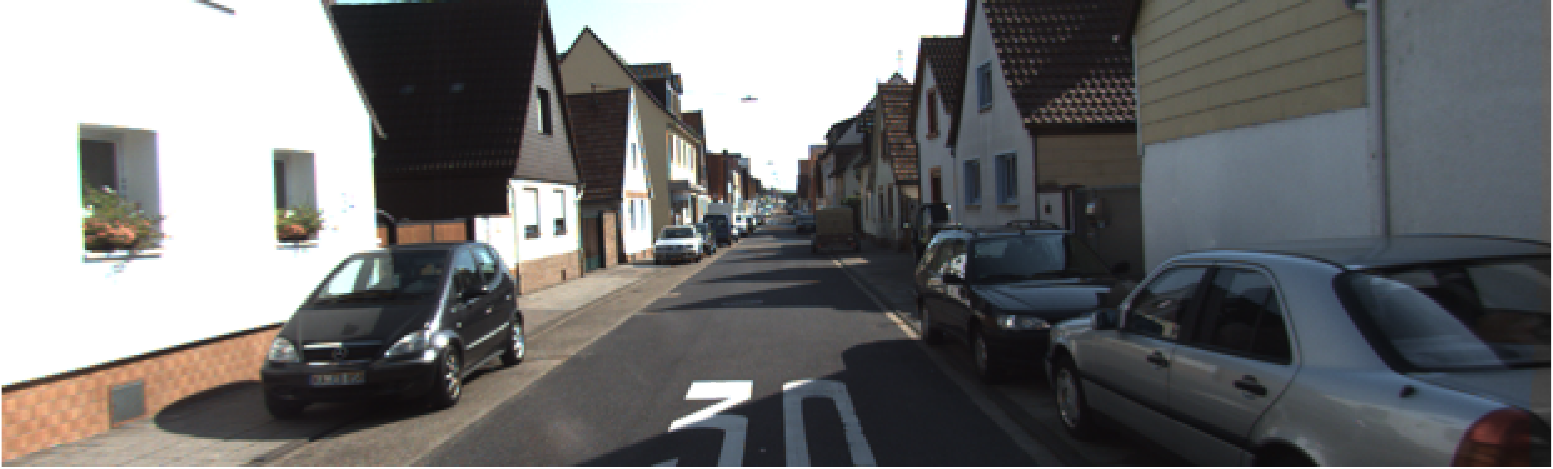}}\\
\hspace{-0.3cm}\includegraphics[width=0.5\linewidth]{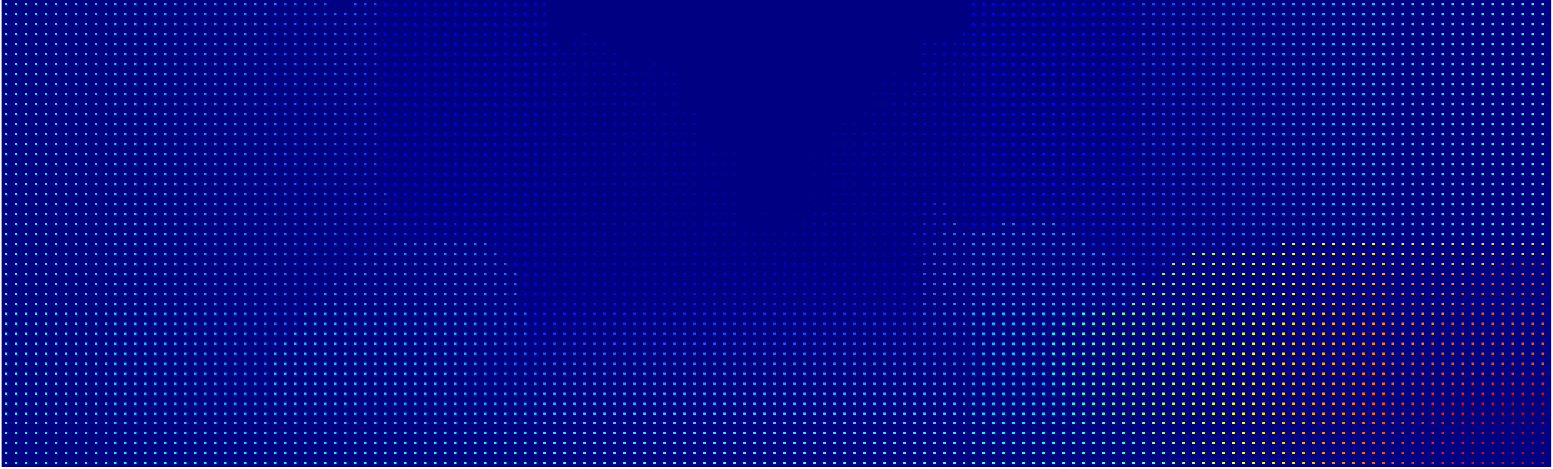}&
\hspace{-0.3cm}\includegraphics[width=0.5\linewidth]{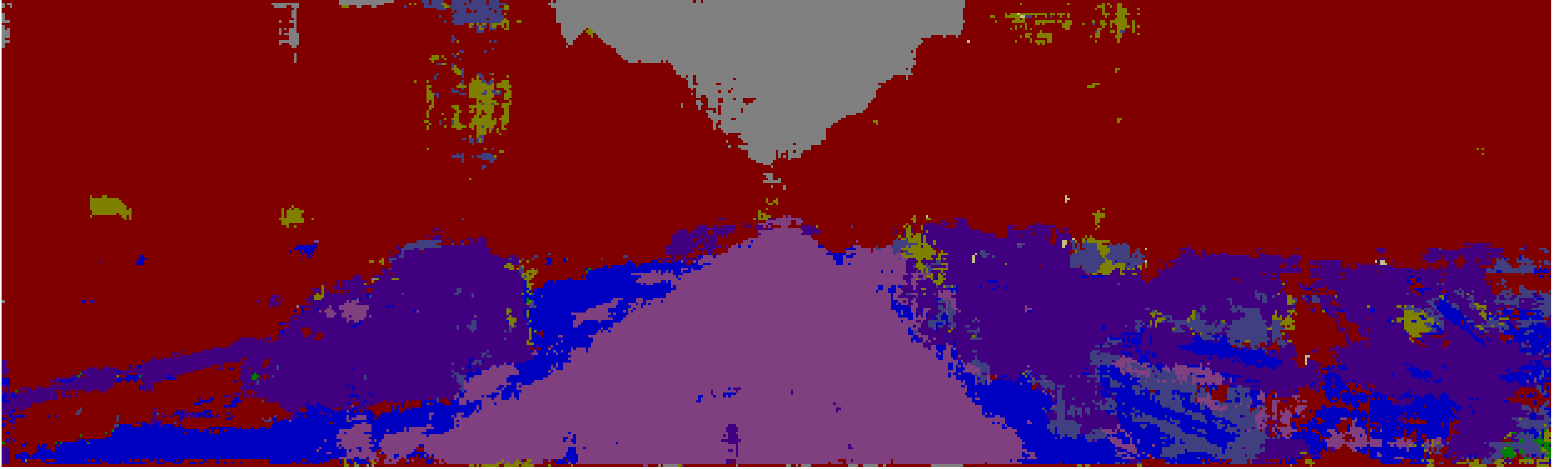}\\
\hspace{-0.3cm}\includegraphics[width=0.5\linewidth]{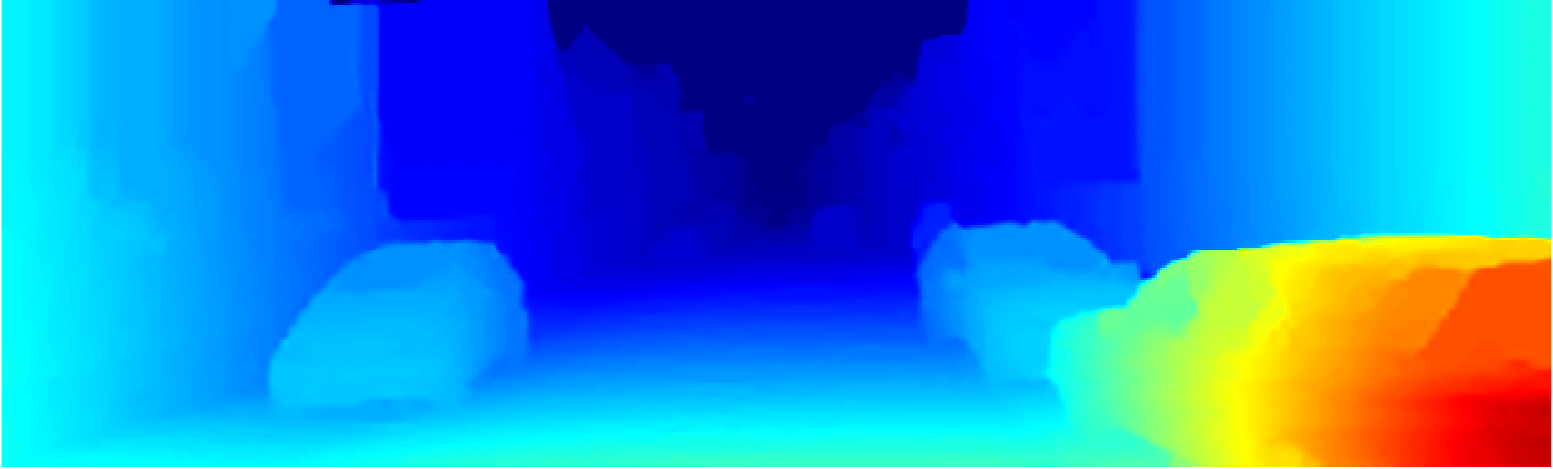}&
\hspace{-0.3cm}\includegraphics[width=0.5\linewidth]{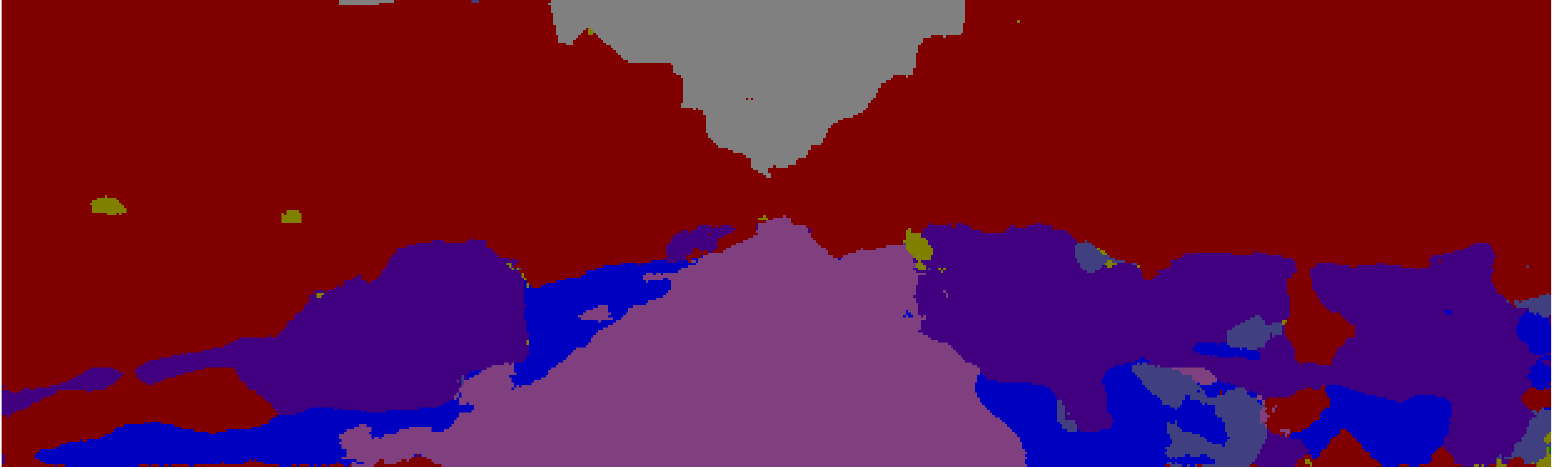}
\end{tabular}
\end{small}
\caption{{\bf Semantic-aware depth super-resolution in outdoor scenes.}{~\bf Top}: color image. {\bf Middle}: sparse disparity observations (left) and noisy semantics (right). {\bf Bottom}: estimated high-resolution disparity (left) and improved semantics using our method (right).~Best viewed in color.}\vspace{-3mm}
\end{figure}

In the past decade, many depth super-resolution methods have been proposed, such as filtering-based techniques~\cite{qi2013structure,yang2007spatial}, learning-based approaches~\cite{lu2014depth,gong2014guided} and CRF-based methods~\cite{Diebel05,Pktbk11ICCV}. In particular, a popular trend consists of exploiting the high-resolution intensity image corresponding to the low-resolution depth map to constrain the depth super-resolution process~\cite{yang2007spatial,gong2014guided}. The intuition behind such an approach is that the discontinuities in intensity space correspond to those in depth. Therefore, the high-resolution depth maps obtained in this manner should be less prone to over-smoothing. While the assumption of corresponding discontinuities is valid in nicely engineered environments, such as in the Middleburry dataset, which is the most popular benchmark for depth super-resolution algorithms, it typically does not hold in more realistic scenarios, and in particular, as illustrated in Fig.~\ref{fig:introduction}, in outdoor scenes, where disturbances such as strong shadows are common. More importantly, state-of-the-art algorithms typically assume to have access to high-resolution depth maps in a training stage to learn the operators they will apply at test time. Unfortunately, for outdoor scenes where depth sensors can only produce low-resolution depth maps, this kind of training data is unavailable.

In this paper, we introduce an approach to performing depth super-resolution in these challenging conditions. In particular, in addition to image gradient, we propose to rely on  semantic information, in the form of pixel-wise image labelings, to better constrain the depth super-resolution process. Semantic maps do not directly suffer from lighting conditions, and transitions between objects often correspond to depth discontinuities. Furthermore, thanks to the popularity of semantic labeling in computer vision, many datasets with ground-truth semantic maps are available for outdoor scenes, and existing methods produce increasingly accurate predictions. In a training stage, our approach therefore jointly exploits intensity and ground-truth semantics, and, importantly, tackles the realistic scenario where only {\it low-resolution} training depth maps are available. At test time, given an  image and low-resolution depth measurements, we generate a high-resolution depth map while simultaneously denoising a predicted semantic labeling.

More specifically, our approach relies on the co-sparse analysis model introduced in~\cite{kiechle2013}. The analysis model learns operators, \ie, filters, such that the response of those filters is sparse when applied to valid data patches. In our case, during training, we therefore seek to reconstruct a high-resolution depth map, while simultaneously learning operators that generate sparse responses from intensity, depth and semantics patches. At test time, since we cannot assume to have access to ground-truth semantic labels, we make use of existing frameworks~\cite{Gould2012JMLR,Krah11} to predict a semantic map. We then aim to jointly denoise this semantic map and predict a high-resolution depth map from low-resolution measurements, such that the learned operators yield sparse responses when applied to the predicted semantics and depth, together with the corresponding intensity. This process can be written as a series of convolutions, which we implemented on a GPU.

We evaluated our approach on two challenging outdoor datasets: the KITTI benchmark~\cite{geiger2012we} and Make3D~\cite{Saxena05NIPS}. Our experiments evidence the benefits of exploiting semantic information for depth super-resolution. Furthermore, they also show that our approach to learning operators from low-resolution depth maps comes at very little loss compared to training with high-resolution ones, which are typically unavailable in practice. Finally, as an additional benefit, our approach also yields improved semantic labelings.

%Depth information is broadly applied in many application fields, such as 3D movie, virtual reality and robotics.  These information are generally acquired by depth sensors, such as Microsoft Kinect,  LIDAR and ASUS XtionPro. However, the captured depth maps usually have limited spatial resolution and missing data.  In order to obtain high resolution depth image, literature works mainly rely on the observation that the discontinuities in color and depth image for the same scene are co-aligned~\cite{Diebel05}. 

% second paragraph talk about more realistic senario

% Third paragraph in talking about the semantic labeling and propose the 
%In recent years, semantic image segmentation has achieved much progress in parsing and analyzing the scene. Efficient semantic labeling algorithms have made it a
\section{Related Work}\label{sec:related}

%Depth super-resolution.
 
Densifying depth maps has attracted much attention in recent years due to the popularity of 3D sensors providing semi-dense measurements for both indoor and outdoor scenes. The resulting depth super-resolution approaches can be roughly categorized into two classes: the methods that rely on depth only, and those that also exploit other modalities. While  a large body of work addresses the problem of multiview depth fusion (\eg,~\cite{schuon2009lidarboost,dolson2010upsampling,castaneda2011stereo}), here, we focus on depth super-resolution from a single view.  

The first kind of methods treats depth as a 2D image and adopts techniques from the image super-resolution literature. For instance, Mac Aodha et al.~\cite{Aodha12} infer high-resolution depth from a single low-resolution depth image based on a generic database of depth patches at high-resolution. Hornacek et al.~\cite{hornacek2013depth} use the concept of self-similarity in 3D and super-resolve depth from a single low-resolution depth image. While effective in their context, these methods cannot leverage additional cues about the underlying scene, such as image intensity. In real-world applications, however, measurements from other modalities are typically available.

%\todo{add depth from single image? maybe not.}
%Patch Based Syntrhesis for Single Depth Image, ECCV 2012, (Infer a high resolution depth image from a Single low resolution depth, given a generic database of training patches in high resolution.) Dataset: MiddleBury stereo dataset, their own laser scanned depth.\\

%Depth Super Resolution by Rigid Body Self-Similarity in 3D. CVPR 2013. Datase: MiddleBury Dataset\\
The second category of methods therefore emerged as a solution to exploit the correlation between two modalities, typically using high-resolution intensity images to guide depth super-resolution. For instance, Diebel and Thrun~\cite{Diebel05} use an image contrast aware pairwise MRF to super-resolve a low-resolution range image. Park et al.~\cite{Pktbk11ICCV} improve this MRF model by incorporating outlier detection, richer image cues and smoothness priors over larger neighborhoods. In~\cite{wang2008stereoscopic}, color and depth images are jointly upsampled from low-resolution stereo images. In essence, however, these methods only capture pairwise intensity-sensitive depth smoothness, and thus lack the capacity to model higher-order depth patterns. %{\bf IS THIS ALSO TRUE FOR THE BILATERAL FILTER APPROACH? IF NOT, IT SHOULD BE MOVED TO THE NEXT PARAGRAPH.}

To explore longer-range dependencies, Yang et al.~\cite{yang2007spatial} perform depth super-resolution via a cross bilateral filtering operation on depth given the corresponding intensity image. Yu et al.~\cite{yu2013shading} use a shape-from-shading model to refine the depth estimation. Shen and Cheung~\cite{shen2013layer} propose a layered model for depth completion. More recently, Lu et al.~\cite{lu2014depth} introduced a low-dimensional subspace model for RGB-D patches, which was used to jointly complete and denoise depth maps. 
Perhaps most related to our work is the co-sparse analysis model of~\cite{kiechle2013}, which learns a set of joint analysis operators in intensity and depth to regularize depth super-resolution. By considering depth patches, this method can capture more complex patterns and enforce longer-range constraints on the depth map. The analysis model was also employed in~\cite{gong2014guided}, but by using pre-defined operators applied on depth only.

Despite this progress, however, all the existing works solely focus on depth completion in indoor scenes, such as  Middlebury images~\cite{scharstein2002taxonomy}, where the boundaries in intensity images are well aligned with depth discontinuities. Here, we consider the more challenging case of outdoor scenes, and introduce the use of semantics to handle large illumination changes and shadows in outdoor images. Furthermore, most learning-based approaches rely on high-resolution depth maps in the training stage, \eg, ~\cite{Aodha12,kiechle2013}, which are typically challenging to obtain in outdoor scenes. By contrast, we design a learning algorithm that only requires low-resolution depth data. %measurements  

%Joint depth reconstruction and semantic labeling.
%{\bf WE SHOULD PROBABLY CITE METHODS THAT USE RGB-D AS INPUT FOR SEMANTIC LABELING.}
While our main focus is depth super-resolution, our model also lets us improve noisy semantic labelings. The relationship between depth estimation and semantic labeling has been studied in several recent works. In particular, RGB-D images have been employed as input to semantic labeling algorithms~\cite{koppula2011semantic,ren2012rgb,gupta2013perceptual}. More related to our goal, Liu et al.~\cite{Liu10} use predicted semantic labels to estimate depth in outdoor scenes. Similarly, Ladick{\'y} et al.~\cite{Ladicky14} learn a joint classifier to predict semantic labels and depth values of image patches. Other approaches integrate depth reconstruction and semantic labeling using stereo images or image sequences as input~\cite{ladicky2012joint,Haene13}. Unlike our approach, however, these methods do not tackle the problem of depth super-resolution, and typically use high-resolution depth maps or 3D models in the training process. 

%\section{Proposed Approach}
\setlength{\belowdisplayskip}{2mm} %\setlength{\belowdisplayshortskip}{1mm}
\setlength{\abovedisplayskip}{2mm} %\setlength{\abovedisplayshortskip}{1mm}

\section{Semantic-Aware Depth Super-Resolution}
\label{sec:method}
Our goal is to estimate a high-resolution depth map ${\bf D}\in\mathbb {R}^n$ from  sparse and noisy measurements $\hat{{\bf D}}\in \mathbb{R}^m$, where, typically, $m \ll n$. This depth super-resolution process can be expressed as the problem of finding ${\bf D}$, such that
\begin{equation}
\hat{{\bf D}} = \mathcal{A}{\bf D}+ {\bf e}_D,\label{eq:basic}
\end{equation}
where $\mathcal{A}$ models the down-sampling process and ${\bf e}_D\in \mathbb{R}^m$ denotes the noise.
% and sampling errors. 
Computing ${\bf D}$ from Eq.~\ref{eq:basic} is an ill-posed problem since $m$ is significantly smaller than $n$. Here, we aim to address this problem by incorporating both intensity and pixel-level semantics to regularize depth estimation.     

In particular, we adopt the analysis model framework of~\cite{HaweKD13TIP} and introduce an approach to building a joint prior on image, depth and semantic patches. The analysis model captures the signal structure by learning an analysis operator ${\bf \Omega}\in \mathbb{R}^{k\times n}$, such that applying ${\bf \Omega}$ to the input signal yields a sparse output vector. Here, we design a trimodal co-sparse analysis approach to depth super-resolution in outdoor scenes by exploiting the strong correlations in the discontinuity patterns of multimodal cues, \ie, intensity, depth and semantics. Furthermore, we introduce a new learning method that prevents the requirement for high-resolution training depth maps, and thus better suits the outdoor setting in which most sensors simply cannot produce high-resolution depth. We first discuss our depth super-resolution framework in this section, and then present our learning approach in Section~\ref{sec:learn}.

\subsection{A Trimodal Co-Sparse Analysis Model}
\label{sec:trimodal}
%In sparse coding, signal information is encoded in the non-zero entries of the sparse code. 
We now introduce our trimodal co-sparse analysis model. To this end, let us first consider the case of a single patch from a single modality, \eg, an image patch. Given an image patch $\bX$ (in vector form) and an operator ${\bf \Omega}$, the 
analysis model encodes the structure of the signal by the zero entries in the vector ${\bf \Omega X}$. In other words, the signal ${\bf X}$ lies in the null space of the matrix composed by a subset of the rows of ${\bf \Omega}$. %corresponding to the zero entries of the analyzed vector. 
Specifically, this subset, defined as the~\emph{co-support} of ${\bf X}$, can be written as
\begin{equation}
\Gamma:=\{j|({\bf \Omega X})_j = 0\},
\end{equation}  
where$(\cdot)_j$ denotes the $j$-th element of a vector.
In our case, we aim to capture the patterns common to intensity, depth and semantic patches. To this end, let ${\bf I}\in \mathbb{R}^{n_I}$ and ${\bf D}\in \mathbb{R}^{n_D}$ denote a vectorized intensity patch and depth patch, respectively. For the semantics, we encode the class label of a pixel as an $L$-dimensional vector representing the probability for the pixel to belong to each of the $L$ classes of interest. The semantic information of the pixels in a patch can therefore be grouped in a vector ${\bf S}\in \mathbb{R}^{Ln_S}$.

Let us then denote by ${\bf \Omega_I} \in \mathbb{R}^{k\times n_I}$, ${\bf \Omega_D} \in \mathbb{R}^{k\times n_D}$, and ${\bf \Omega_S} \in \mathbb{R}^{k\times L n_S}$ the operators corresponding to ${\bf I}$, ${\bf D}$ and ${\bf S}$, respectively.  Similarly, let $\Gamma_I$, $\Gamma_D$, and $\Gamma_S$ be the co-support of ${\bf I}$, ${\bf D}$, and ${\bf S}$. Our model assumes that the structures of the different modalities are correlated, and thus that $\Gamma_I$ ,$\Gamma_D$, and $ \Gamma_S$ are statistically highly dependent. In other words, the probability for a row index $j\;, 1\leq j \leq k$, to be in the co-support of one modality should be higher if $j$ also is in the co-support of the other modalities.

%\todo{Do we really care about this probabilistic interpretation? Is this coming from the JID paper? I would tend to remove this.}
%We can further explain it as the conditional probability of $j \in \Gamma_D$ given $j \in \Gamma_I$ and $j \in \Gamma_S$ is higher than the unconditional probability, i.e. 
%\begin{equation}
%\label{eq:cosparseCondition}
%Pr(j\in \Gamma_D|j\in \Gamma_I,j\in \Gamma_S) \geq Pr(j\in \Gamma_D) .
%\end{equation}

More concretely, for each row index $j$, the dependency between the co-supports of different modalities can be modeled by the function
\begin{align}\label{eq:sparse_j}
&g_j({\bf\Omega_I I},{\bf \Omega_D D},{\bf \Omega_S S}) =\nonumber \\
&\log\left(1+\nu_I({\bf \Omega}_I{\bf I})_j^2+\nu_D({\bf \Omega}_D {\bf D})_j^2+ \nu_S({\bf \Omega}_S {\bf S})_j^2\right)\;,
\end{align}
where $\nu_I$, $\nu_D$, and $\nu_S$ are the weights of the different modalities. This function, which was shown to be more effective than other sparsity-inducing functions such as the $l_0$ and $l_1$ norms~\cite{Chen14TIP}, will be zero only if the three operators yield a zero value for index $j$, \ie, if $j$ is in the co-support of all modalities. Therefore, the global cost function
\begin{equation}\label{eq:sparse}
g({\bf \Omega_I I},{\bf\Omega_D D},{\bf\Omega_S S}) = \sum_{j=1}^k  g_j({\bf\Omega_I I},{\bf \Omega_D D},{\bf \Omega_S S})
\end{equation}
will reach its minimum if the co-supports of the three modalities coincide exactly. 

While the prior induced by this cost favors discontinuities aligned in the three modalities, thanks to the weighted contributions of semantics and intensity, it still allows depth boundaries to occur between regions of the same semantic class.
%. Nevetheless, we note that the depth continuities are affected by a weighted contribution from both semantic and image mode, and thus the model also allows depth boundaries between regions in the same semantic class.}   
%\textcolor{red}{Note that although this three modality co-sparsity constraint prefers co-aligned smoothness and discountinuities, we can still keep the depth discontinuities for the same object by adjusting the contribution weight among the three different modes in addition to be close to the observations in the data term.}
As such, and as discussed in the remainder of this paper, this cost function is therefore very-well suited to both perform depth super-resolution and learn the analysis operators.
Below, we start by formulating the depth super-resolution problem.

\subsection{Depth Super-Resolution with Co-Sparsity}

We now turn to the problem of performing depth super-resolution for an entire image, given the co-sparse analysis operators ${\bf \Omega_I}$, ${\bf \Omega_D}$, ${\bf \Omega_S}$ for the three modalities, \ie, intensity, depth and semantics. Since these operators are defined on patches, we use the convolution operation to apply them to the entire image. We then define an energy that combines a regularizer based on our trimodal co-sparse analysis model with data terms accounting for the sparse depth measurements and the noisy semantic label predictions.

More specifically, let $\hat{{\bf D}}$ be the vector of sparse depth measurements given as input with corresponding intensity image ${\bf I}$. From ${\bf I}$, and given a pre-trained semantic classifier, we can predict a noisy semantic map $\hat{{\bf S}}$, as is commonly done in image-based semantic labeling. Given these two inputs, we seek to find the high-resolution depth map ${\bf D}$ and noise-free semantic labels ${\bf S}$ that minimize a data term of the form
%\vspace{-1mm}
\begin{equation}
E_d({\bf D},{\bf S}) = \|\mathcal{A}{\bf D}-\hat{{\bf D}}\|^2_2 + \lambda\|{\bf S}-\hat{{\bf S}}\|^2_2,
\end{equation}
where $\lambda$ encodes the relative importance of the two terms. This data term encourages the estimated depth to be close to the sparse observations, and the estimated semantic map to be close to the noisy one. On its own, of course, this data term is not sufficient since, for instance, nothing constrains the depth values with no measurements and nothing relates the depth to the semantics. We propose to make use of our trimodal co-sparse analysis model to address this issue.

As mentioned above, we convolve our patch-based analysis operators with the entire image. To this end, let us define position operators $\{\bP^I_{ij},\bP^D_{ij},\bP^S_{ij}\}$ that extract vectorized local patches centered at pixel location $(i,j)$ and of size $\sqrt{n_I}\times\sqrt{n_I}$, $\sqrt{n_D}\times\sqrt{n_D}$, $\sqrt{n_S}\times\sqrt{n_S}$ from ${\bf I}$, ${\bf D}$, and ${\bf S}$, respectively. Following the trimodal co-sparse analysis model of Section~\ref{sec:trimodal}, we define a regularizer of the form
%\vspace{-3mm}
\begin{equation}
\scalebox{0.9}{$E_s({\bf D},{\bf S} | {\bf I}) $}= \frac{1}{n}\sum_{r=1}^ng({\bf \Omega}_I\bP_r^I{\bf I},\;{\bf \Omega}_D\bP_r^D{\bf D},\;{\bf \Omega}_I\bP_r^S{\bf S}),
\end{equation}
where $n$ is the total number of pixels in the image. This regularizer now relates the estimated depth to the estimated semantics, and links them to the given intensity image.

By combining this regularizer with our data term, we formulate joint depth super-resolution and semantic labeling as the optimization problem
%\vspace{-1mm}
\begin{equation}
({\bf D}^\star,{\bf S}^\star)= \argmin_{\{{\bf D}\in \mathbb{R}^n,{\bf S}\in \mathbb{R}^{L n}\}} \eta E_s({\bf D},{\bf S} | \bI) + E_d({\bf D},{\bf S})\;,
\label{eq:superres_final}
\end{equation}
where $\eta$ is a weight that adjusts the influence of the regularizer. In practice, we make use of  a conjugate gradient descent algorithm to solve~\eqref{eq:superres_final}.

\vspace{-2mm}
\section{Learning Multimodal Analysis Operators}\label{sec:learn}

The semantic-aware depth super-resolution method introduced in Section~\ref{sec:method} relies on given analysis operators. We now turn to the problem of learning these operators from training data.
In particular, we consider two different settings: one where we exploit high-resolution training depth maps, and a more realistic one where we only rely on low-resolution training depth.
In both cases, we assume that the ground-truth depth, the intensity image and the semantic probability map are registered. 

\subsection{Learning with High-Resolution Depth}
Given a set of high-resolution multimodal patches $\{{\bf I}_m,{\bf D}_m,{\bf S}_m\}_{m=1}^{M}$, we make use of the trimodal co-sparse analysis model to learn the analysis operators corresponding to the three modalities. Intuitively, we search for operators such that the co-supports of those three modalities are highly dependent. By making use of the cost defined in Eq.~\ref{eq:sparse} for a single patch, we can express this as the loss
%\vspace{-0mm}
\begin{equation}
\label{eq:sparsity_learn}
\scalebox{0.9}{$L_s({\bf \Omega}_I, {\bf \Omega}_D, {\bf \Omega}_S)  
=\frac{1}{M}$}\sum_{m=1}^Mg({\bf \Omega}_I{\bf I}_m, {\bf \Omega}_D {\bf D}_m, {\bf \Omega}_S {\bf S}_m).
\end{equation}
%\vspace{-1mm}
Minimizing this loss alone, however, suffers from trivial solutions (\ie, the operators can simply go to 0). To avoid these trivial solutions, we make use of the constraints and priors introduced in~\cite{HaweKD13TIP}.

More specifically, as a first constraint, we assume that the (transposed) analysis operators lie on the oblique manifold. In other words, each operator ${\bf \Omega}_{X}$, where $X$ can be either $D$, $I$, or $S$, is constrained as
\begin{equation}
\begin{split}
&\hspace{-2cm}{\bf \Omega}^T_X \in OB(n,k)= \\\{ {\bf \Omega}^T\in \mathbb{R}^{n \times k} |
& \mathrm{rank}({\bf \Omega}) = n,\mathrm{ddiag}({\bf \Omega} {\bf \Omega}^T) = \mathbf{I}_k \}\;,
\end{split}
\end{equation}
which indicates that ${\bf \Omega}_X$ must have full rank and unit-norm rows. Furthermore, we also make use of two additional priors introduced in~\cite{HaweKD13TIP} that have proven effective to learn meaningful operators. These priors can be expressed as
\begin{equation}
\begin{split}
h({\bf \Omega}_X)& =-\frac{1}{n\log(n)}\mathrm{logdet}\left(\frac{1}{k}{\bf \Omega}_X^T{\bf \Omega}_X\right)\;,\\
r({\bf \Omega}_X)& =-\sum_{1\leq i<j\leq k} \log\left(1-(\langle{\bf \omega}_X^i, {\bf \omega}_X^j\rangle)^2\right)\;,
\end{split}
\end{equation}
where ${\bf \omega}_X^i$ is the $i^{th}$ row of ${\bf \Omega}_X$, and $\langle\cdot\rangle$ denotes the inner product between two vectors. The function $h({\bf \Omega})$ encodes a rank constraint, and $r({\bf \Omega})$ encourages the rows of ${\bf \Omega}$ to be linearly independent. These priors can then be grouped in a regularizer of the form
\begin{align}
L_c({\bf \Omega}_I, {\bf \Omega}_D, &{\bf \Omega}_S)=\kappa_Ih({\bf \Omega}_I) +\kappa_Dh({\bf \Omega}_D)+\kappa_Sh({\bf \Omega}_S)\nonumber\\
&+\mu_Ir({\bf \Omega}_I)+\mu_Dr({\bf \Omega}_D)+\mu_Sr({\bf \Omega}_S)\;,
\label{eq:regul}
\end{align}
where $\kappa_{\{I,D,S\}}$ and $\mu_{\{I,D,S\}}$ are weights that adjust the influence of the individual terms.

By combining the loss of Eq.~\ref{eq:sparsity_learn} and the regularizers discussed above, learning can be expressed as 
\begin{align}
\label{eq:learn}
&\hspace{0.3cm}\min_{{\bf \Omega}_{\{I,D,S\}}} \eta L_s({\bf \Omega}_I, {\bf \Omega}_D, {\bf \Omega}_S) + L_c({\bf \Omega}_I, {\bf \Omega}_D, {\bf \Omega}_S) \\
&\resizebox{0.7\columnwidth}{!}{${\rm s.t.}\;\; {\bf \Omega}_{I}^T \in OB(n_I,k), \; {\bf \Omega}_{D}^T \in OB(n_D,k),\; {\bf \Omega}_{S}^T \in OB(Ln_S,k),$} \nonumber
\end{align}
where $\eta$ denotes the relative weight of the two terms. In practice, we use the geometric conjugate gradient descent method of~\cite{AbsMahSep2008} to solve this problem.

\subsection{Learning with Low-Resolution Depth}

In a more realistic scenario, such as for outdoor scenes, high-resolution depth maps are typically not available, even at training time. To tackle this scenario, we introduce an approach that exploits low-resolution depth information during training. More precisely, our training set is now comprised of registered high-resolution intensity patches, low-resolution depth patches and ground-truth semantic patches, which we denote by $\{{\bf I}_m,{\hat {\bf D}}_m,{\bf S}_m\}_{m=1}^{M}$. 

We follow a similar intuition as in the high-resolution case, and search for operators that yield co-support sets which are highly dependent among the three modalities. To this end, we can re-use the loss of Eq.~\ref{eq:sparsity_learn}, as well as the regularizer of Eq.~\ref{eq:regul}. Note, however, that the loss is defined on a complete depth map, which we do not have access to here. To address this issue, we propose to simultaneously estimate the high-resolution depth map ${\bf D}_m$ of each training patch and the analysis operators. 

To this end, we make use of a similar data term as for depth super-resolution. %\textcolor{red}{\st{This can be expressed as}}
This data term encourages the reconstructed depth map to be consistent with the low-resolution depth map during the training process, and can thus be written as
%\vspace{-4mm}
\begin{equation}
L_d(\{{\bf D}_m\})  = \frac{1}{M}\sum_{m=1}^M\|\hat{{\bf D}}_m - \mathcal{A}{\bf D}_m\|_2^2\;,
\end{equation}
where $\mathcal{A}$ is the downsampling operator.

This lets us express learning with low-resolution depth maps as the optimization problem 
\begin{align}
\label{eq:learn}
&\resizebox{0.65\columnwidth}{!}{$\min\limits_{{\bf \Omega}_{\{I,D,S\}},\{{\bf D}_m\}} L({\bf \Omega}_I, {\bf \Omega}_D, {\bf \Omega}_S,\{{\bf D}_m\}) + L_d(\{{\bf D}_m\})$} \\
&\resizebox{0.7\columnwidth}{!}{${\rm s.t.}\;\; {\bf \Omega}_{I}^T \in OB(n_I,k), \; {\bf \Omega}_{D}^T \in OB(n_D,k),\; {\bf \Omega}_{S}^T \in OB(Ln_S,k),$} \nonumber
\end{align}
where
\begin{align}
\label{eq:learn2} 
L({\bf \Omega}_I, {\bf \Omega}_D, {\bf \Omega}_S,\{{\bf D}_m\}) &= \eta L_s({\bf \Omega}_I, {\bf \Omega}_D, {\bf \Omega}_S,\{{\bf D}_m\}) + L_c({\bf \Omega}_I, {\bf \Omega}_D, {\bf \Omega}_S)
\end{align}
%is the objective function of the high-resolution case, which 
is now a function of the estimated depth. %To obtain a solution to this problem, we follow the alternating minimization strategy described below.

%&({\bf \Omega}_I, {\bf \Omega}_D, {\bf \Omega}_S, \{{\bf D}_m\}_{m=1}^M) \\ 

%Therefore, our training process is  to estimate the operators (${\bf \Omega_I}$, ${\bf \Omega_D}$, ${\bf \Omega_S}$) as well as reconstructing the high resolution patches $\{{\bf D}_m\}_{m=1}^M$, which can be achieved by solving the following equation

%\paragraph{Data Term}
%The data term enforces that the reconstructed depth map is consistent with the low resolution depth map during the training process. We define the data term as
%\begin{equation}
%L_d({\bf D}_m,\hat{\bD}_m)  =\|\hat{{\bf D}}_m - \mathcal{A}{\bf D}_m\|_2^2.
%\end{equation}
%where $\mathcal{A}$ is the downsampling operator and $\hat{\bD}$ is the observed low resolution depth. 
%\vspace{-0.3cm}
%\paragraph{Alternating Minimization:}\mbox{}\\
Problem~\eqref{eq:learn} has two different types of variables, \ie, the operators ${\bf \Omega}_I$, ${\bf \Omega}_D$ and ${\bf \Omega}_S$ that lie on the oblique manifold, and the depth maps $\{ {\bf D}_m \}_{m=1}^M$ that are in Euclidean space. We therefore follow an alternating approach to computing these variables. In particular, we initialize $\{ {\bf D}_m \}_{m=1}^M$ by interpolation of the low-resolution depth maps. As a first step, we fix the depth variables and optimize the operators by using geometric conjugate gradient descent on the manifold~\cite{AbsMahSep2008}. Then, we fix the resulting operators and optimize the depth variables using a conjugate gradient descent method in Euclidean space. We perform this alternating optimization for a fixed number of iterations. As evidenced by our experiments, this alternating strategy proved sufficient to achieve good results.

\vspace{-3mm}
\section{Experiments}
\label{sec:exp}
We evaluated our approach on two challenging outdoor datasets: the KITTI benchmark~\cite{geiger2012we} and Make3D~\cite{Saxena05NIPS}. To demonstrate the effectiveness of our method, we provide quantitative and qualitative evaluation results on both depth and semantics.

As mentioned in Section~\ref{sec:method}, we train the operators on patches. To this end, we extracted corresponding square patches of size $\sqrt{n_I} = \sqrt{n_D} = \sqrt{n_S} = 5$ from the intensity images, the depth maps and the semantic maps. The intensity and depth patches were then reshaped as $25$-dimensional vectors. For the semantics, since we represent the label of each pixel as an $L$-dimensional vector encoding the probability of each label, vectorizing a semantic patch yields a $25\cdot L$-dimensional vector. We then subtracted the mean of each training patch, but did not normalize the patches. As is common practice with the analysis model, we learned redundant (or over-complete) operators, \ie, operators with more rows than the dimensionality of the vectorized patches. In particular, to trade-off accuracy and efficiency, we chose a factor $1.2$ for the redundancy of ${\bf \Omega}_S$, which corresponds to the modality with highest dimensionality. We validated all the parameters of our model on the validation set of KITTI in a grid-search fashion. During training, we estimated the operators and the high-resolution depth maps by solving~\eqref{eq:learn}. 

While we use the ground-truth semantics to learn the operators during training, we can only realistically have access to noisy semantic maps at test time. To generate such noisy semantic labels, we used existing multi-class classifiers~\cite{Gould2012JMLR,Krah11}. We then represented the semantic information of each pixel as the classification confidence of each class. During test, we solved~\eqref{eq:superres_final} to estimate the high-resolution depth and clean the noisy semantic map.

\subsection{KITTI Dataset}
The original KITTI dataset only provides the raw sparse disparity. Therefore, to quantitatively evaluate our method, we tested it on a subset of the KITTI data provided by Ladick{\'y} et al.~\cite{Ladicky14} and consisting of $60$ images aligned with ground-truth dense disparity maps and semantic labels. We made use of the training and test splits provided with this data, namely, the first $30$ images for training and the remaining ones for testing. We reserved $10$ images from the original training set for validation purpose. We artificially created the low-resolution depth by downsampling the high-resolution ones by a factor $d\in\{2,4,8\}$ in both dimensions of the image. Therefore, our input for training includes the high-resolution images, the low-resolution depth maps and the ground-truth semantic labels. We used $L=9$ classes in the dataset, as suggested in~\cite{Ladicky14}. With our $1.2$ redundancy factor, this yields ${\bf \Omega}_S \in \mathbb{R}^{270\times 225}$ (\ie, $5\times 5 \times 9 = 225$). Since our trimodal co-sparse analysis model assumes that the operators all have the same number of rows, this yields ${\bf \Omega}_I \in \mathbb{R}^{270\times25}$ and similarly ${\bf \Omega}_D \in \mathbb{R}^{270\times25}$. 

In the reconstruction process, the relative weight between the sparsity term and the semantic data term is fixed. However, we iteratively changed $\eta$ and restarted the conjugate gradient descent ten times in order to reach a better local minimum. In particular, following the strategy of~\cite{kiechle2013}, we started with $\eta = 30$ and kept shrinking it to a final value $\eta = 0.04$. Our validation procedure resulted in the parameter values $\nu_I = \nu_D = 3, \nu_S = 30$ and $\lambda = \eta$.

%and vectorized them for learning (${\bf \Omega}_I,{\bf \Omega}_D$). We represent the semantics as volume data. In particular, we represent the semantic label for each pixel by using a binary vector. For instance, we use $[0\; 1\; 0\; 0\; 0\; 0\; 0\; 0\; 0]$ to denote a certain pixel belonging to $class~2$. Therefore, the semantic information for the entire image can be represented by a tensor of size $h\times w \times l$, where $h$, $w$ and $l$ denote the image height, width and semantic class number respectively. We extract a volume of size $5\times 5\times l$ in the same location as the image and depth patch. The volume is vectorized to form the training example for learning ${\bf \Omega}_S$. 

\begin{table}[t!]
\vspace{-0mm}
\centering
\begin{tabular}{|c | c| c| c|}
\hline
method & 2x & 4x & 8x\\ 
\hline
 bicubic & 1.6744& 2.2466 & 3.4520\\
\hline
nearest-neighbour& 2.0985 & 3.0034 & 4.1902\\
\hline
JID~\cite{kiechle2013} &  1.0169 & 1.5257 & 2.1921\\
\hline
ours-trainHR&{\bf 0.9673} &{\bf 1.4906} &{\bf 2.1904}\\
\hline
%ours-trainHR&0.9673 & 1.4906 &2.1904\\
%\hline
\end{tabular}
\vspace{1mm}
\caption{Comparison of the results of our approach trained using high-resolution depth maps with three depth super-resolution baselines, including the state-of-the-art JID algorithm. We report the RMSE for three downsampling factors, \ie, 2, 4 and 8 times, respectively. These results evidence that depth super-resolution benefits from using semantic information.}
\label{Tab:compState}
\vspace{-3mm}
\end{table}

\begin{figure}[t!]
	\centering
\hspace{-0.5cm}	\includegraphics[width=0.9\linewidth]{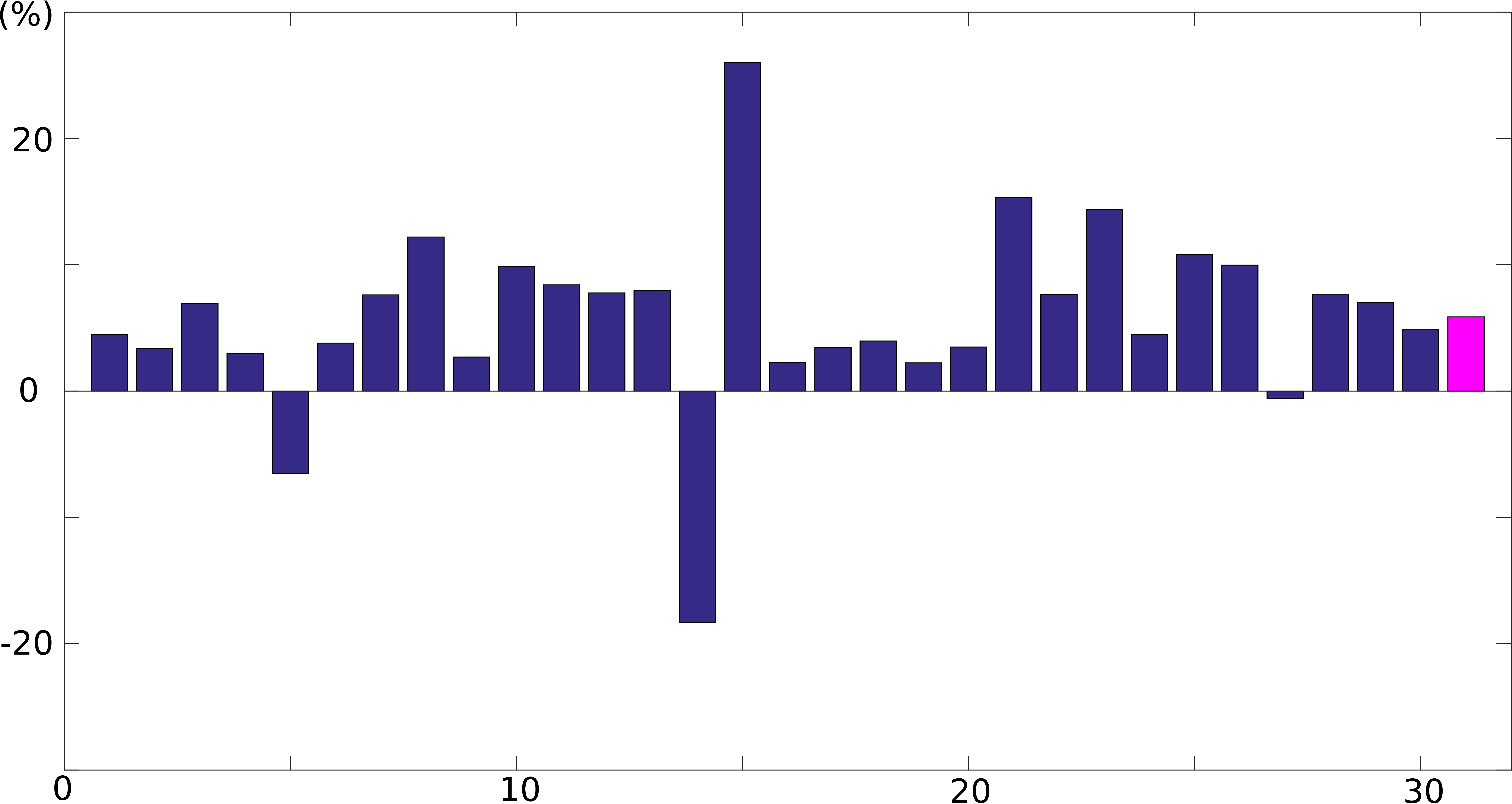} 
	\caption{Quantitative comparison of the results obtained by our method and of the JID results for each test image for 2x up-sampling. We show the relative improvement of our method over JID computed as $(JID_{RMSE} - Ours_{RMSE})/JID_{RMSE}$, where $Ours_{RMSE}$ and $JID_{RMSE}$ denote the RMSE of our results and OF the JID results, respectively. Furthermore, we show the mean relative improvement in magenta, which is $5.8733\%$. These results evidence that, for most images, our improvement is significant. 
	}\vspace{-2mm}
	\label{fig:2xbarfig}
\end{figure}

%\begin{figure}[t!]
%%\centering
%\includegraphics[width=0.9\linewidth]{./figures/kitti/ICCVRes.pdf} 
%\caption{Quantitative comparison of the RMSE obtained by our method (cyan) with those of JID (magenta) for each test image and for 2x up-sampling (lower is better). These results evidence that, for a number of images, our improvements are significant. 
%}
%\label{fig:2xbarfig}
%\vspace{-3mm}
%\end{figure}

As a first quantitative evaluation, we compare the results obtained with our approach when learning the operators with high-resolution depth with the following baselines: nearest-neighbor interpolation, bicubic interpolation, and the state-of-the-art \emph{JID}~\cite{kiechle2013}, which also exploits high-resolution training depth maps.
In Table~\ref{Tab:compState}, we report the root-mean-square-error (RMSE) of all the methods. Our approach yields lower RMSE than all the baselines and of the JID results. In particular, it outperforms~\emph{JID}, which demonstrates the benefit of exploiting semantic information. 
%Note that our reported error are computed as the statistics over the entire dataset. 
%In Fig.~\ref{fig:2xbarfig}, we provide the RMSE value for each test image for 2x up-sampling. Note that, on several images, our approach yields a large improvement over JID. Note also that the improvement achieved by our method is of similar magnitude to what is typically reported in the literature, \eg,~\cite{kiechle2013}.
In Fig.~\ref{fig:2xbarfig}, we show the relative improvement of our method over JID, computed as $(JID_{RMSE} - Ours_{RMSE})/JID_{RMSE}$, where $Ours_{RMSE}$ and $JID_{RMSE}$ denote the RMSE error of our results and of the JID results, respectively. Note that our approach yields a large improvement over JID on most images. Note also that the improvement achieved by our method is of similar magnitude to what is typically reported in the literature, \eg,~\cite{kiechle2013}.

\begin{figure}[t!]
\vspace{-0.15cm}
\begin{small}
\begin{tabular}{cc}
\hspace{-0.0cm}\includegraphics[width=0.5\linewidth]{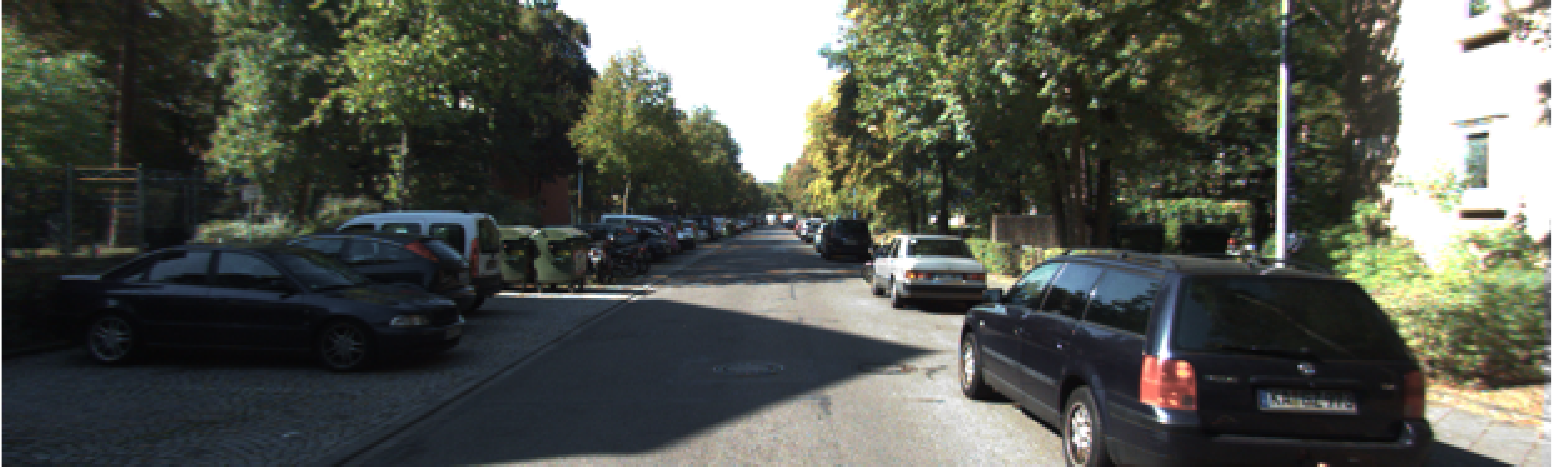} & 
\hspace{-0.0cm}\includegraphics[width=0.5\linewidth]{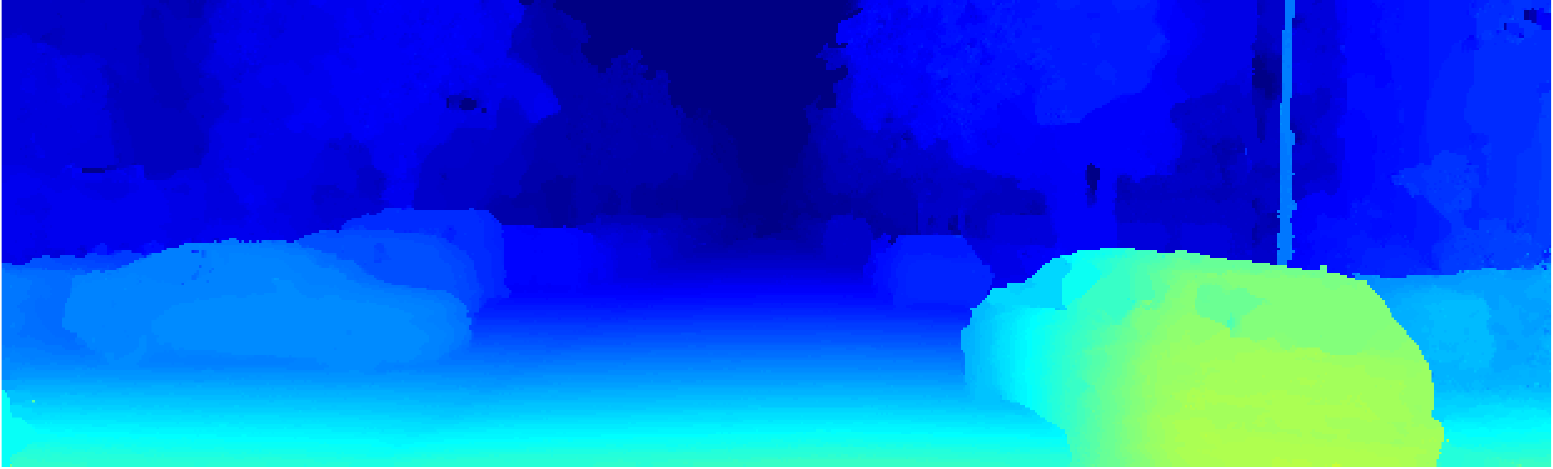}\\
\hspace{-0.0cm}\includegraphics[width=0.5\linewidth]{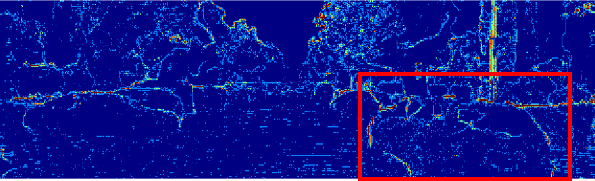} & 
\hspace{-0.0cm}\includegraphics[width=0.5\linewidth]{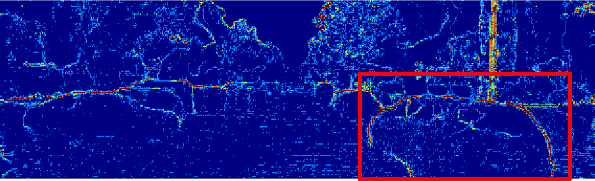}\\
\hspace{-0.0cm}\includegraphics[width=0.3\linewidth]{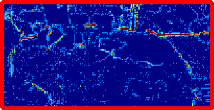} & 
\hspace{-0.0cm}\includegraphics[width=0.3\linewidth]{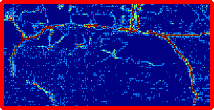}
\end{tabular}
\end{small}
\vspace{-1.5mm}
\caption{Qualitative comparison of the results of our approach trained using high-resolution disparity with~\emph{JID}, which also relies on high-resolution training disparity maps. {\bf Top:} RGB image and ground-truth disparity. {\bf Middle:} Absolute difference between our results and ground-truth, and between the JID results and ground-truth, respectively. Blue denotes small errors and red large ones. {\bf Bottom:} Close-up view of the portion highlighted in red. Note that our approach, which exploits semantics, yields lower errors near object boundaries (\eg, the car on the right). Best viewed in color.}\vspace{-2.5mm}
\label{fig:compJID}
\end{figure}

In Fig.~\ref{fig:compJID}, we provide a qualitative comparison of our results with those of JID, which evidences that semantic information helps reducing the errors near object boundaries. 

To show that our approach effectively exploits the semantic information, we re-ran the previous experiment with ground-truth semantics. This gave the following accuracies: 2x: 0.9418, 4x: 1.4743, 8x: 2.1590. This illustrates that better semantics can indeed further improve our results. Thus, as progress is made in semantic labeling, our method will produce increasingly accurate high-resolution depth maps.

\begin{table}
	\centering
	\begin{tabular}{|c | c| c| c|}
		\hline
		method & 2x & 4x & 8x\\ 
		\hline
		ours-trainLR&1.0546 & 1.5205 & {\bf 2.1861}\\
		\hline
		ours-trainHR&{\bf 0.9673} &{\bf 1.4906} & 2.1904\\
		\hline
	\end{tabular}
	\vspace{0.1cm}
	\caption{Comparison of the results obtained by learning with high-resolution depth maps (ours-trainHR) and with low-resolution ones (ours-trainLR). Note that learning with low-resolution depth maps yields very little loss in accuracy.}\vspace{-4.5mm}
	\label{Tab:oursComp}
\end{table}

\begin{figure*}[t!]
\vspace{-0.2cm}
\begin{small}
\begin{tabular}{cccc}
	\rotatebox{90}{\parbox{0.9cm}{\centering RGB\\ Image}}&\hspace{-0.0cm}\includegraphics[width=0.305\linewidth]{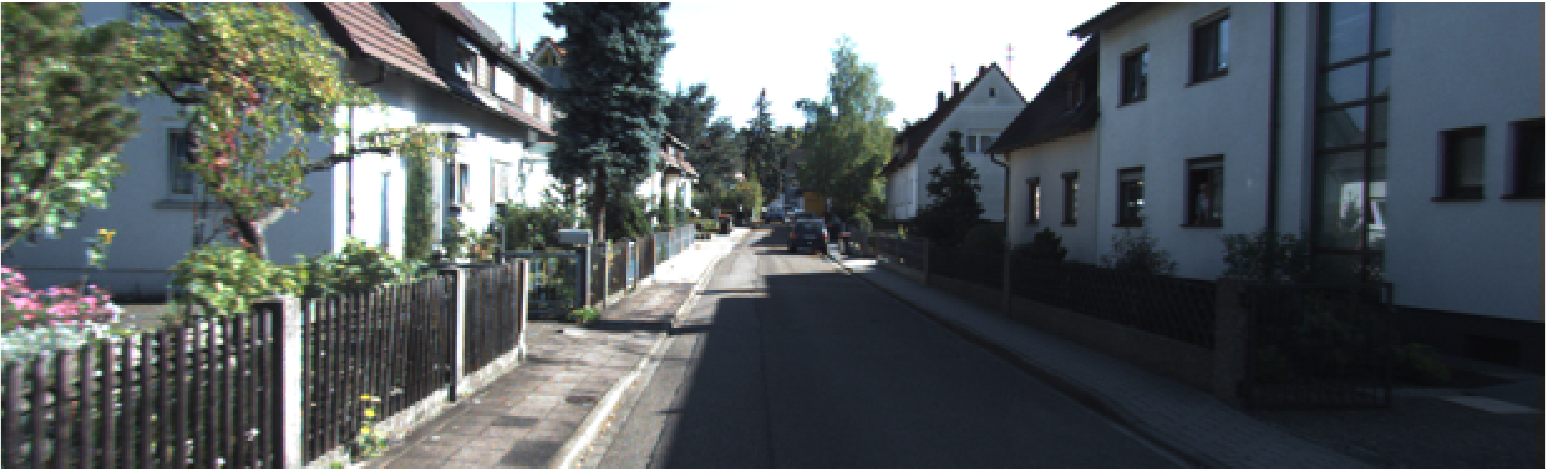} & 
	\hspace{-0.0cm}\includegraphics[width=0.305\linewidth]{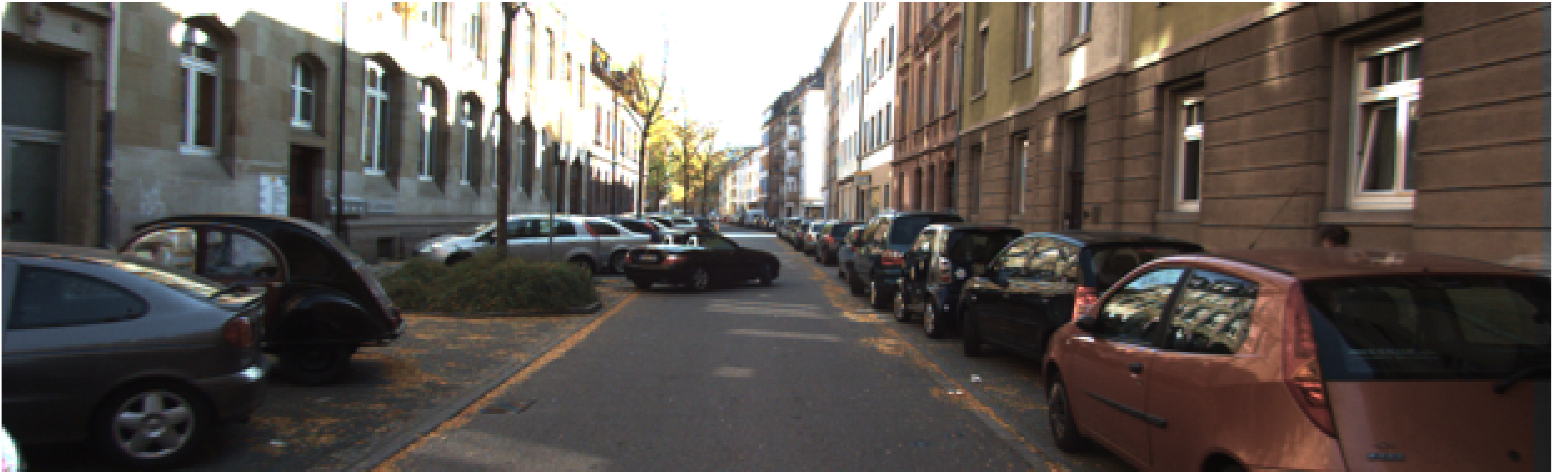} & 
	\hspace{-0.0cm}\includegraphics[width=0.305\linewidth]{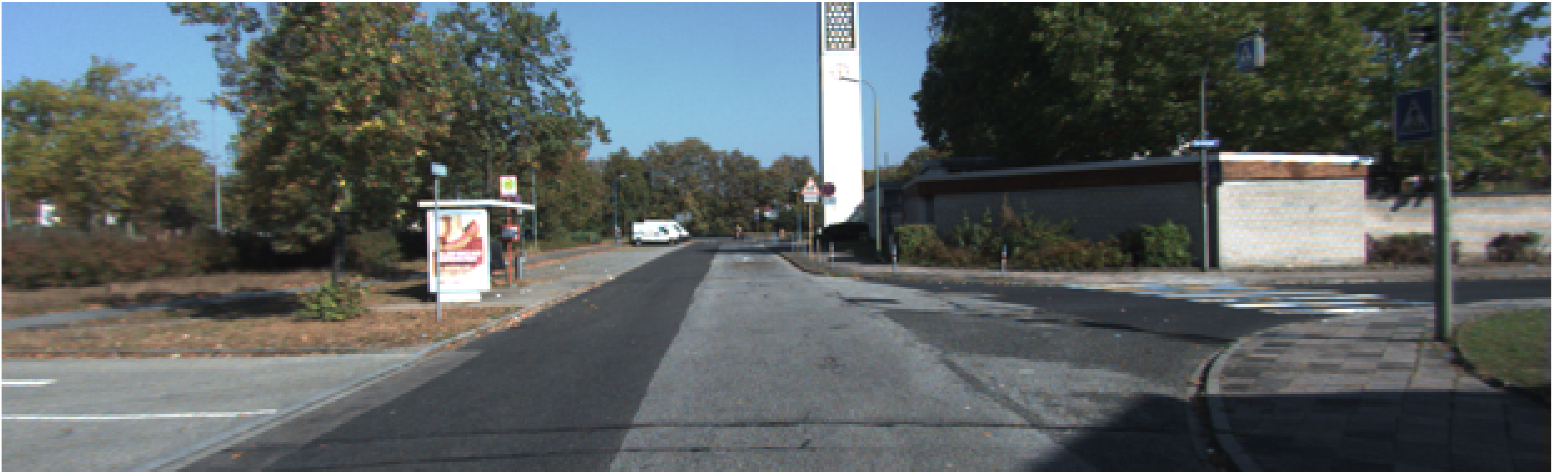}\\
	\rotatebox{90}{\parbox{0.9cm}{\centering GT\\ Semantic}}&\hspace{-0.0cm}\includegraphics[width=0.305\linewidth]{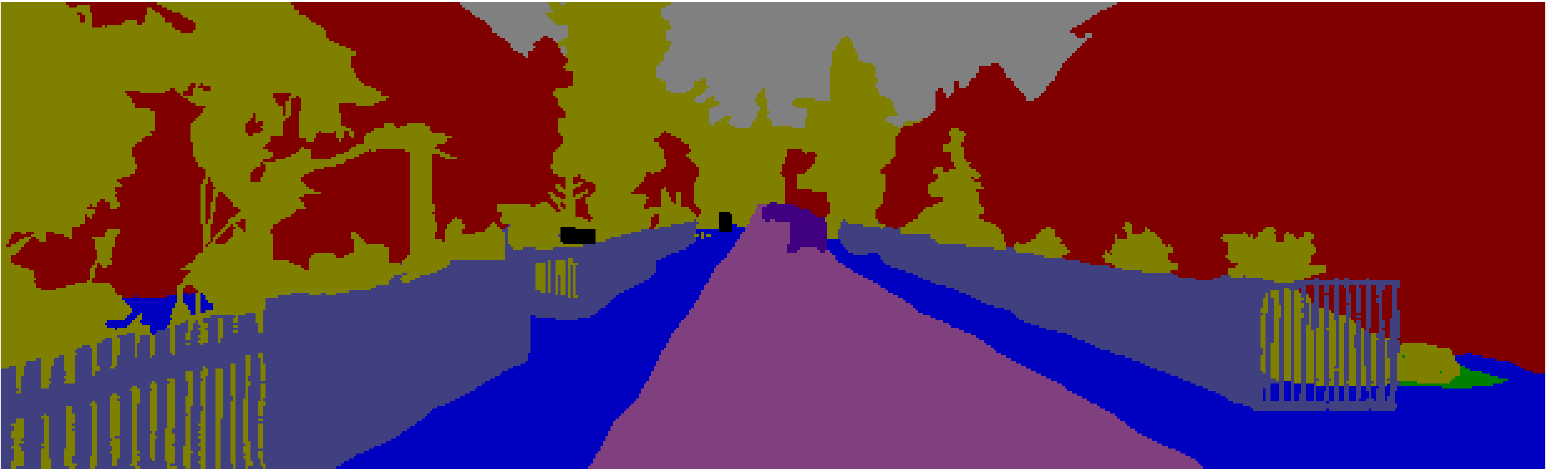} & 
	\hspace{-0.0cm}\includegraphics[width=0.305\linewidth]{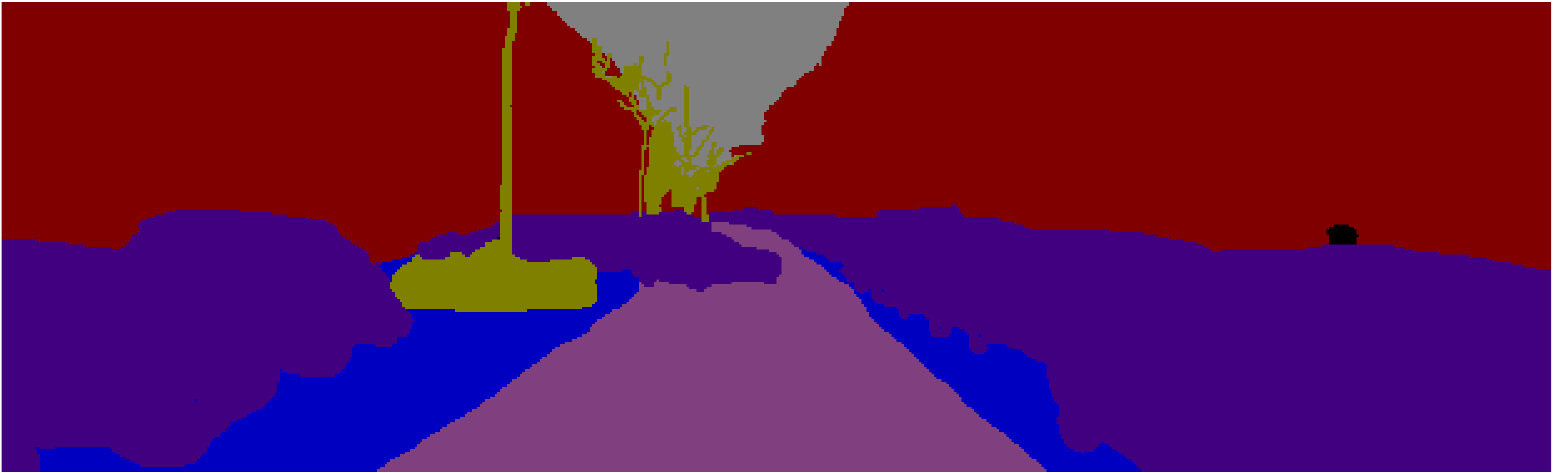} & 
	\hspace{-0.0cm}\includegraphics[width=0.305\linewidth]{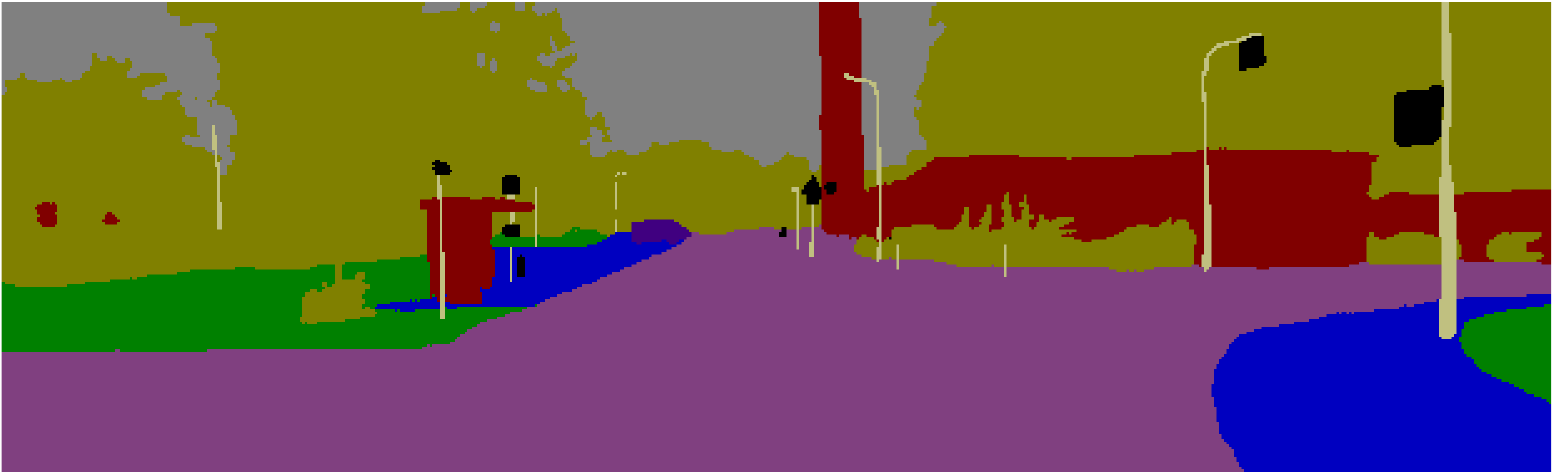}\\
	\rotatebox{90}{\parbox{0.9cm}{\centering GT\\ Disparity}}&\hspace{-0.0cm}\includegraphics[width=0.305\linewidth]{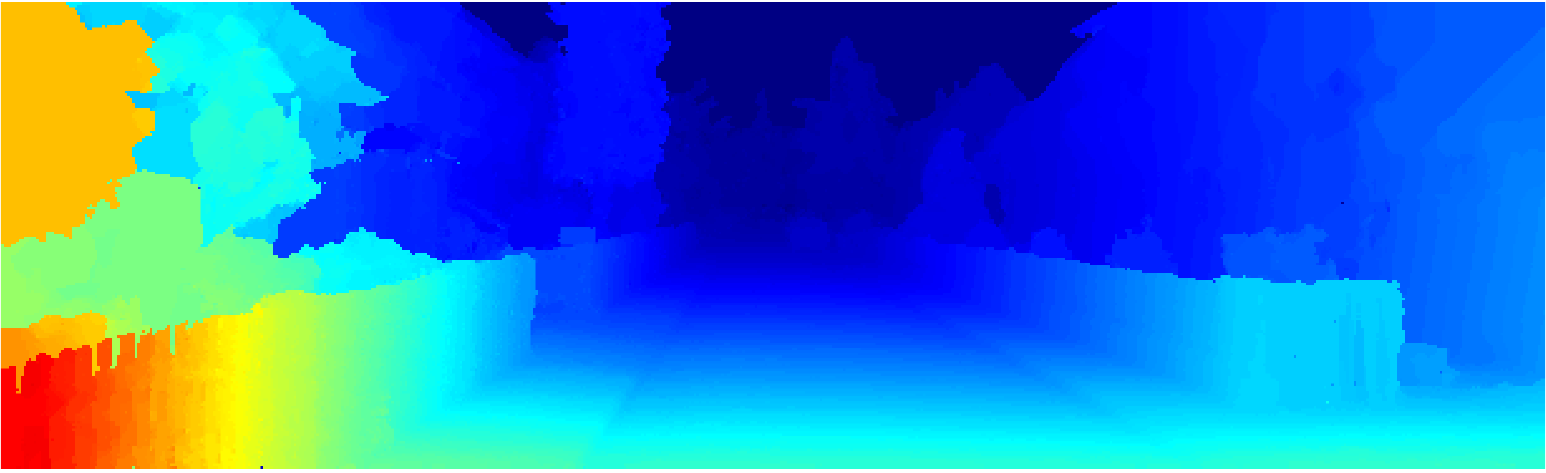} & 
	\hspace{-0.0cm}\includegraphics[width=0.305\linewidth]{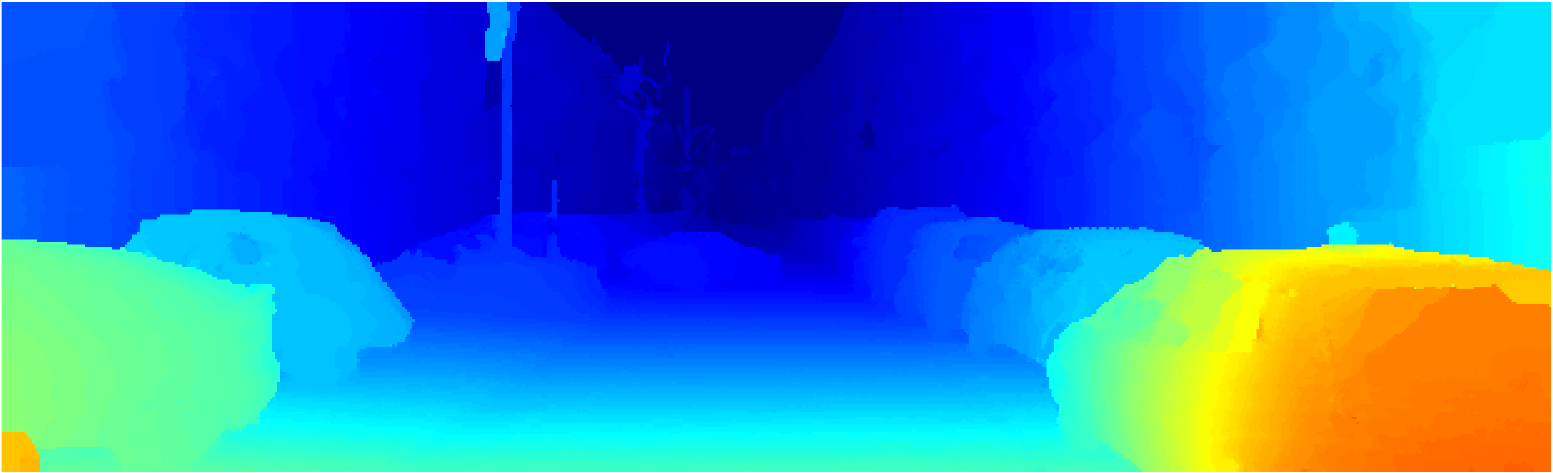} & 
	\hspace{-0.0cm}\includegraphics[width=0.305\linewidth]{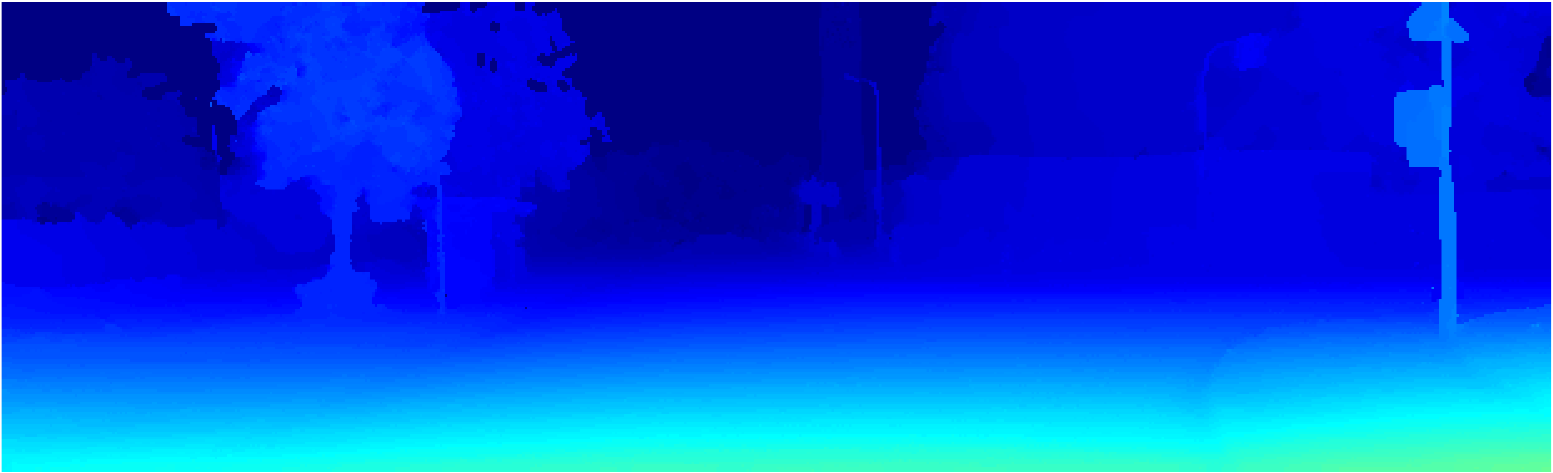}\\
	\rotatebox{90}{\parbox{0.9cm}{\centering Low-Res\\ Disparity}}&\hspace{-0.0cm}\includegraphics[width=0.305\linewidth]{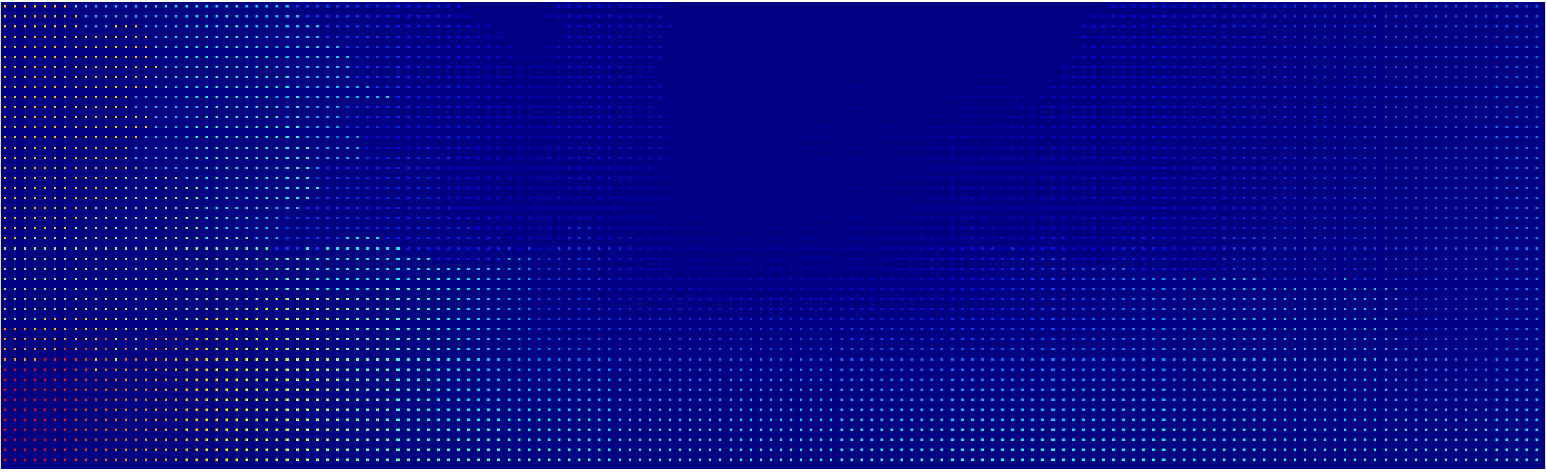} & 
	\hspace{-0.0cm}\includegraphics[width=0.305\linewidth]{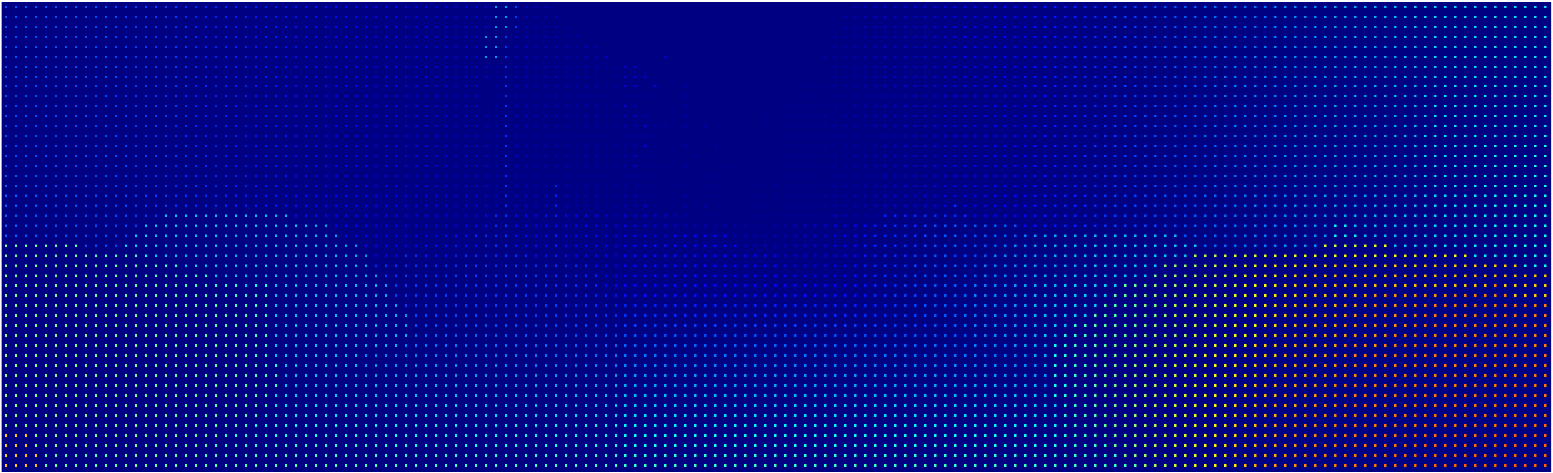} & 
	\hspace{-0.0cm}\includegraphics[width=0.305\linewidth]{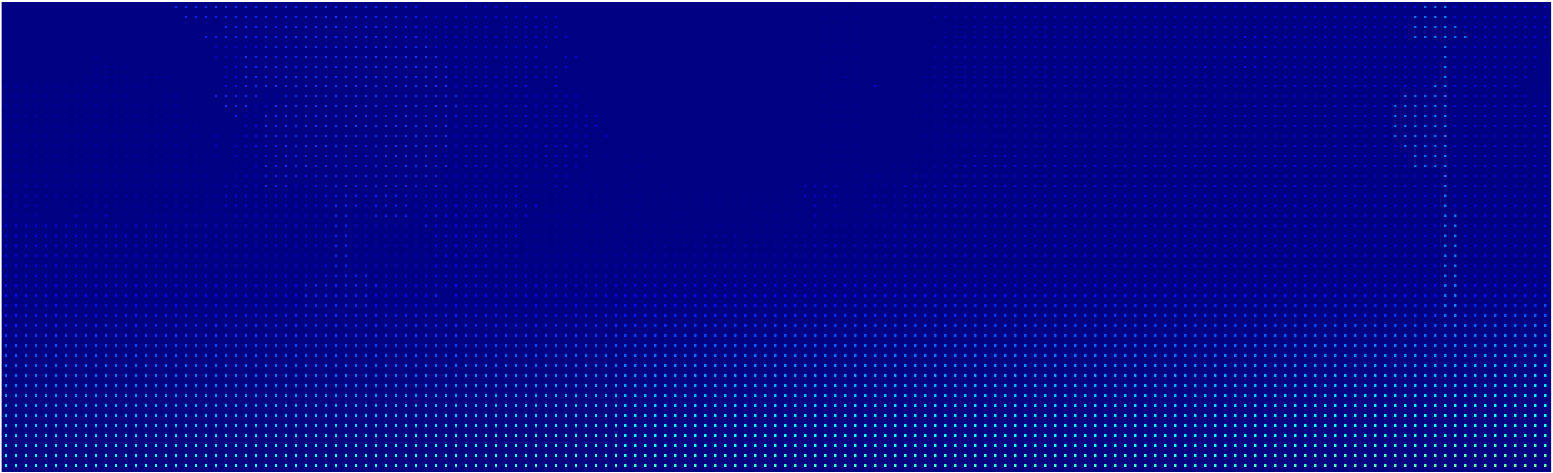}\\
	\rotatebox{90}{\parbox{0.9cm}{\centering Our\\ Disparity}}&\hspace{-0.0cm}\includegraphics[width=0.305\linewidth]{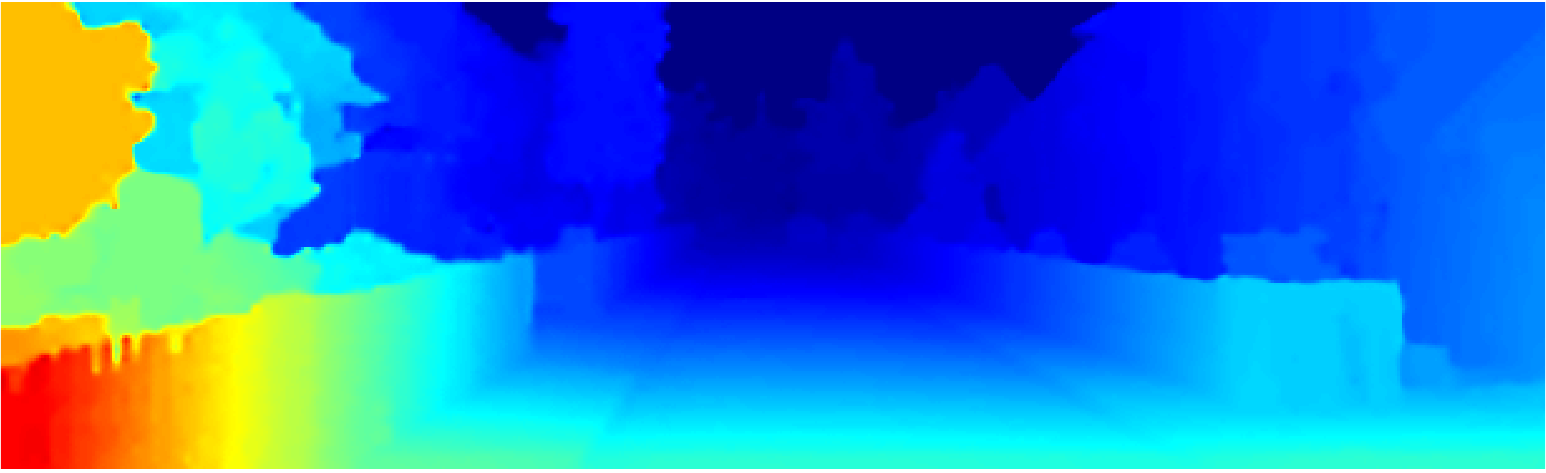} & 
	\hspace{-0.0cm}\includegraphics[width=0.305\linewidth]{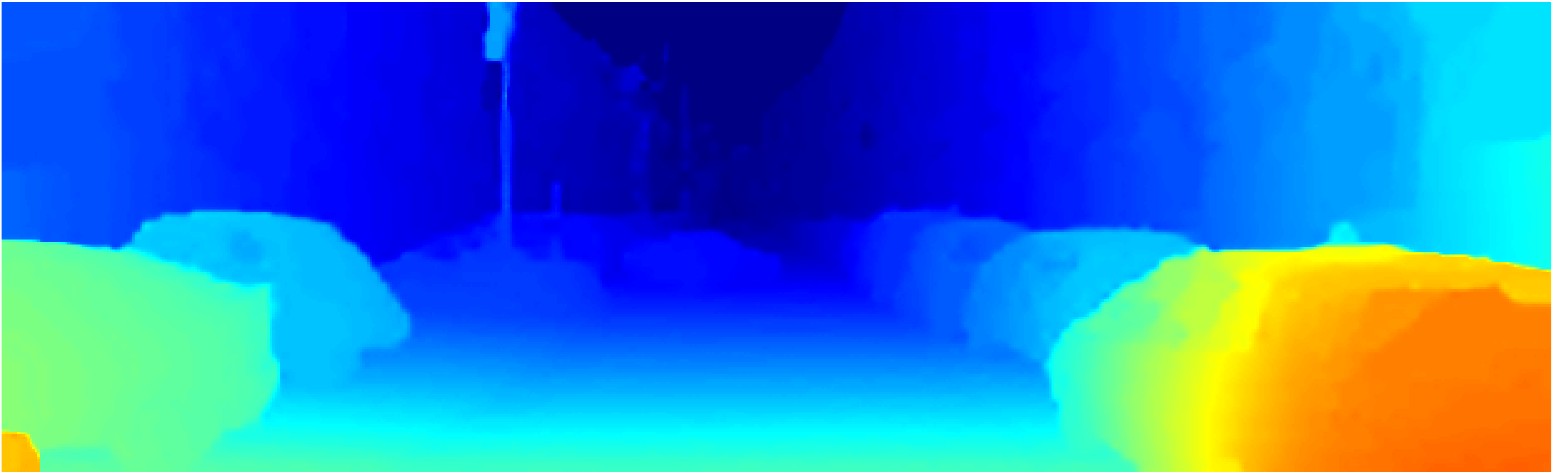} & 
	\hspace{-0.0cm}\includegraphics[width=0.305\linewidth]{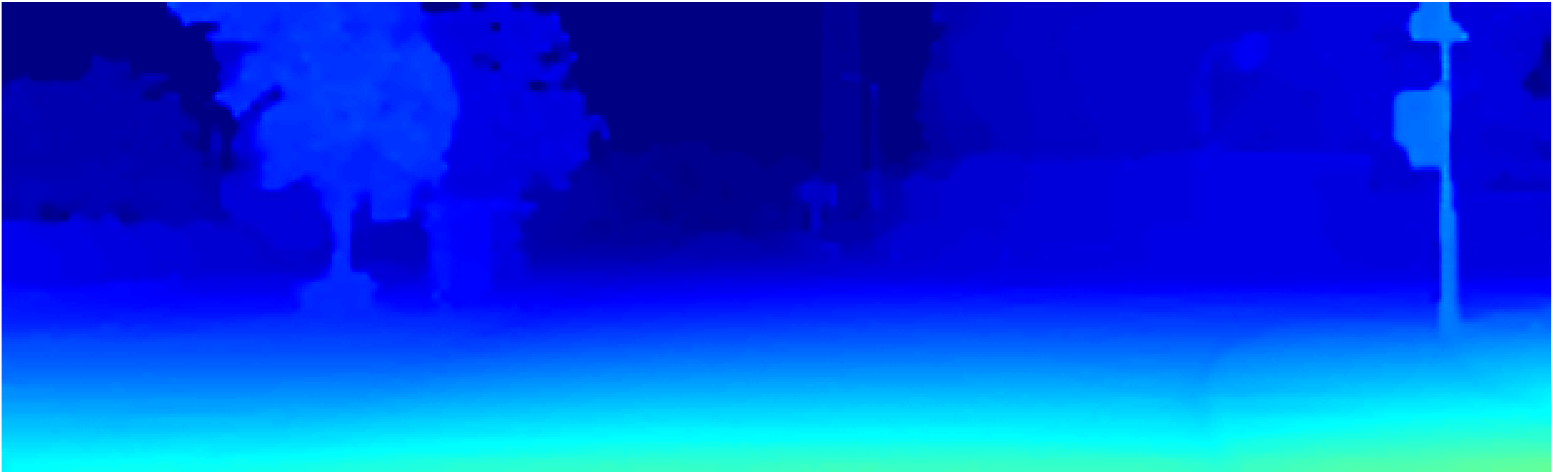}\\
	&\multicolumn{3}{c}{\hspace{-0.1cm}\includegraphics[width=0.8\linewidth]{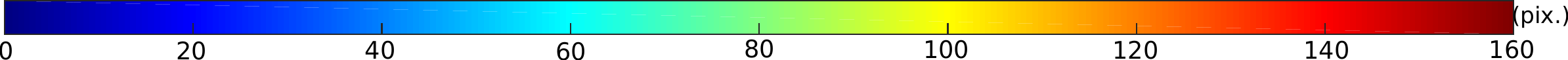}}\\	
	\rotatebox{90}{\parbox{0.9cm}{\centering Noisy\\ Semantic}}&\hspace{-0.0cm}\includegraphics[width=0.305\linewidth]{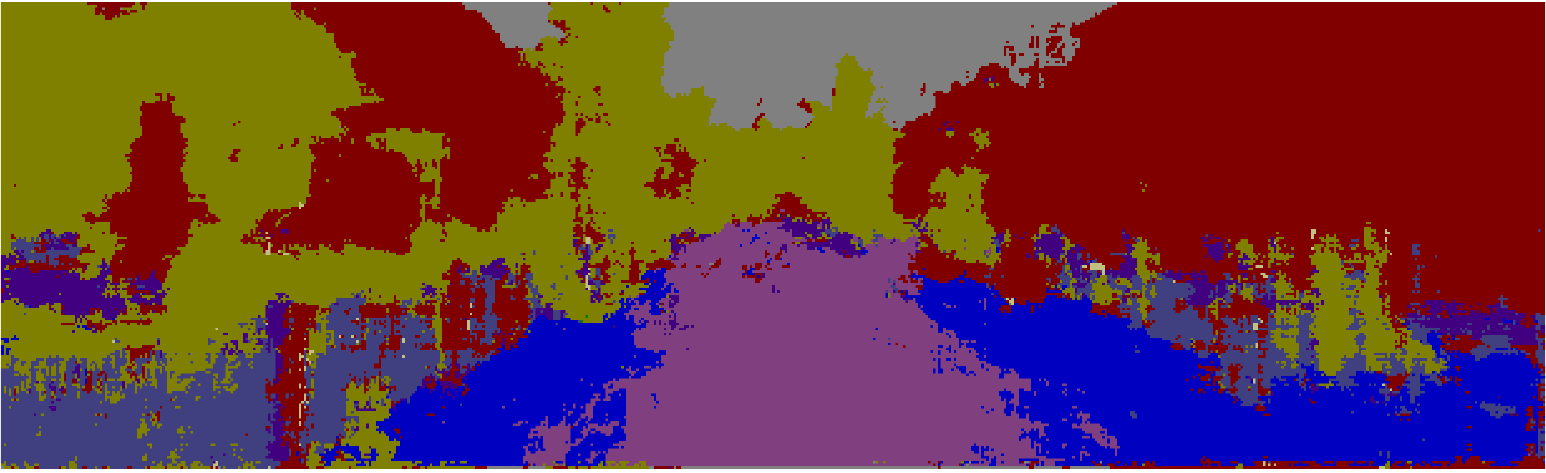} & 
	\hspace{-0.0cm}\includegraphics[width=0.305\linewidth]{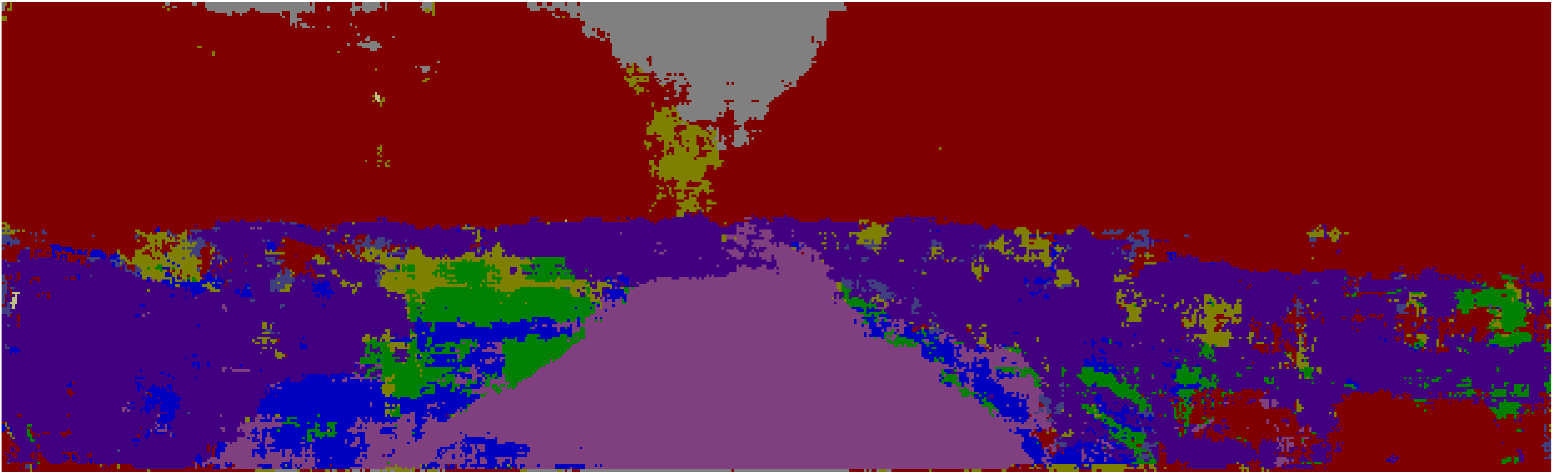} & 
	\hspace{-0.0cm}\includegraphics[width=0.305\linewidth]{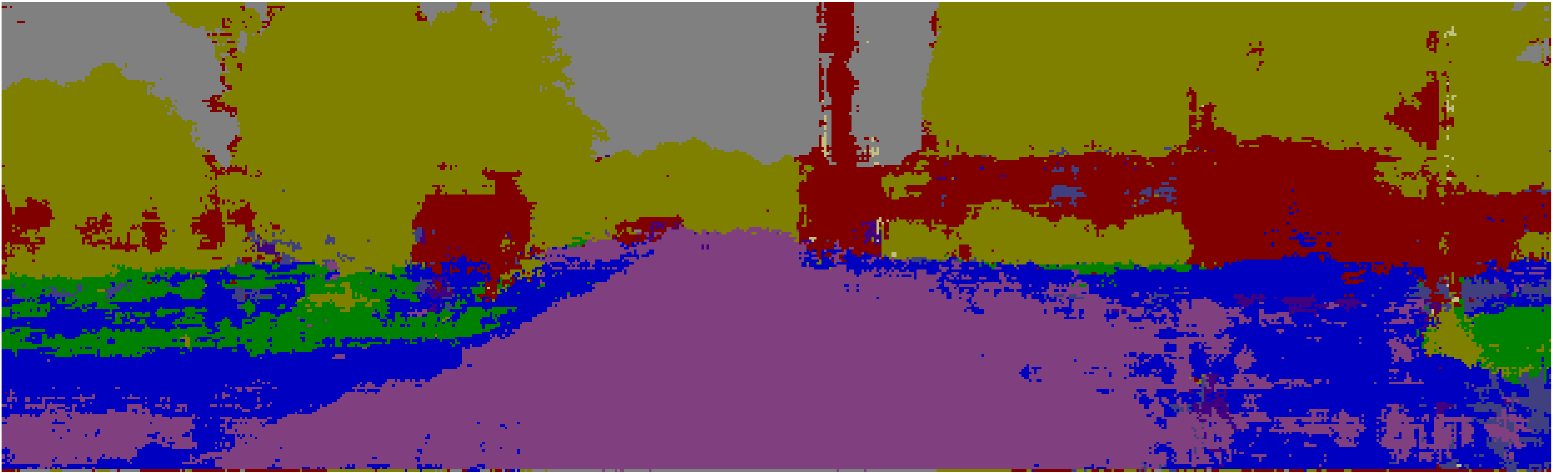}\\
	\rotatebox{90}{\parbox{0.9cm}{\centering Our\\ Semantic}}&\hspace{-0.0cm}\includegraphics[width=0.305\linewidth]{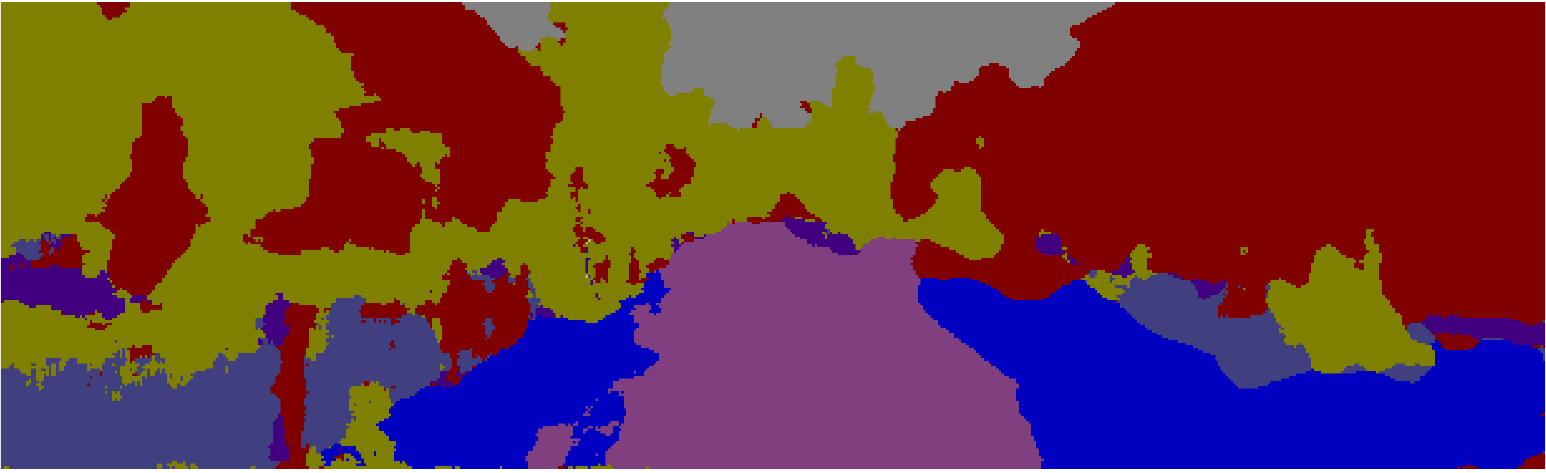} & 
	\hspace{-0.0cm}\includegraphics[width=0.305\linewidth]{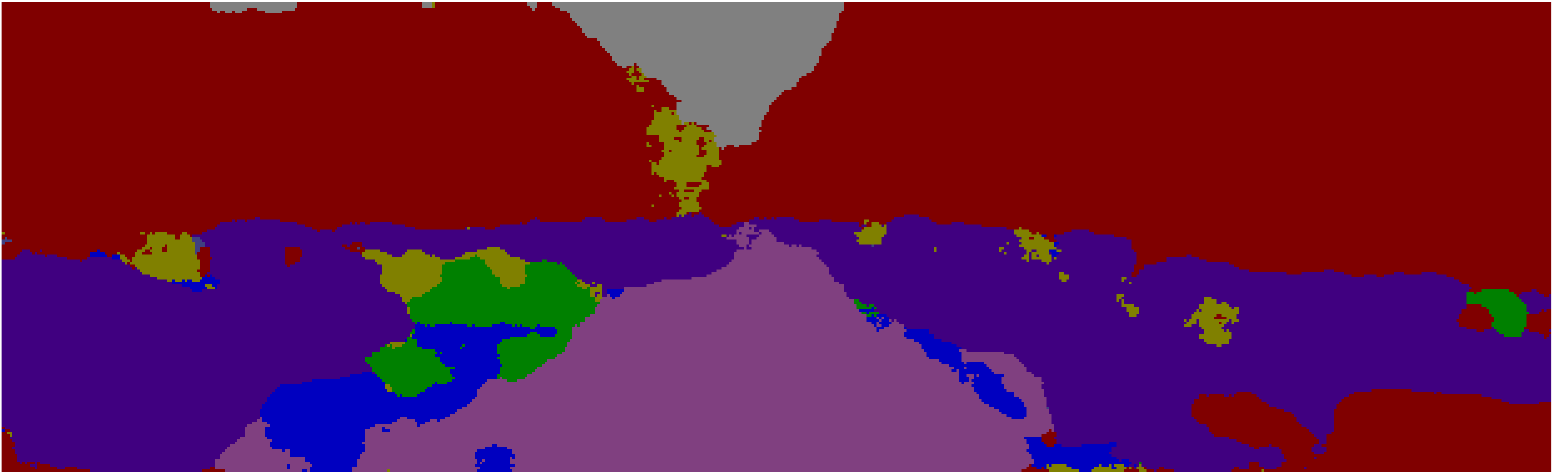} & 
	\hspace{-0.0cm}\includegraphics[width=0.305\linewidth]{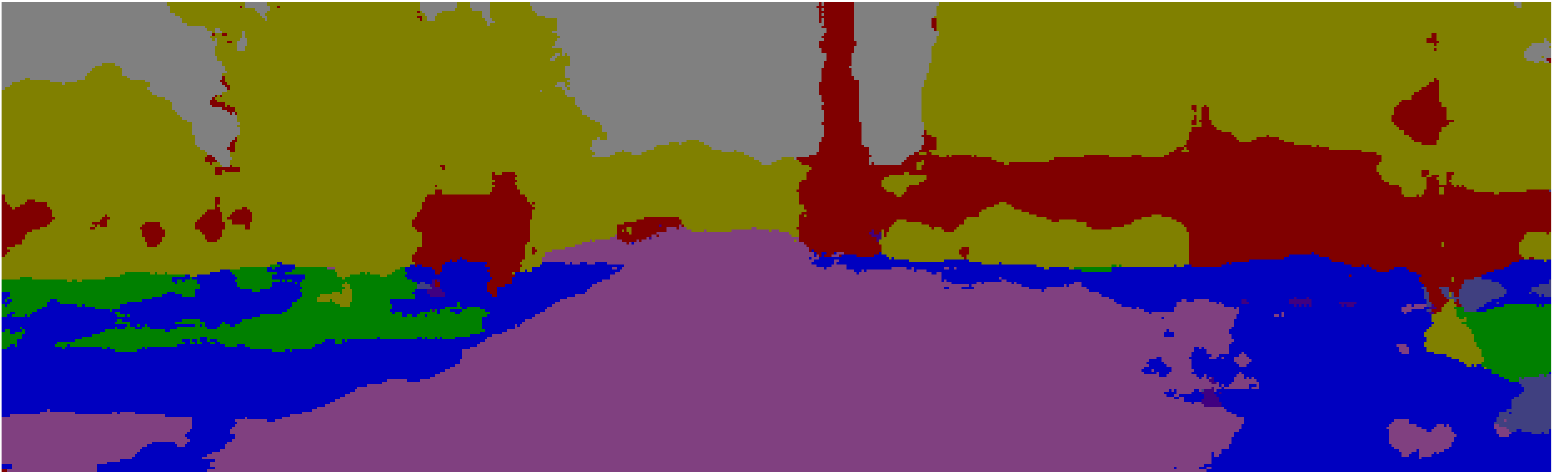}\\
	&\multicolumn{3}{c}{\includegraphics[width=0.8\linewidth]{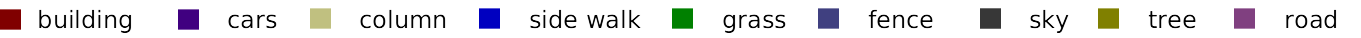}}
\end{tabular}
\end{small}
\vspace{0.05cm}
\caption{{\bf Qualitative results on the KITTI dataset for a downsampling factor of 4.} For the disparity values, red denotes large values, \ie, points close to the camera, and blue denotes small disparity values, \ie, points far from the camera. {\bf From top to bottom:} RGB image, ground-truth semantics, ground-truth dense disparity map, the 6.25\% observations mapped on the grid of the high-resolution image, our estimated high-resolution disparity map, noisy semantics predicted by the method of~\cite{Gould2012JMLR}, our estimated semantics. Our method yields accurate high-resolution depth maps, and improves the semantic labeling results. Best viewed in color.}
\label{fig:kittires}
\vspace{-0.3cm}
\end{figure*}

In Table~\ref{Tab:oursComp}, we compare the results of our approach trained with low-resolution depth maps with those obtained by training with high-resolution ones. Note that training on low-resolution depth maps comes at very little loss in accuracy. Importantly, to the best of our knowledge, this realistic scenario has never been demonstrated in the past.

Since our method also denoises the semantic labels, we report the average per pixel and average per class accuracies of semantic labeling. As a first experiment, we used the multi-class classifier of~\cite{Gould2012JMLR} to generate noisy labels, which yields average per pixel and per class accuracies of 77.26\% and 60.71\%, respectively. After denoising with our approach, these accuracies increased to 81.19\% and 63.98\%. This is comparable to the state-of-the-art MRF model of~\cite{Krah11}, which yields accuracies of 81.82\%~and~64.31\%. As a second experiment, we then started from this slightly more accurate result. After our denoising process, the accuracies were further improved to 82.32\% and 64.81\%. We observed that these results are not sensitive to the sparsity of the depth map. In Fig.~\ref{fig:kittires}, we provide a qualitative evaluation of our results.

\begin{figure*}[t!]
\vspace{-0.1cm}
\begin{small}
\begin{tabular}{ccccccccc}
%\hspace{-2.5cm}{\bf Image} &  \hspace{-2.5cm}{\bf Sparse Depth} & & \hspace{-2.5cm}{\bf Noisy Semantics} & \hspace{-2.5cm}{\bf semantic result}\hspace{-2.5cm}{\bf depth result} \\
\raisebox{0.0em}{\rotatebox{90}{{ RGB Image}}}&\hspace{-0.0cm}\includegraphics[width=0.118\linewidth]{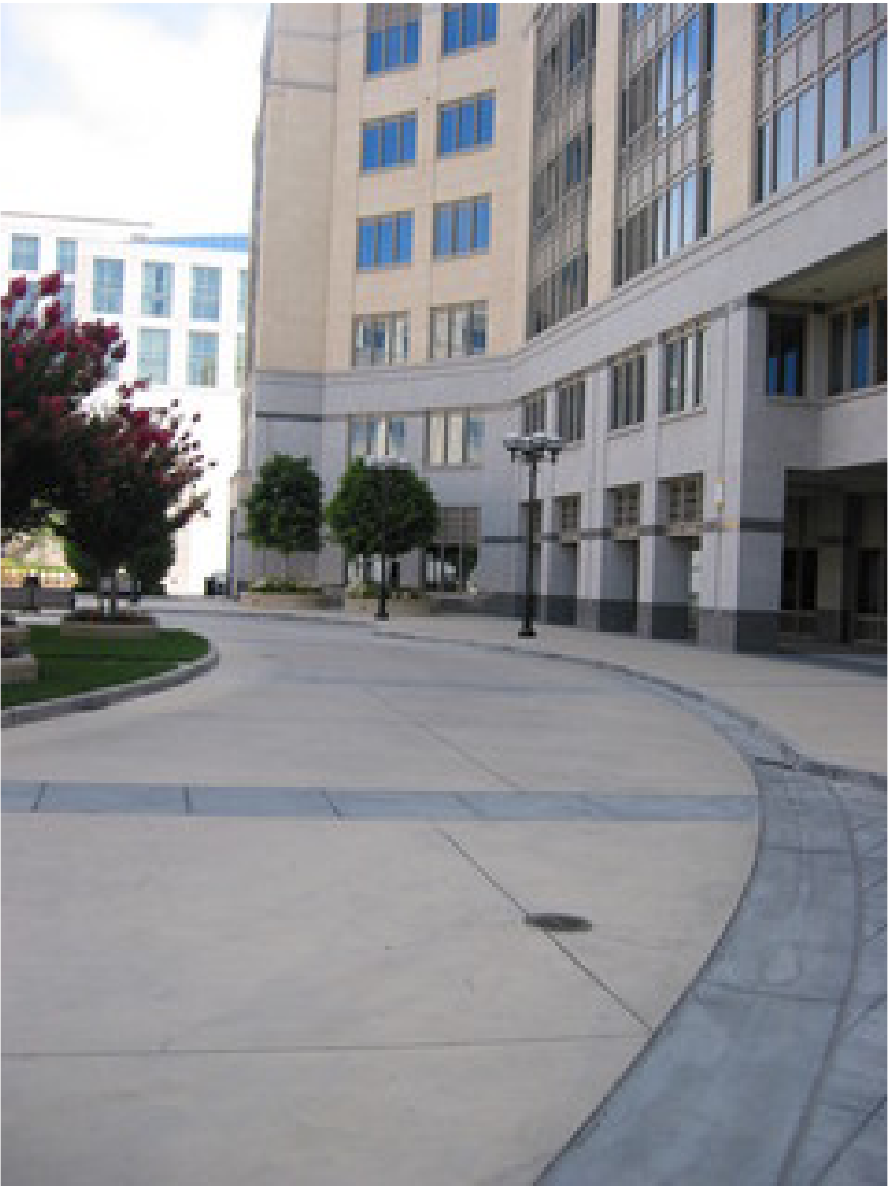} & 
\hspace{-0.0cm}\includegraphics[width=0.118\linewidth]{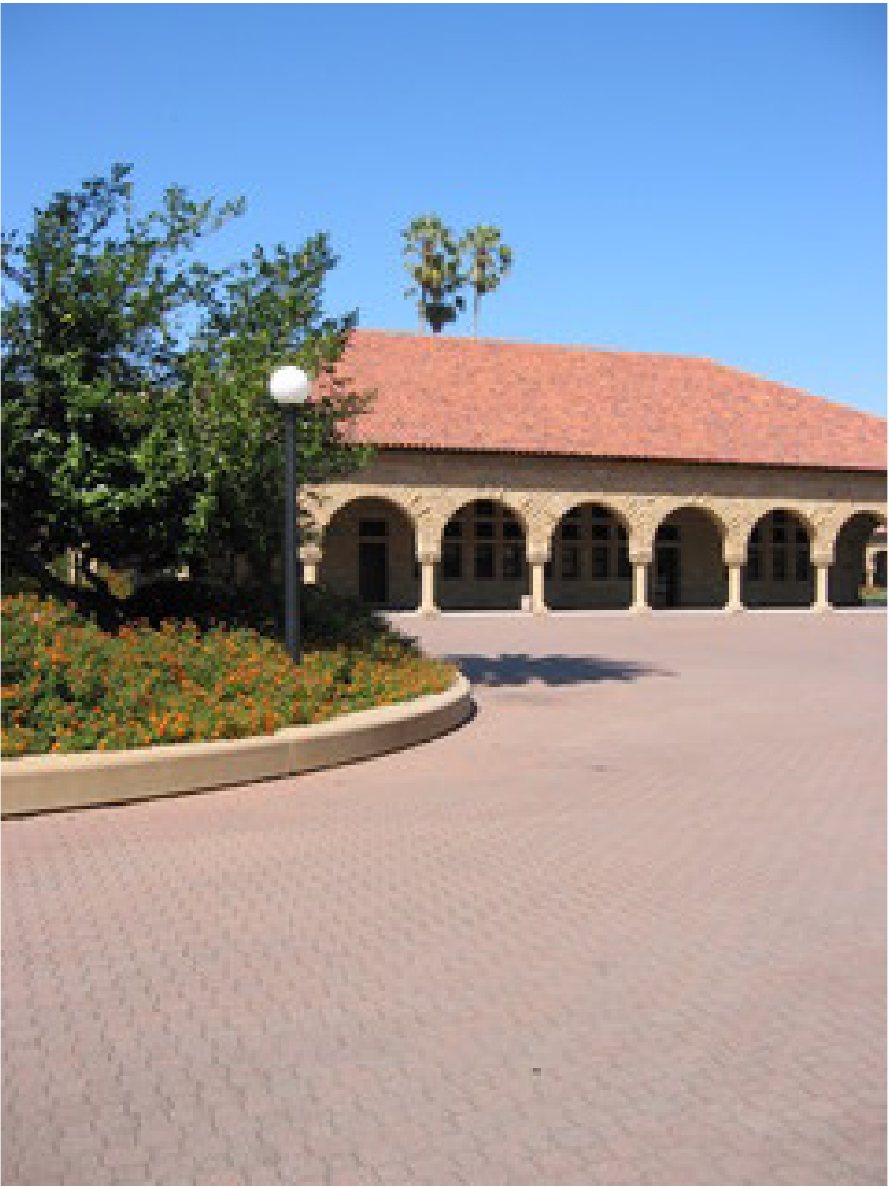}&
\hspace{-0.0cm}\includegraphics[width=0.118\linewidth]{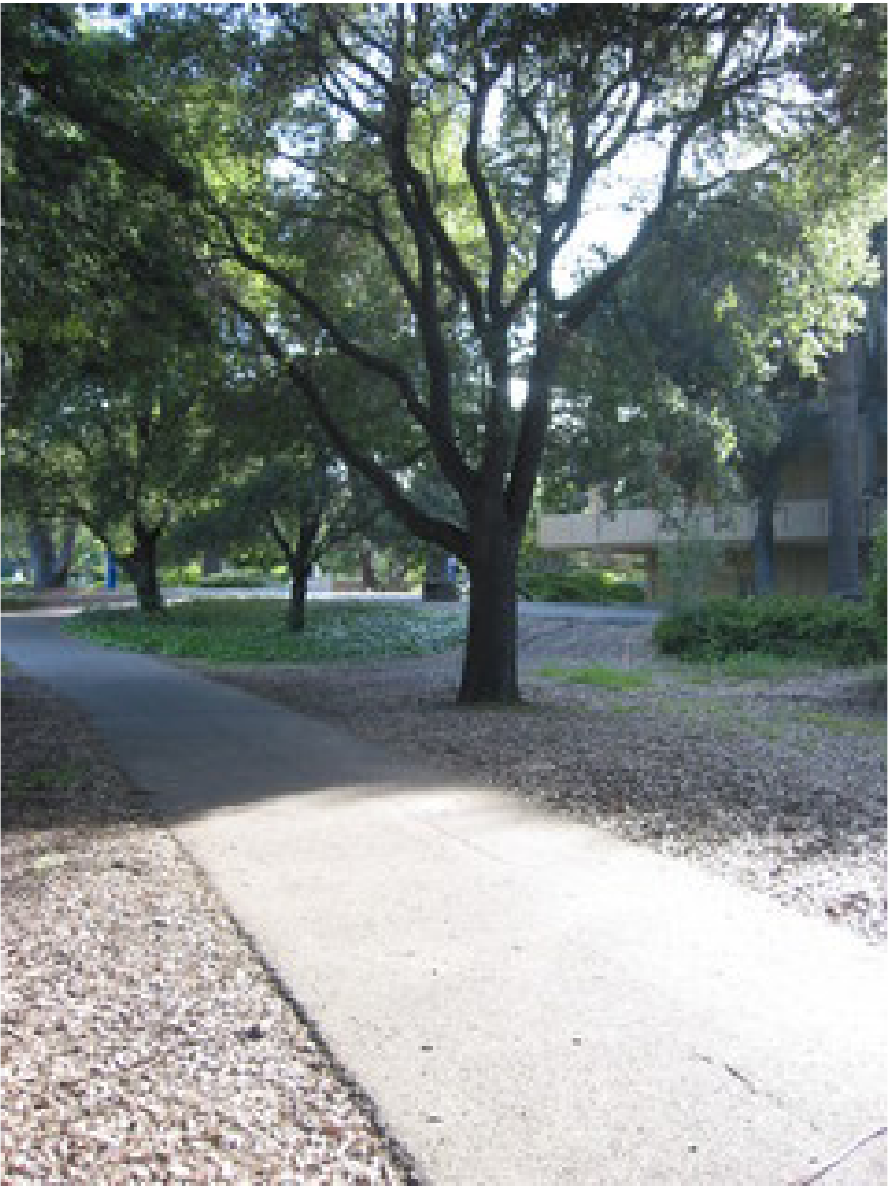} & 
\hspace{-0.0cm}\includegraphics[width=0.118\linewidth]{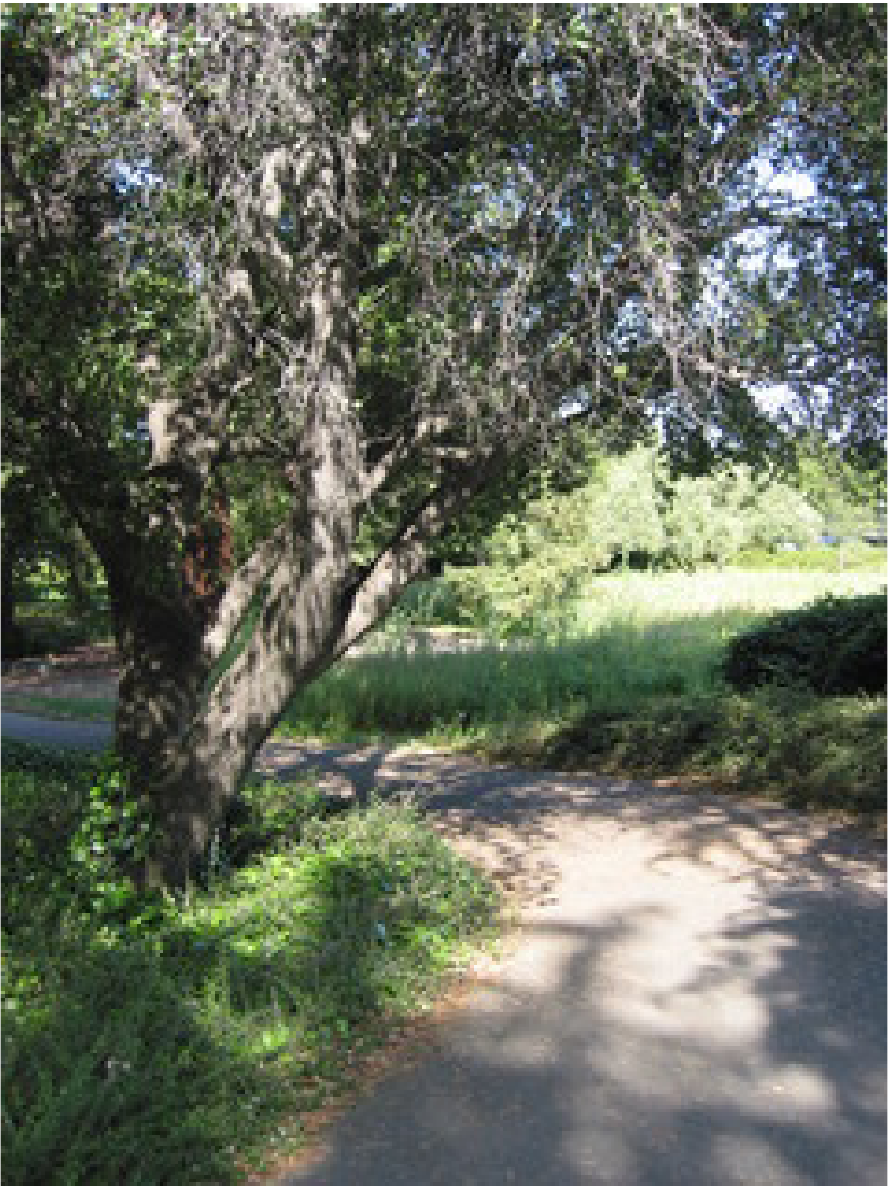}&
\hspace{-0.0cm}\includegraphics[width=0.118\linewidth]{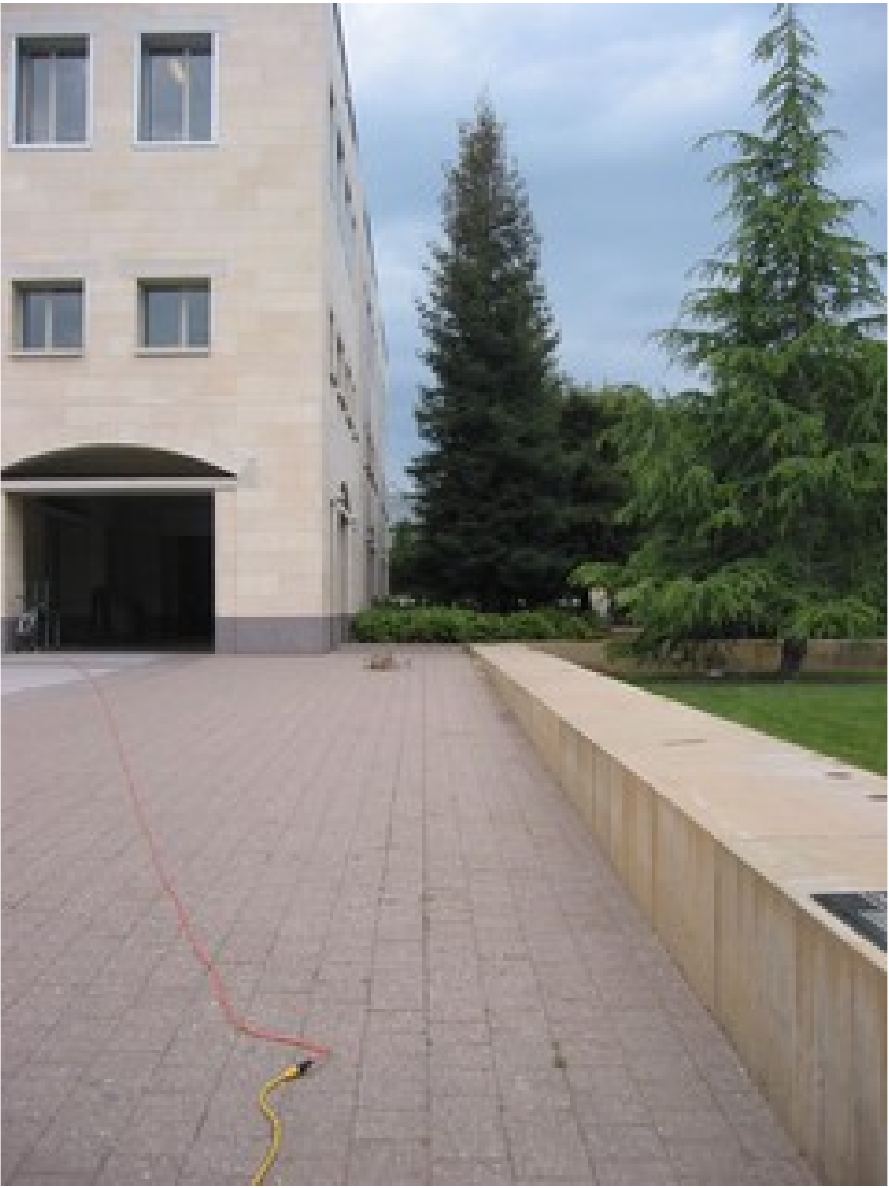} & 
\hspace{-0.0cm}\includegraphics[width=0.118\linewidth]{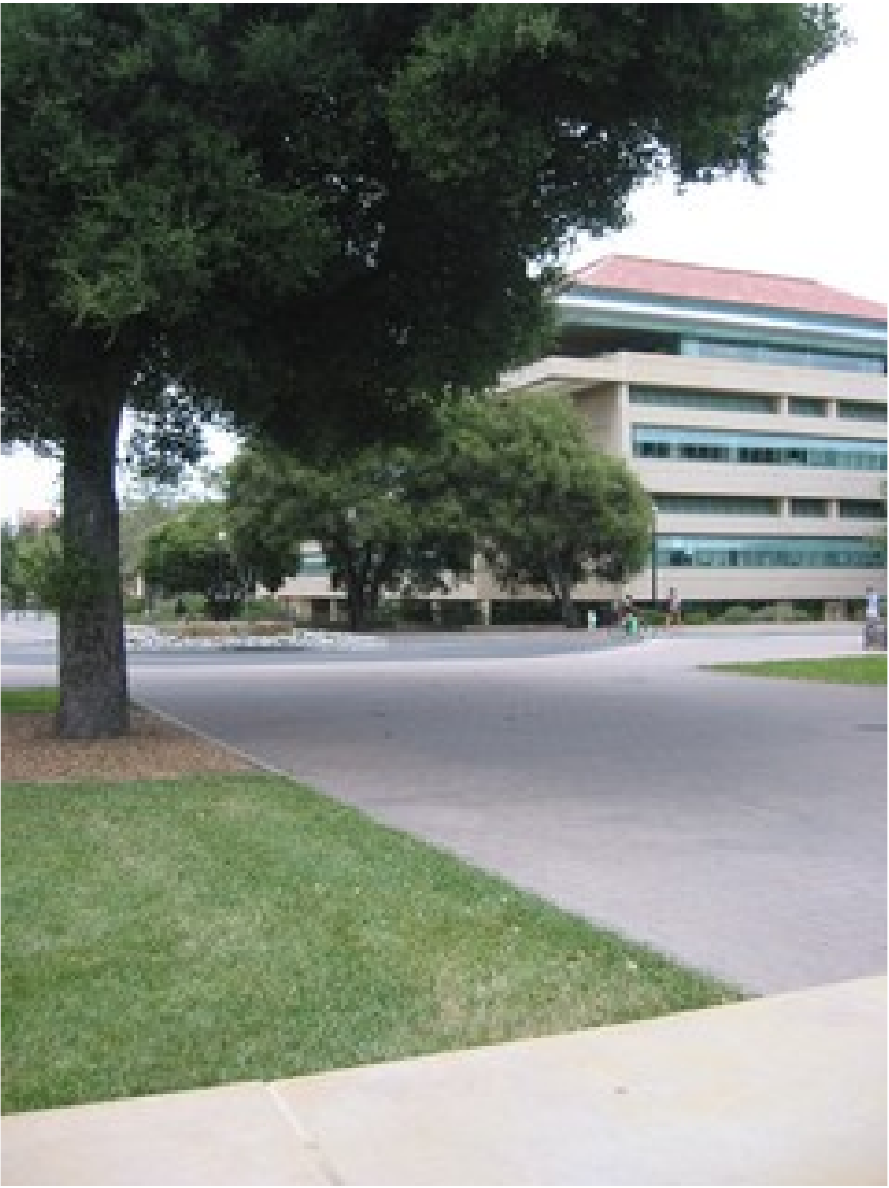} & 
\hspace{-0.0cm}\includegraphics[width=0.118\linewidth]{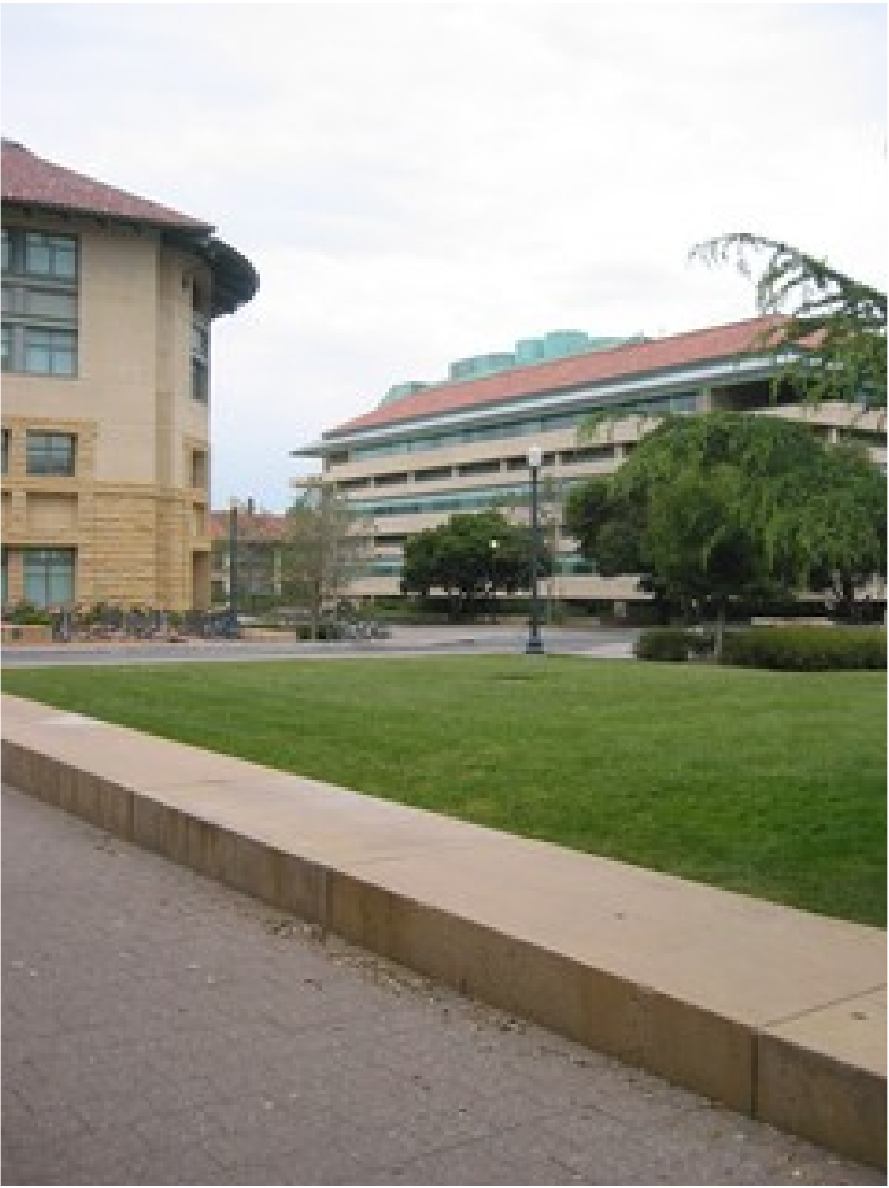} & 
\hspace{-0.0cm}\includegraphics[width=0.118\linewidth]{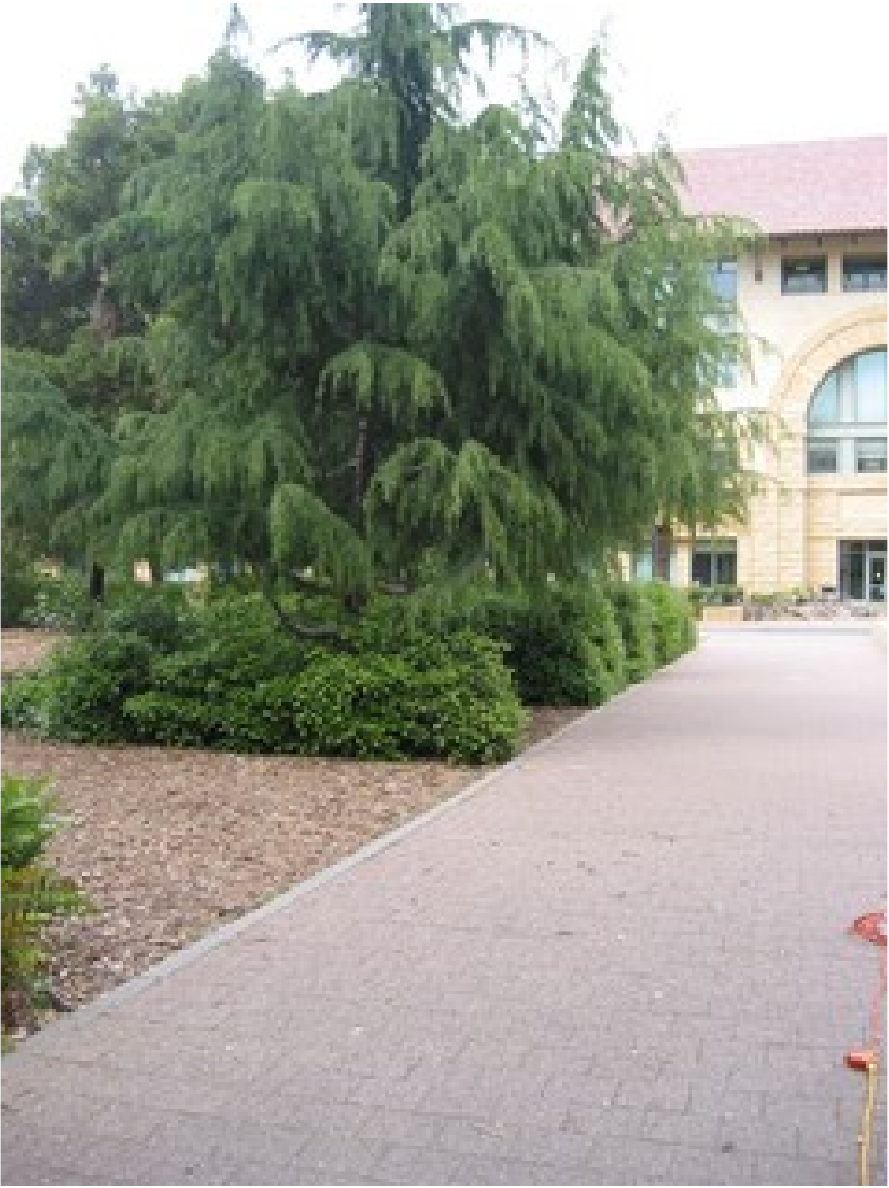}\\
\raisebox{-.0em}{\rotatebox{90}{\footnotesize{ GT Semantic}}}&\hspace{-0.0cm}\includegraphics[width=0.118\linewidth]{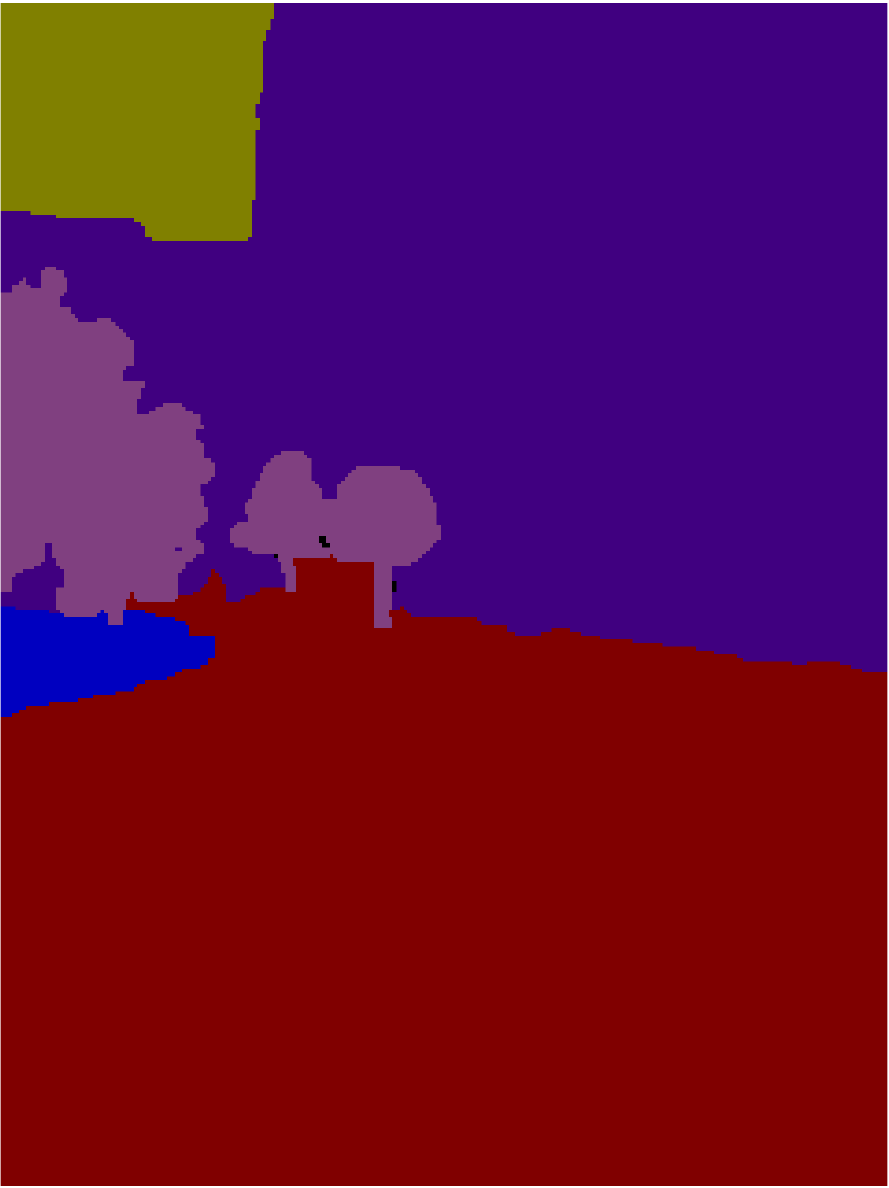} & 
\hspace{-0.0cm}\includegraphics[width=0.118\linewidth]{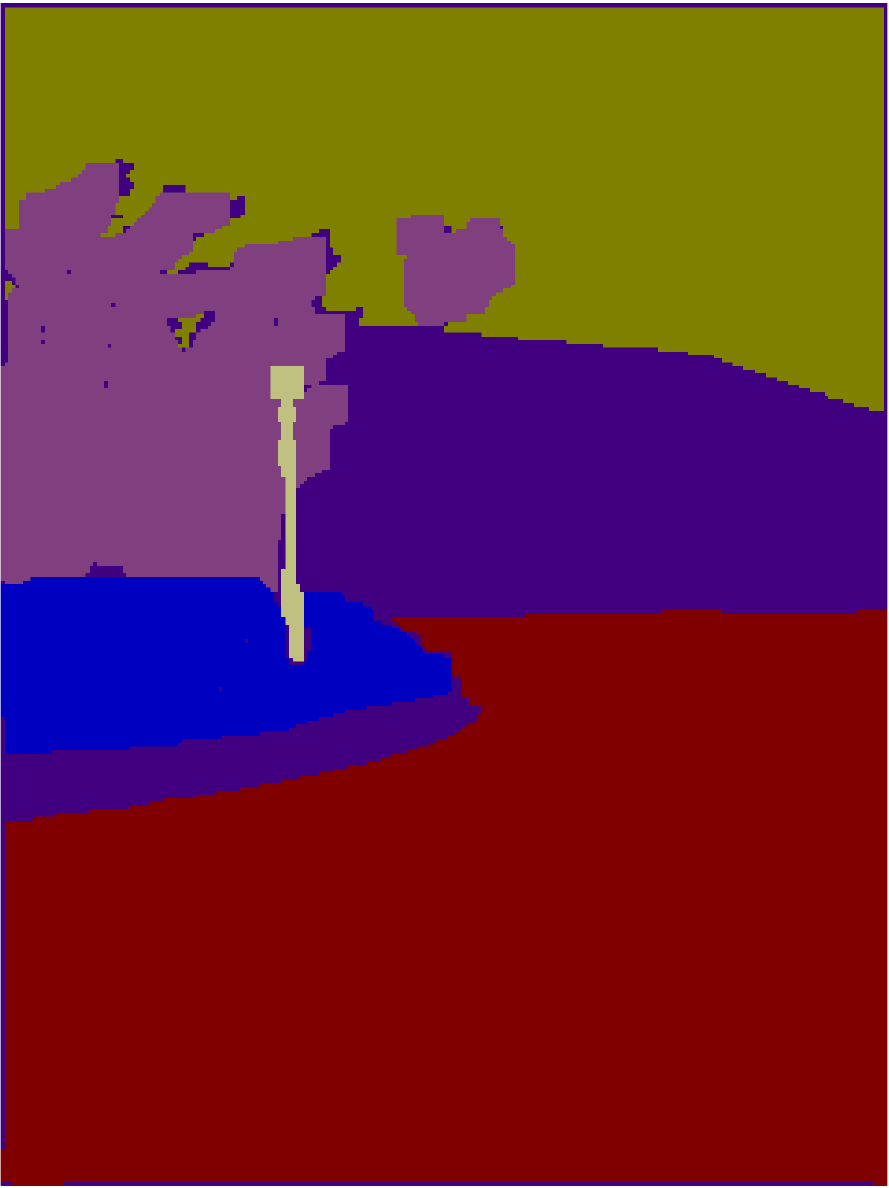}&
\hspace{-0.0cm}\includegraphics[width=0.118\linewidth]{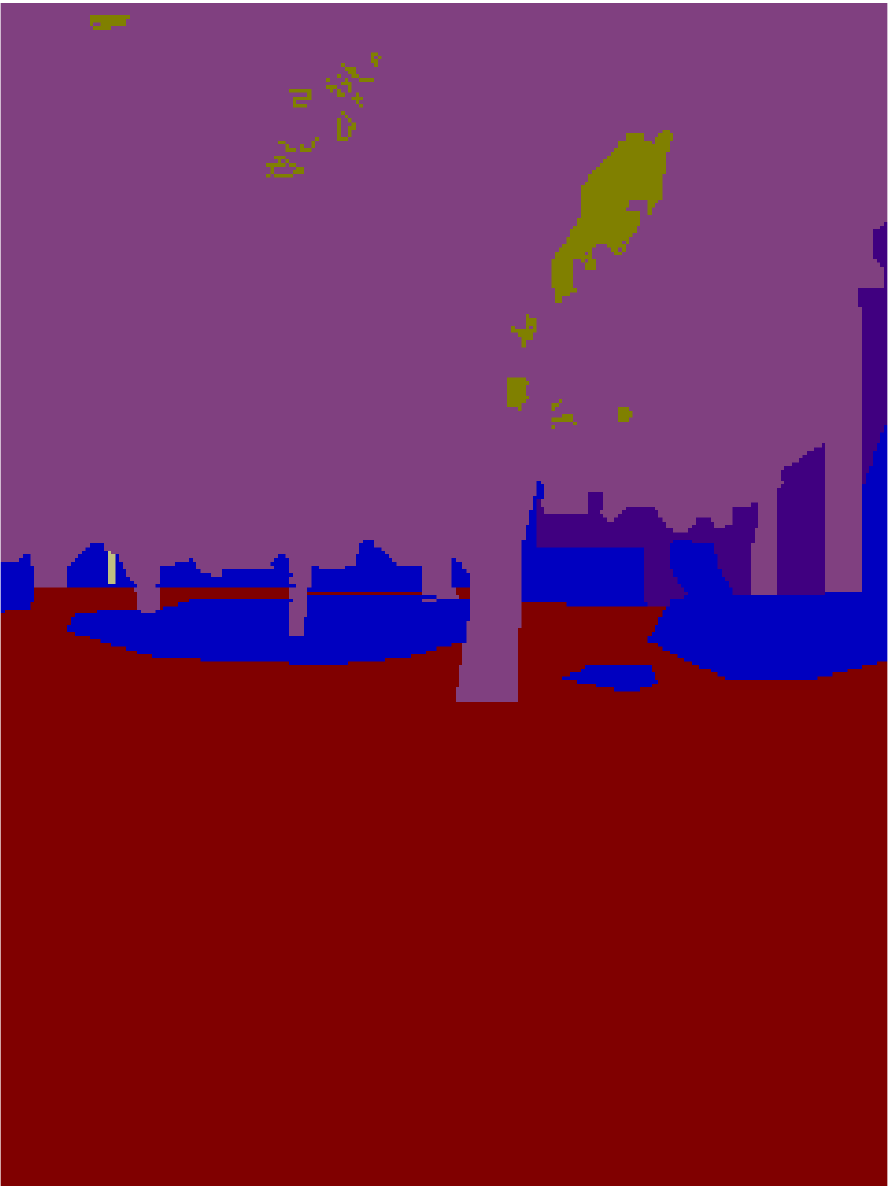} & 
\hspace{-0.0cm}\includegraphics[width=0.118\linewidth]{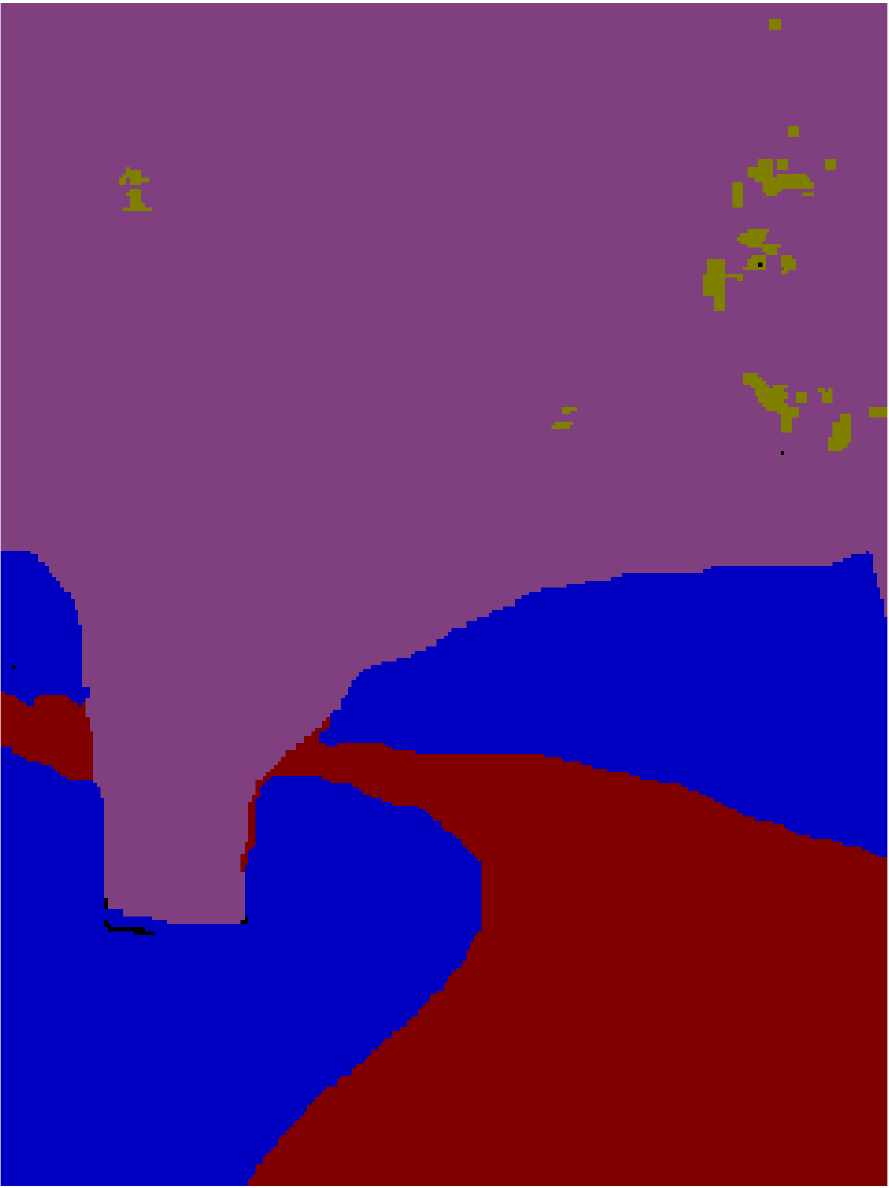}&
\hspace{-0.0cm}\includegraphics[width=0.118\linewidth]{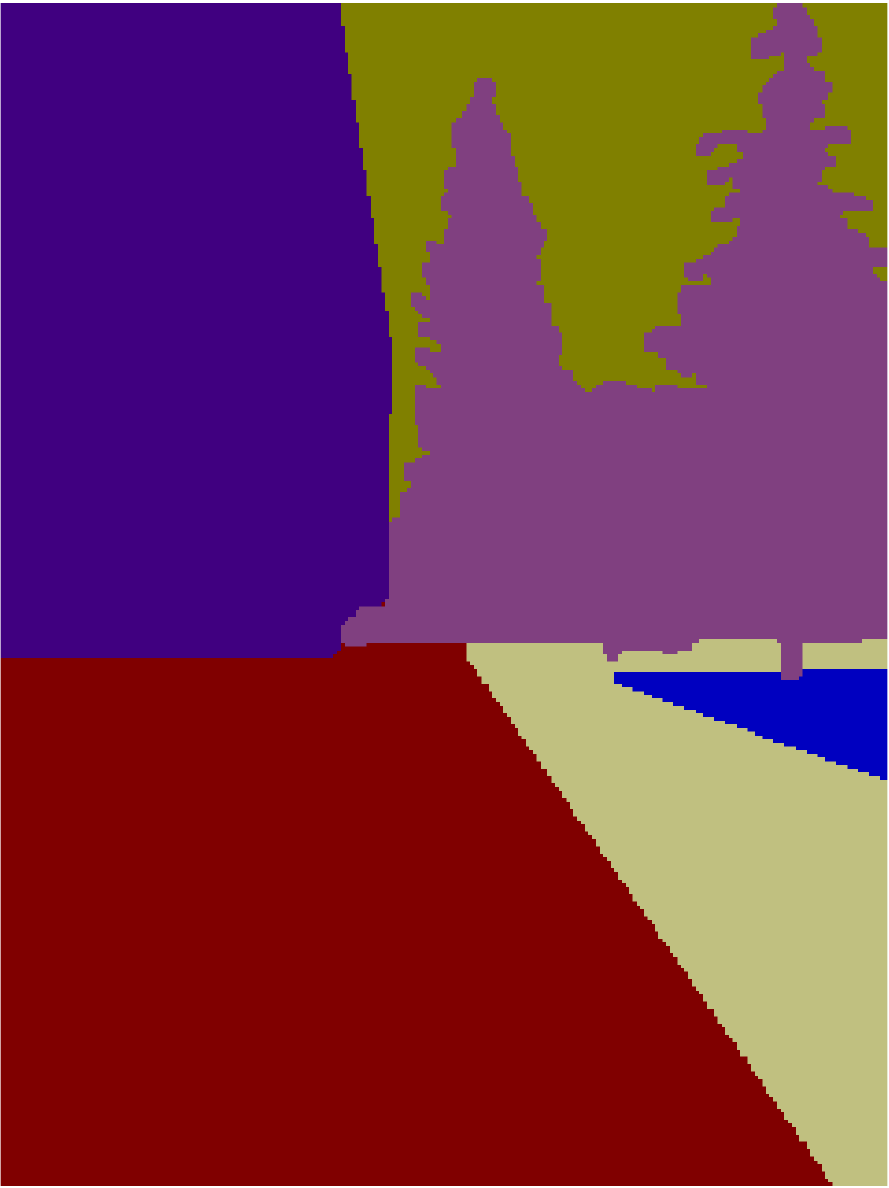} & 
\hspace{-0.0cm}\includegraphics[width=0.118\linewidth]{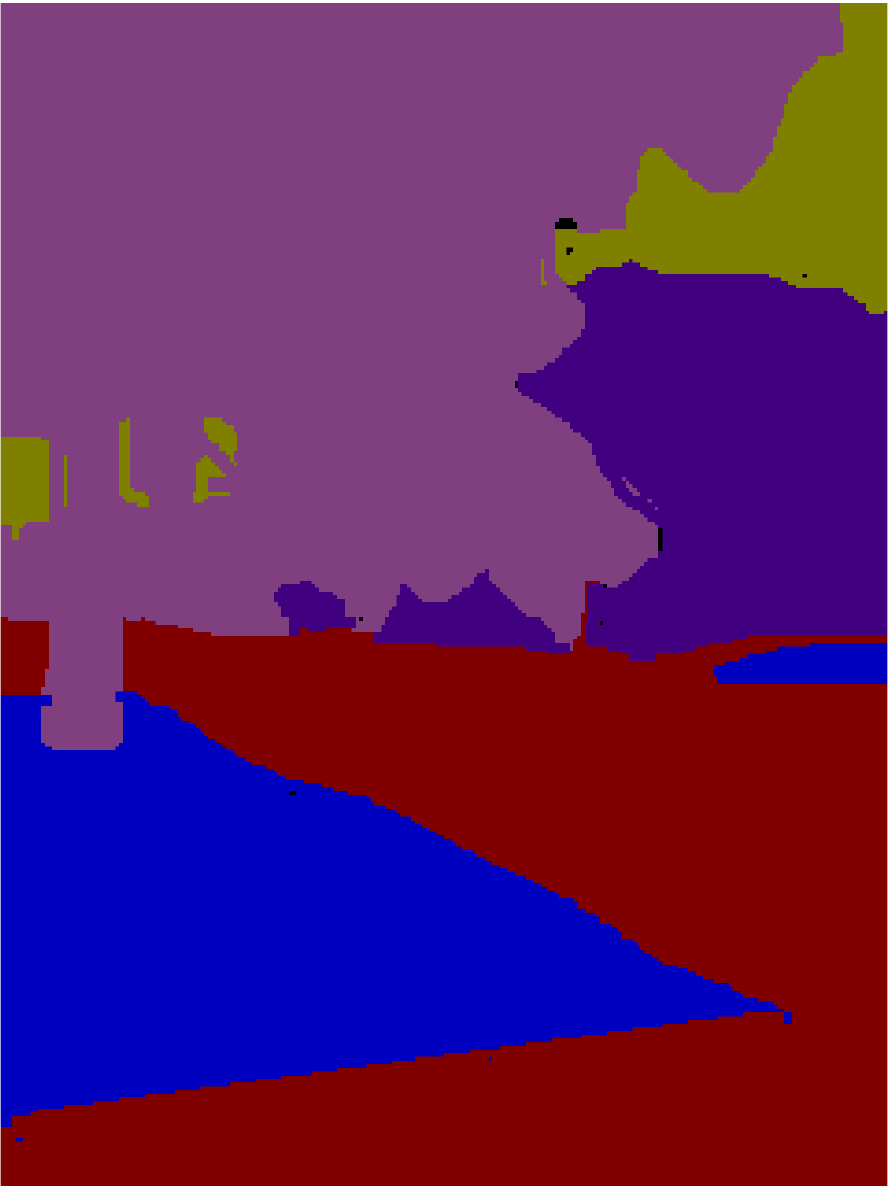} & 
\hspace{-0.0cm}\includegraphics[width=0.118\linewidth]{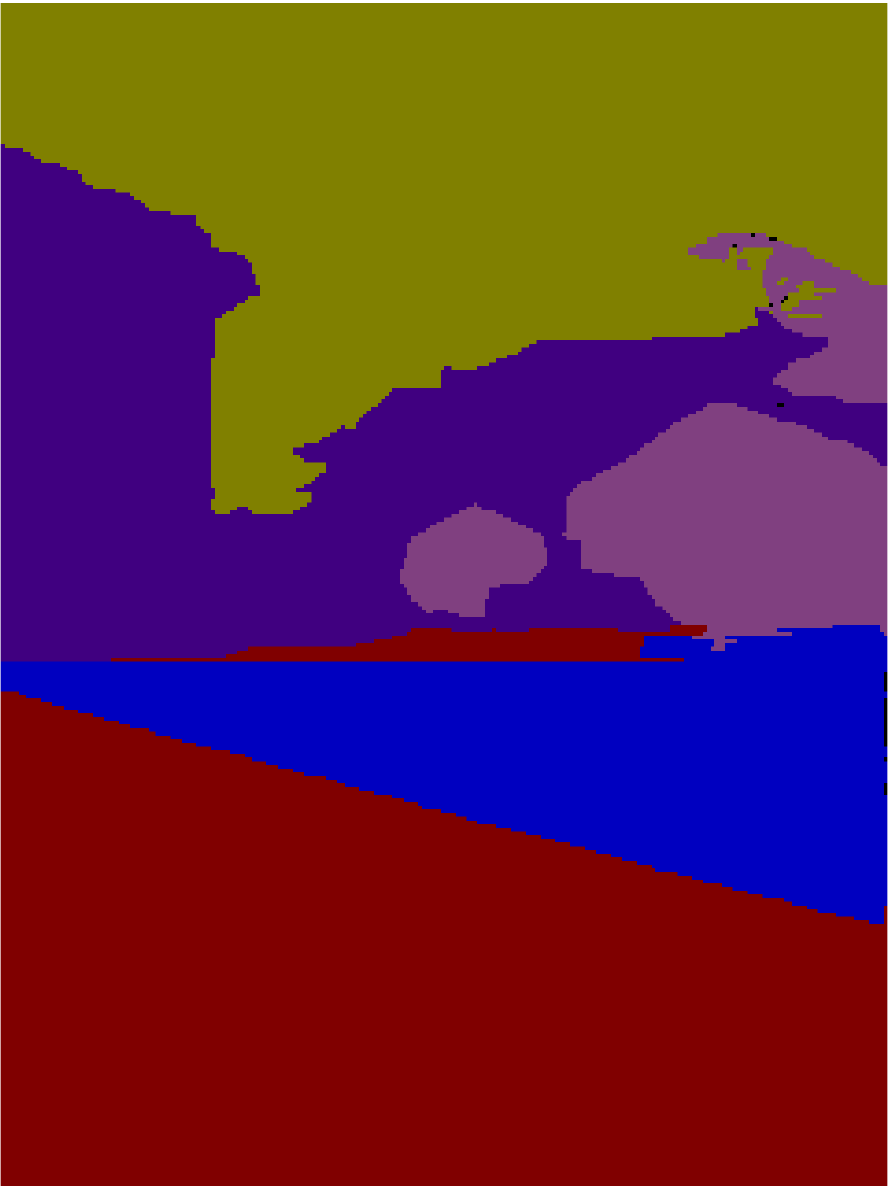} & 
\hspace{-0.0cm}\includegraphics[width=0.118\linewidth]{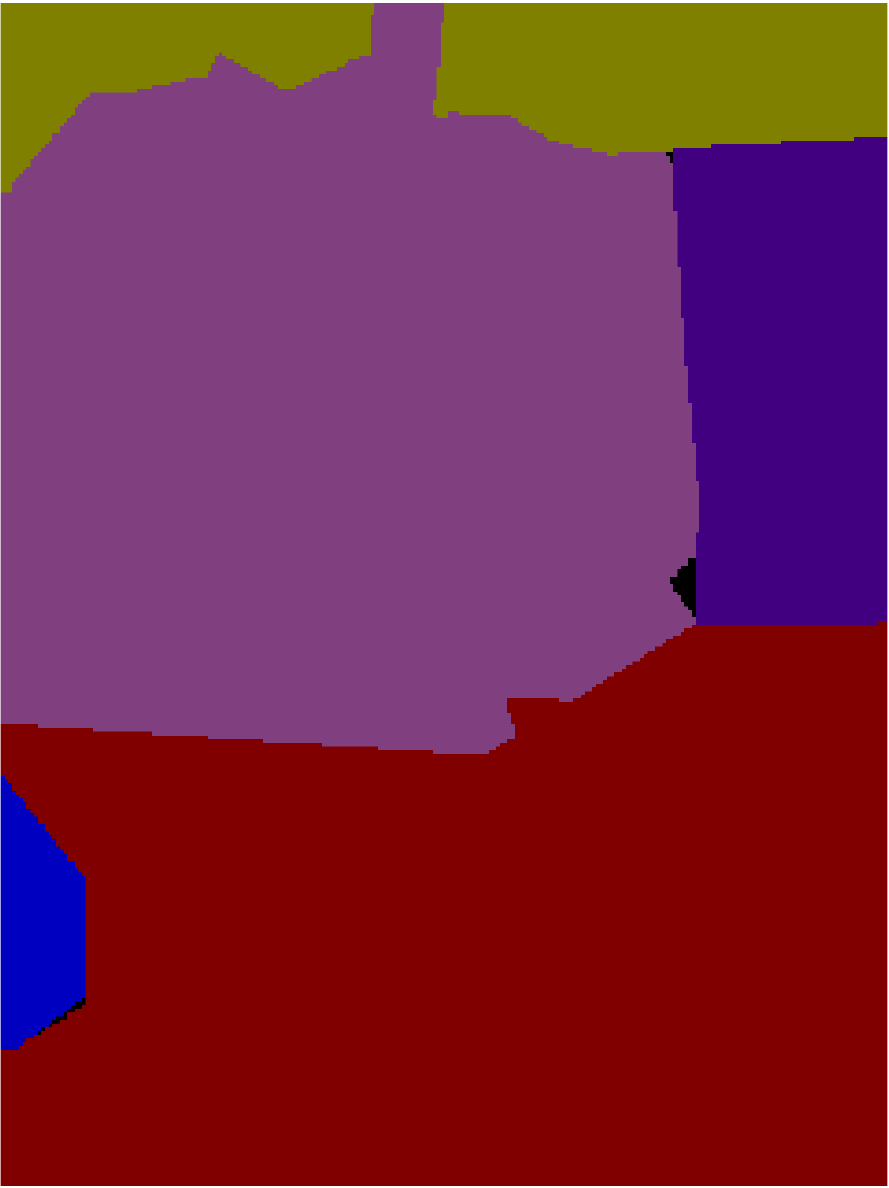}\\
\raisebox{.5em}{\rotatebox{90}{{ Low-Res Depth}}}&\hspace{-0.0cm}\includegraphics[width=0.118\linewidth]{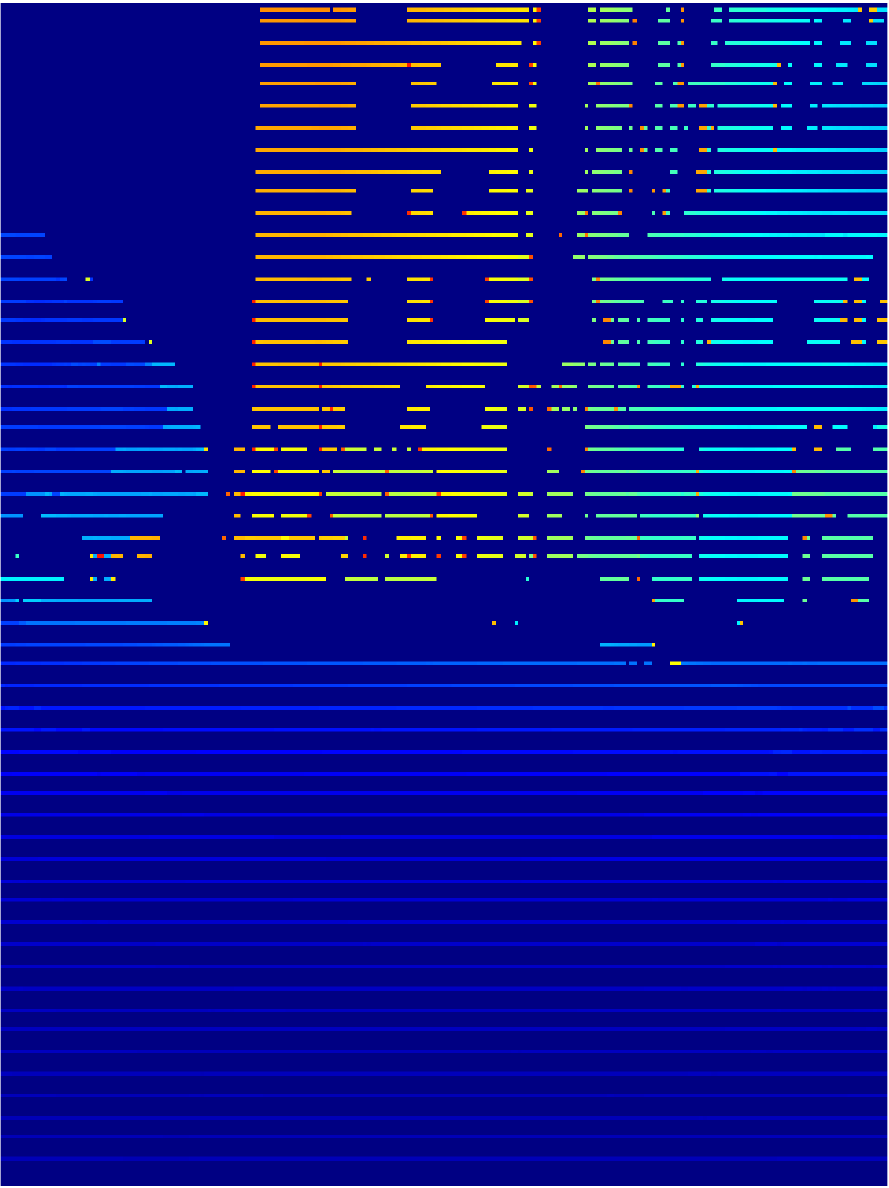} & 
\hspace{-0.0cm}\includegraphics[width=0.118\linewidth]{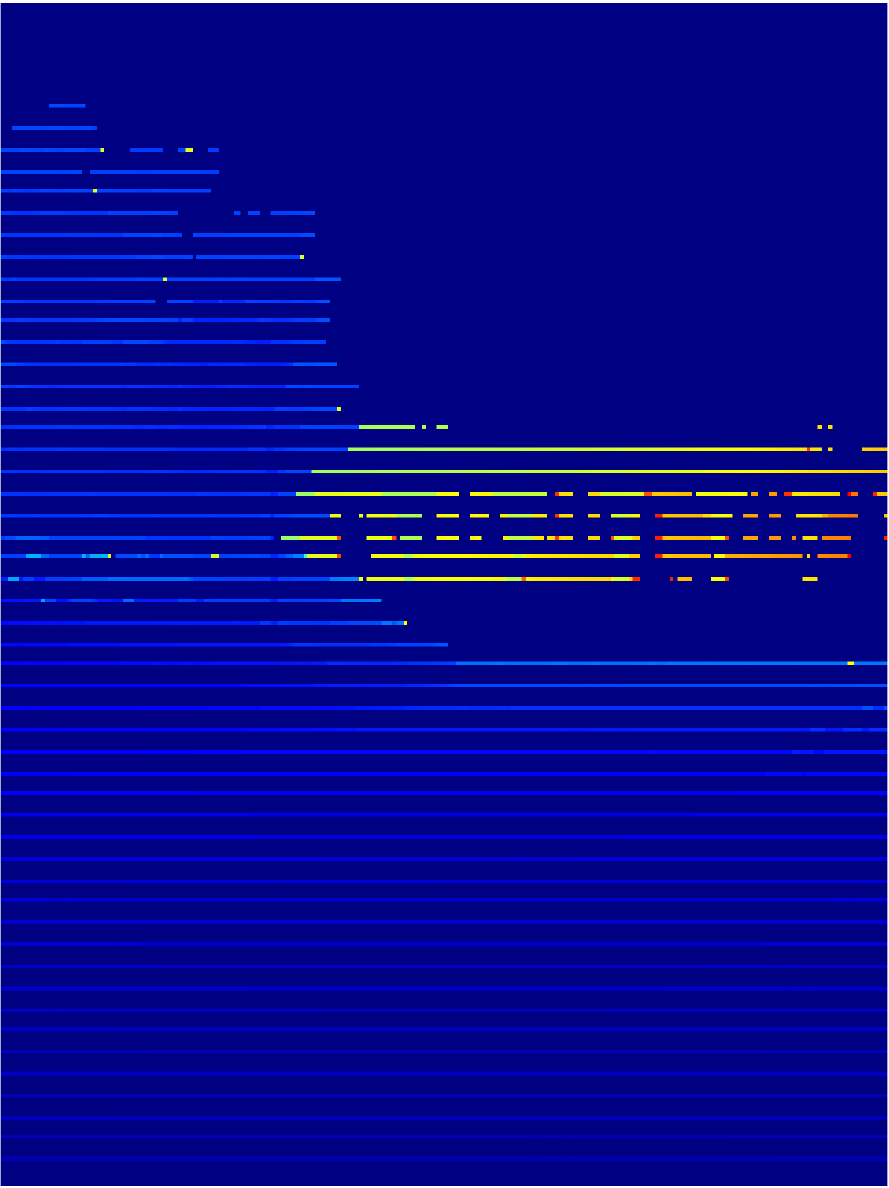}&
\hspace{-0.0cm}\includegraphics[width=0.118\linewidth]{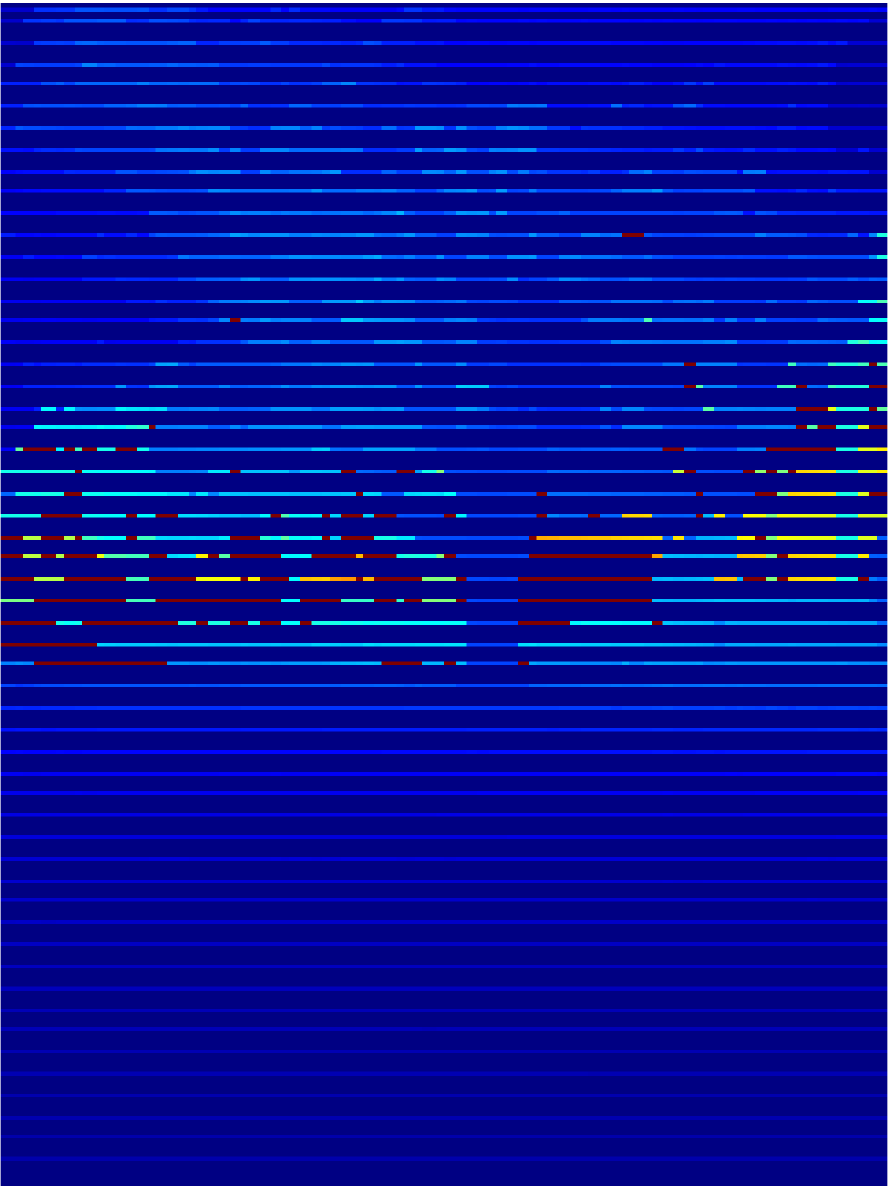} & 
\hspace{-0.0cm}\includegraphics[width=0.118\linewidth]{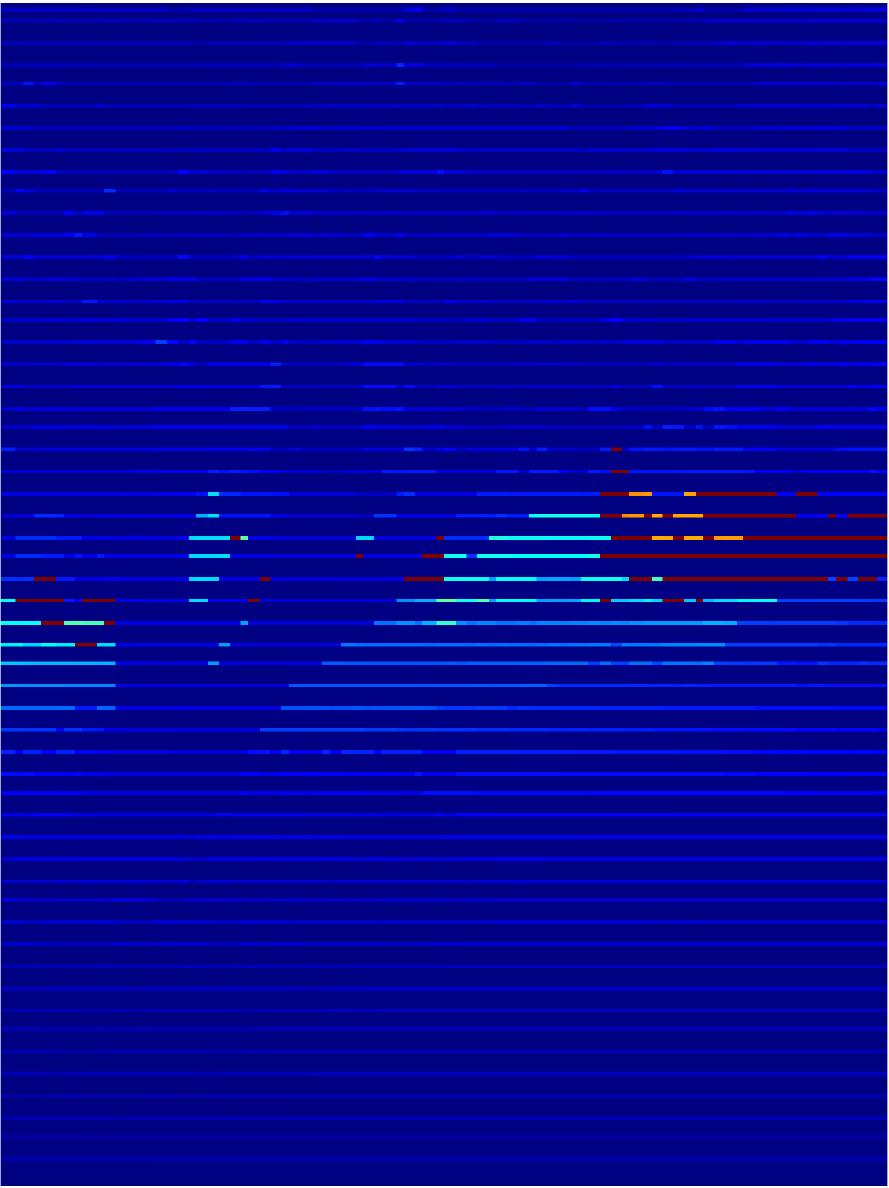}&
\hspace{-0.0cm}\includegraphics[width=0.118\linewidth]{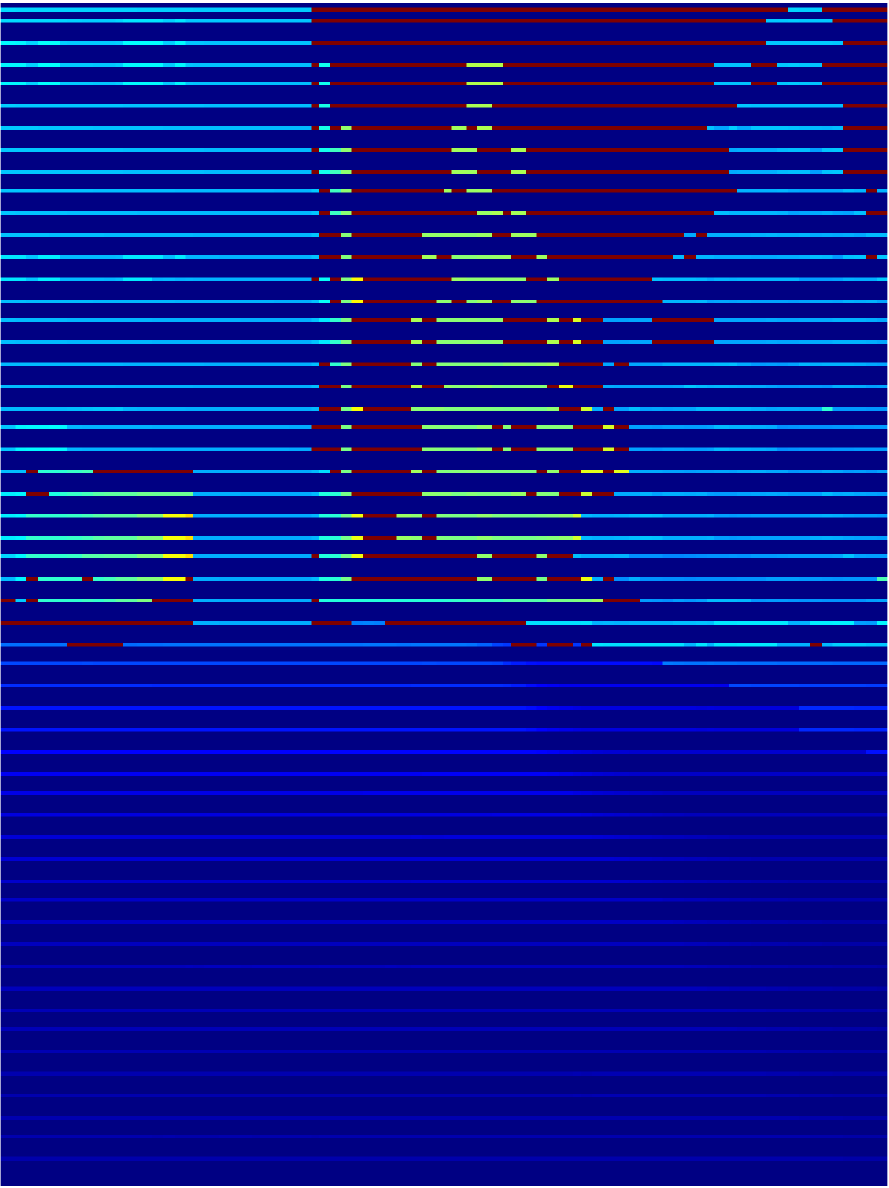} & 
\hspace{-0.0cm}\includegraphics[width=0.118\linewidth]{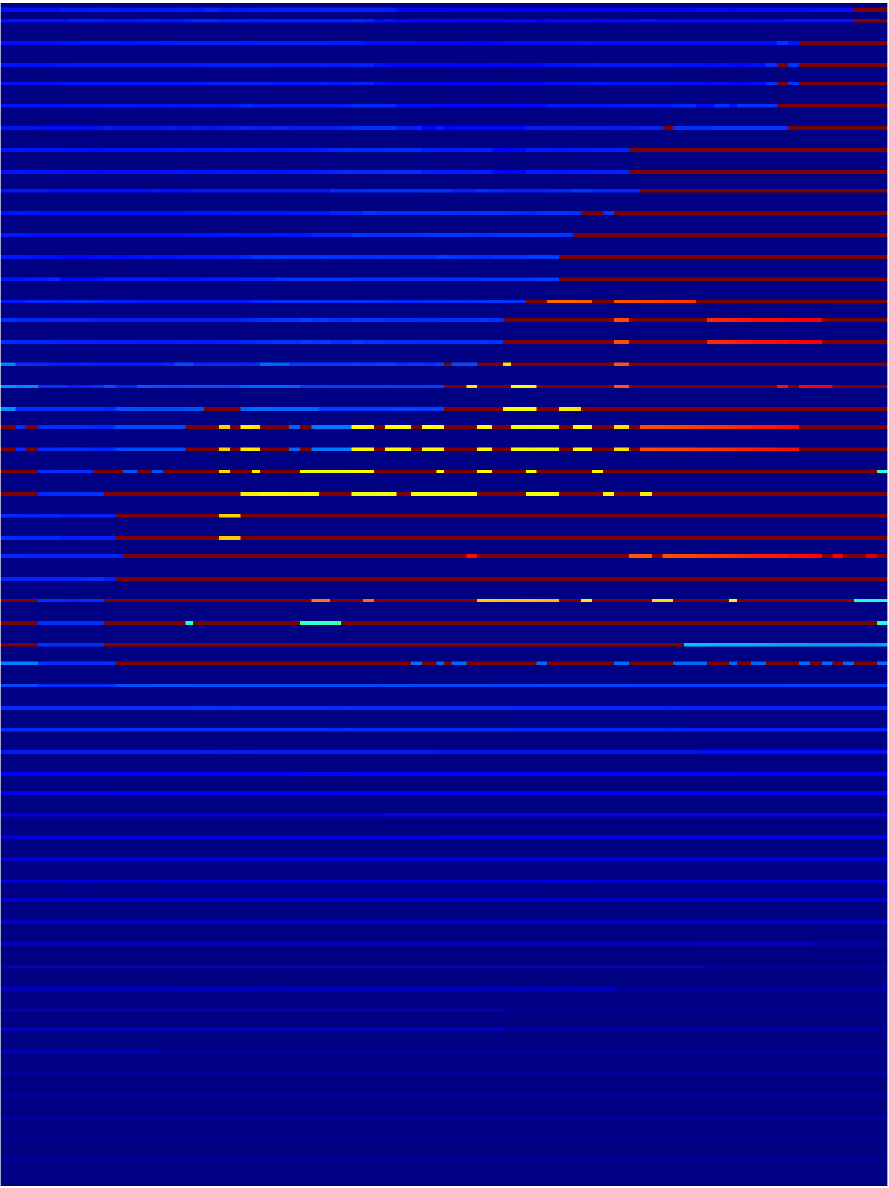} & 
\hspace{-0.0cm}\includegraphics[width=0.118\linewidth]{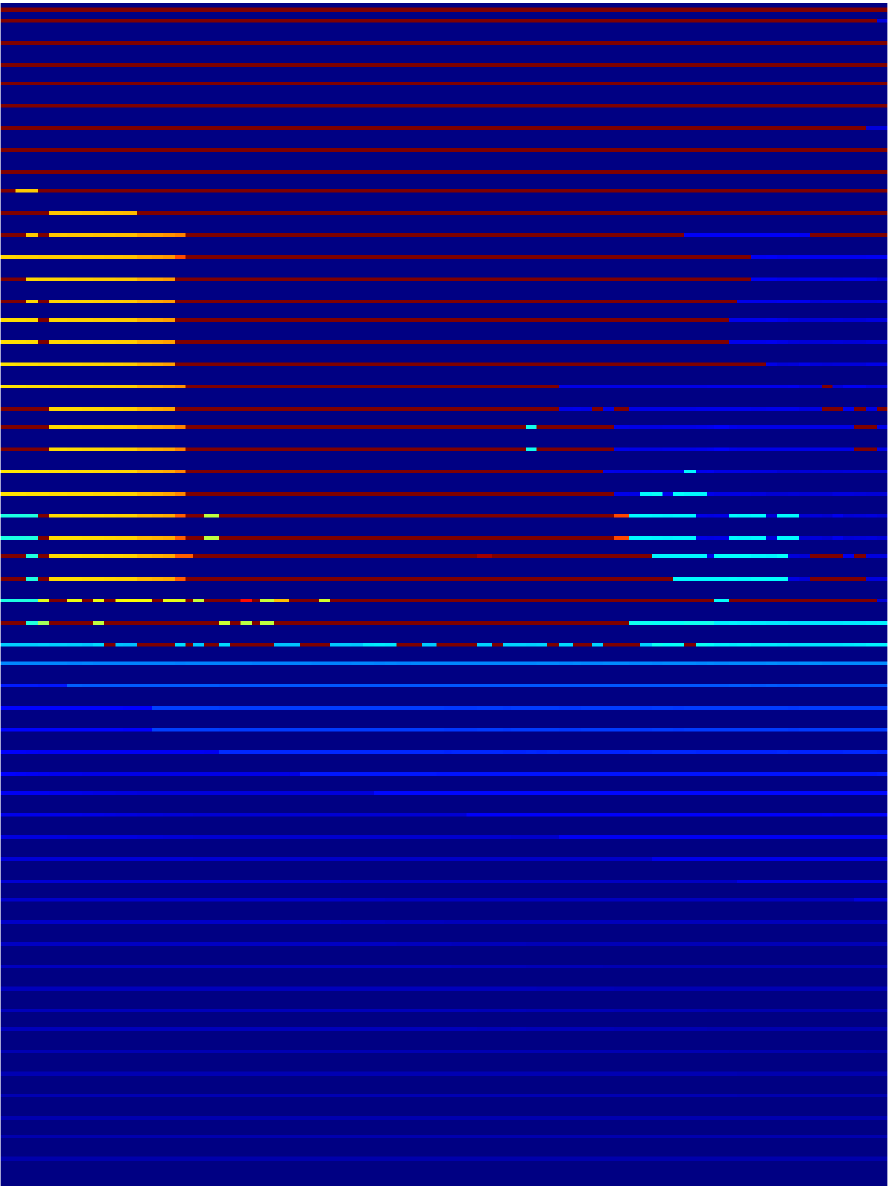} & 
\hspace{-0.0cm}\includegraphics[width=0.118\linewidth]{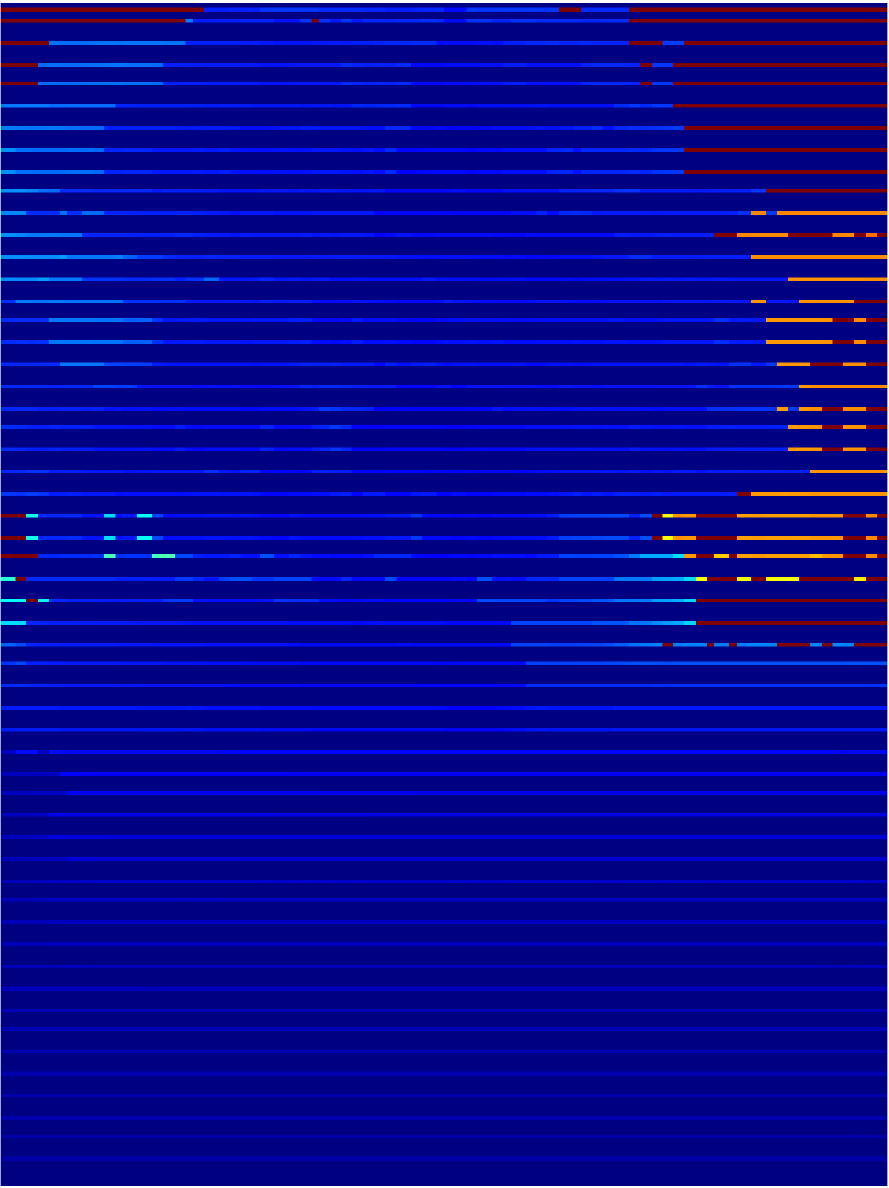}\\
\raisebox{0.0em}{\rotatebox{90}{{ Our Depth}}}&\hspace{-0.0cm}\includegraphics[width=0.118\linewidth]{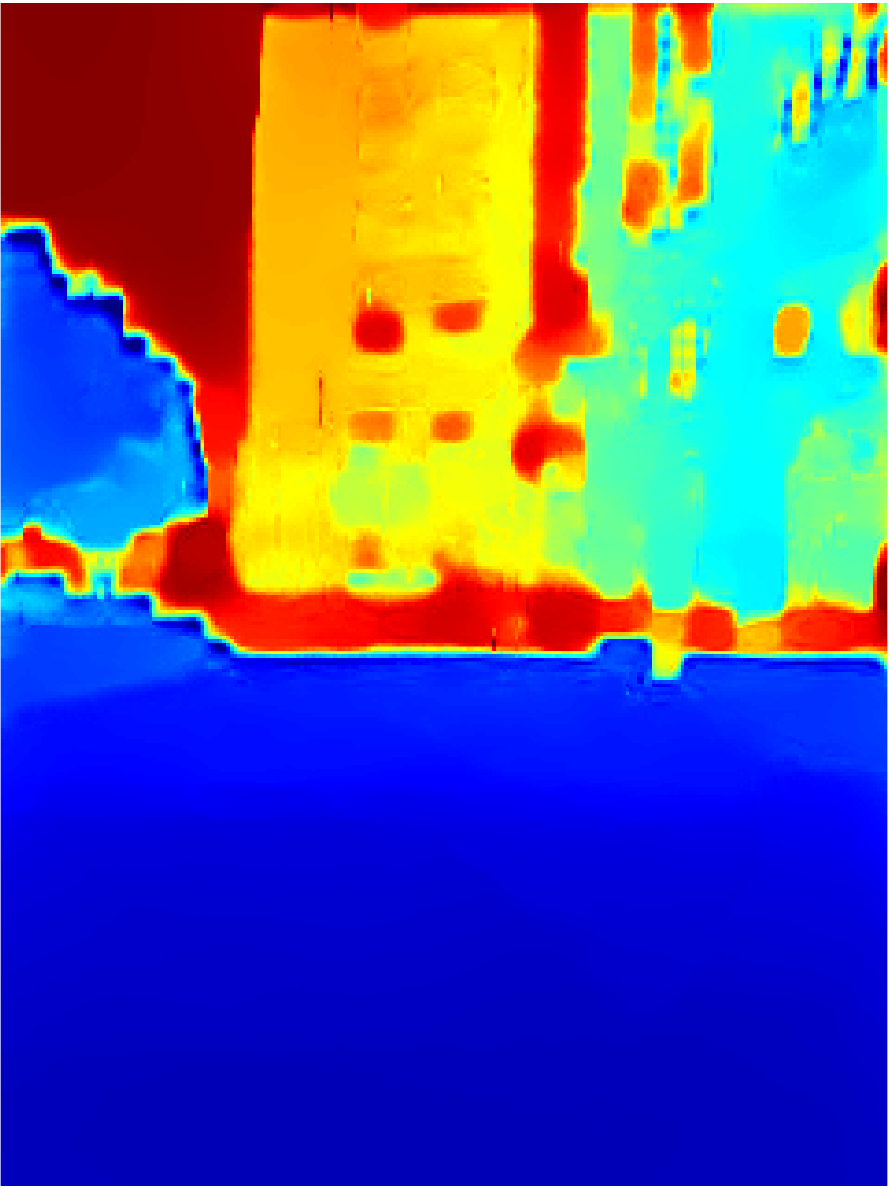} & 
\hspace{-0.0cm}\includegraphics[width=0.118\linewidth]{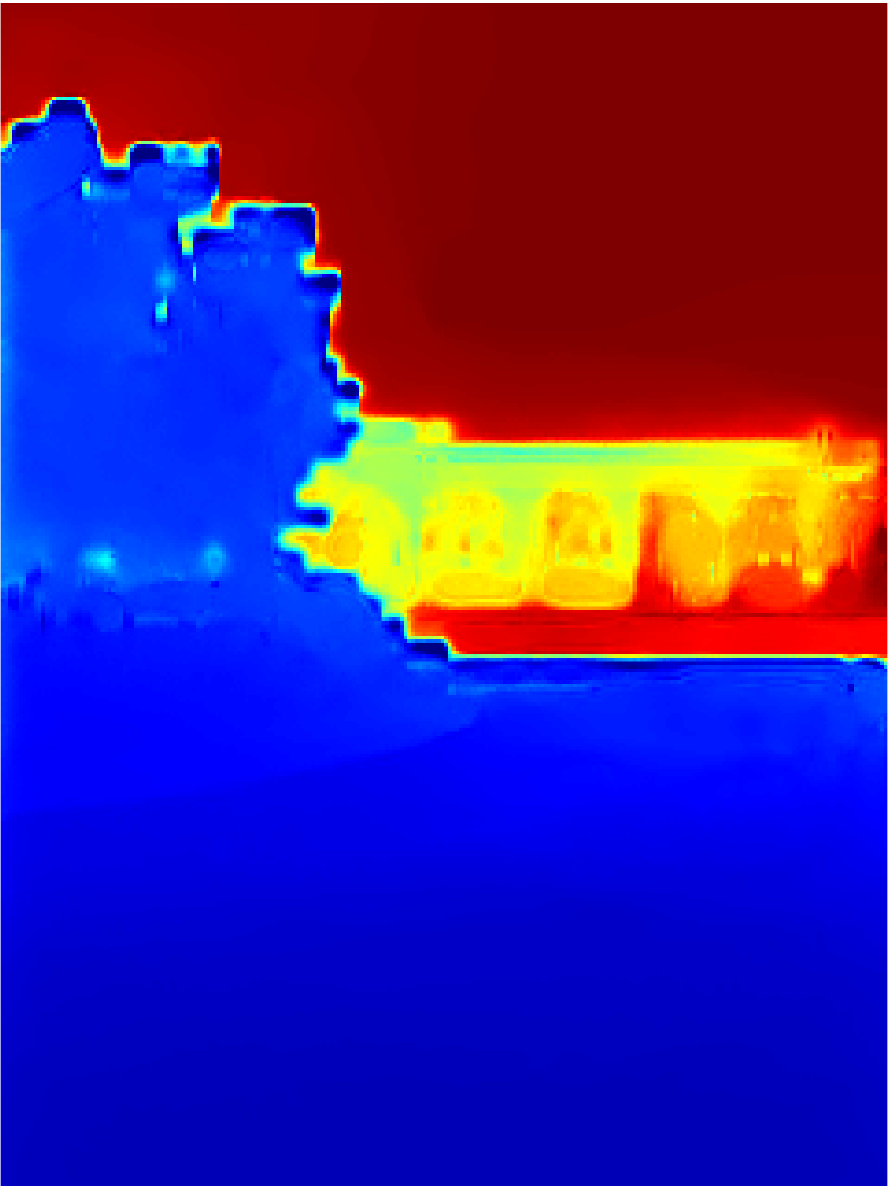}&
\hspace{-0.0cm}\includegraphics[width=0.118\linewidth]{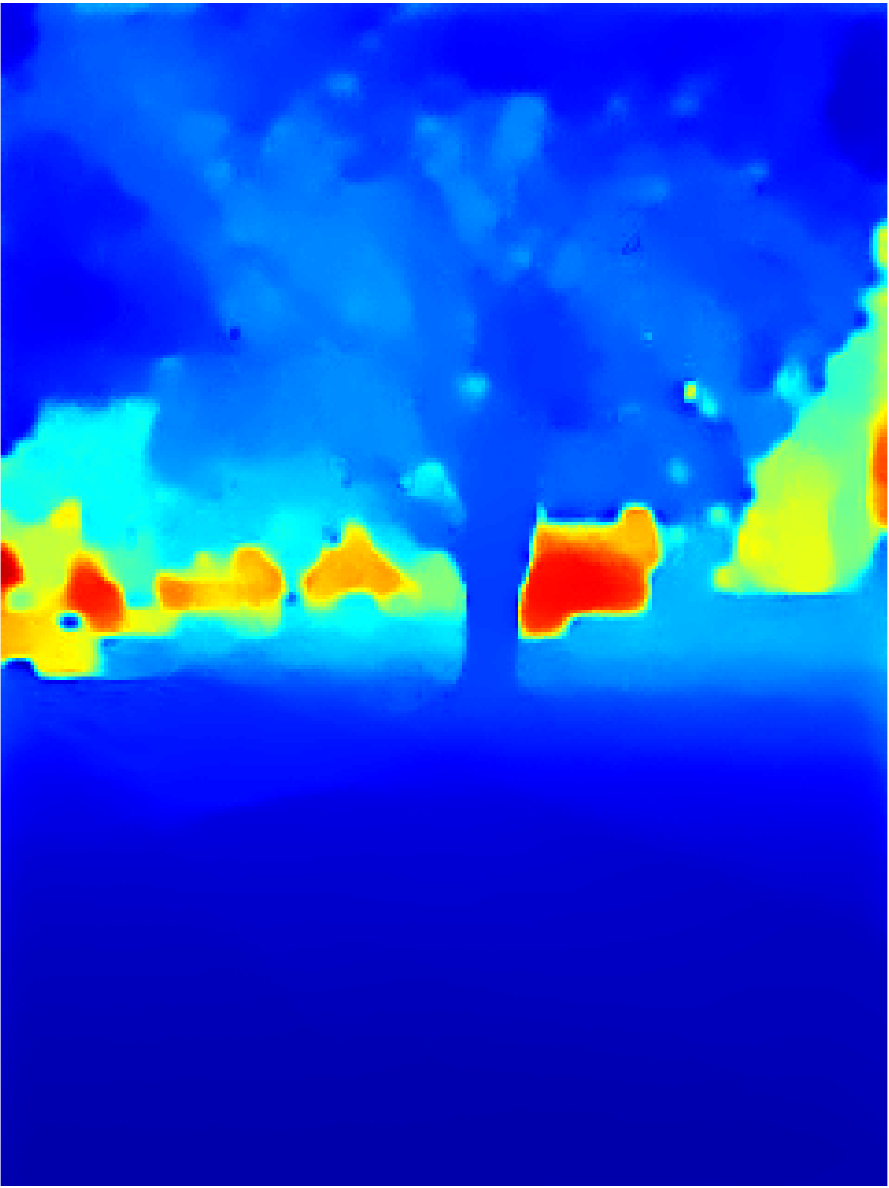} & 
\hspace{-0.0cm}\includegraphics[width=0.118\linewidth]{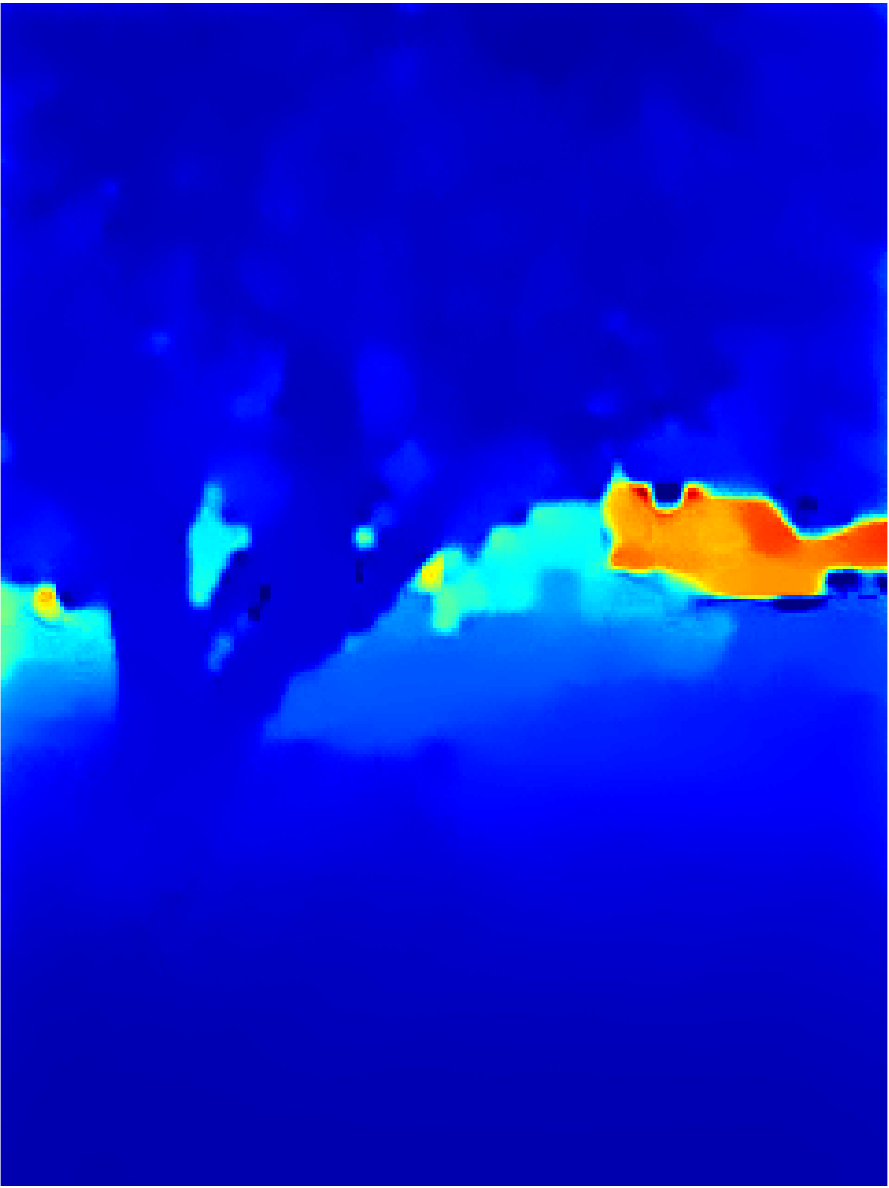}&
\hspace{-0.0cm}\includegraphics[width=0.118\linewidth]{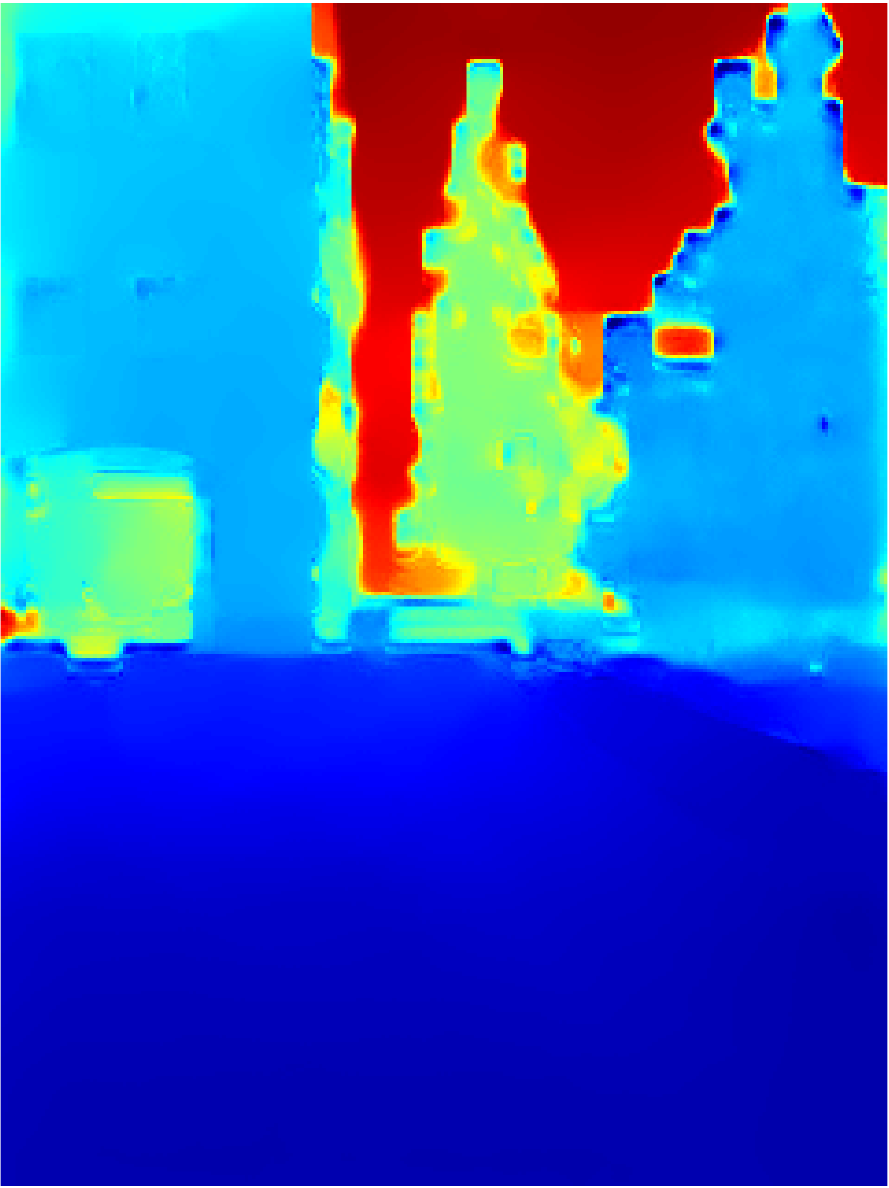} & 
\hspace{-0.0cm}\includegraphics[width=0.118\linewidth]{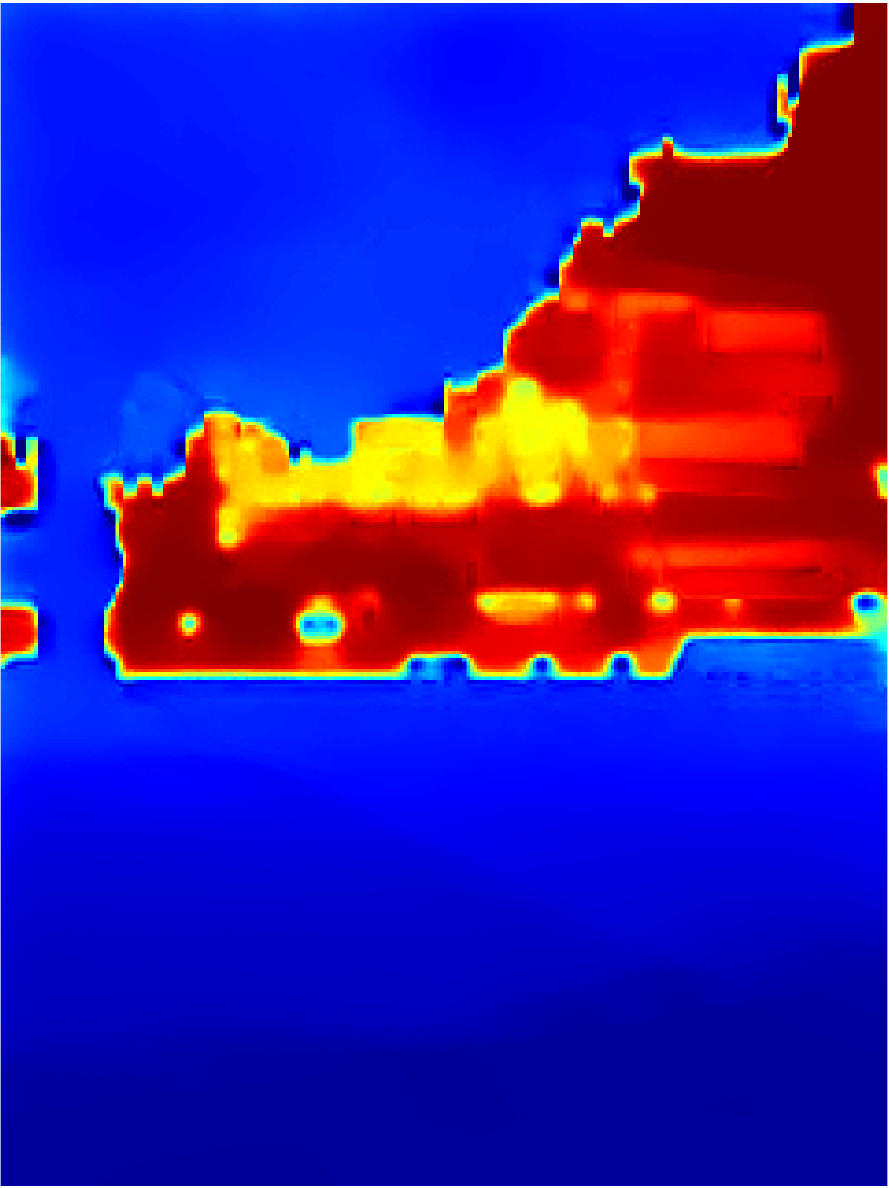} & 
\hspace{-0.0cm}\includegraphics[width=0.118\linewidth]{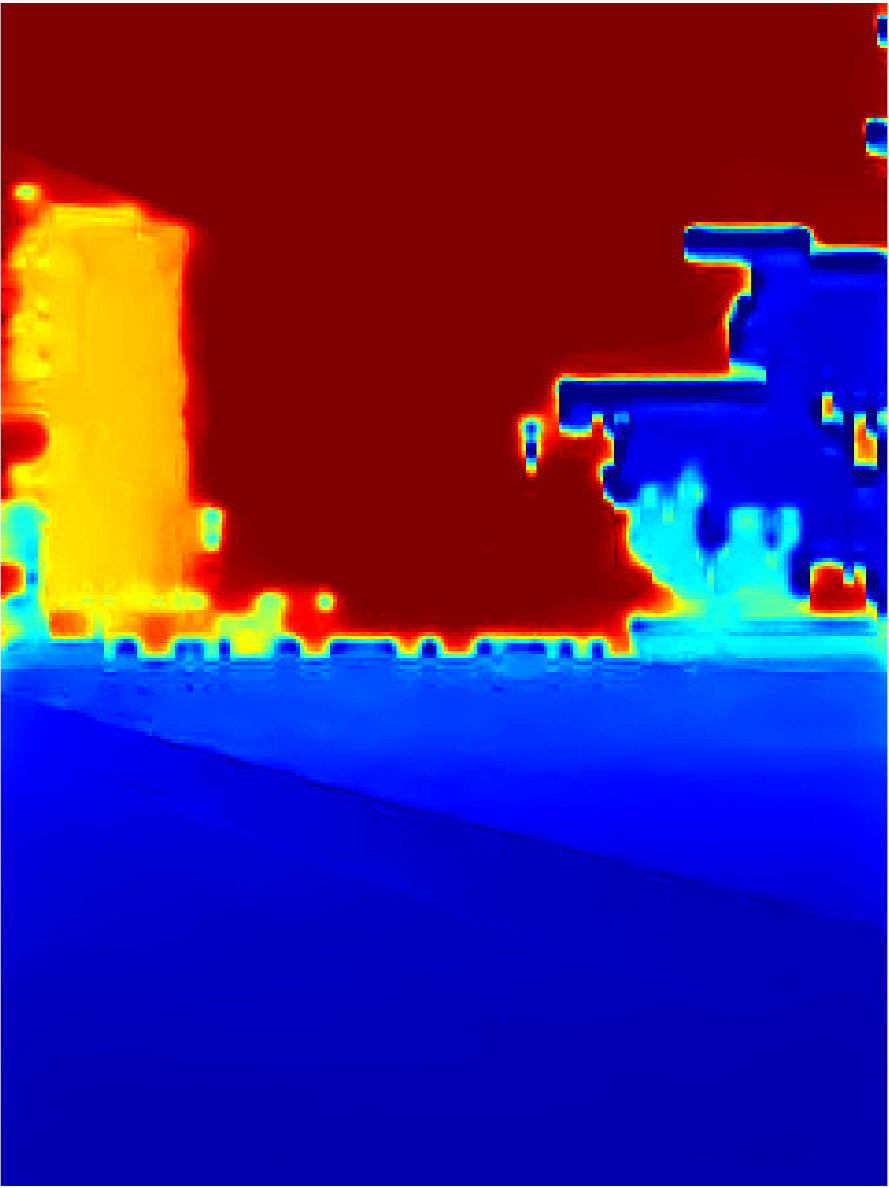} & 
\hspace{-0.0cm}\includegraphics[width=0.118\linewidth]{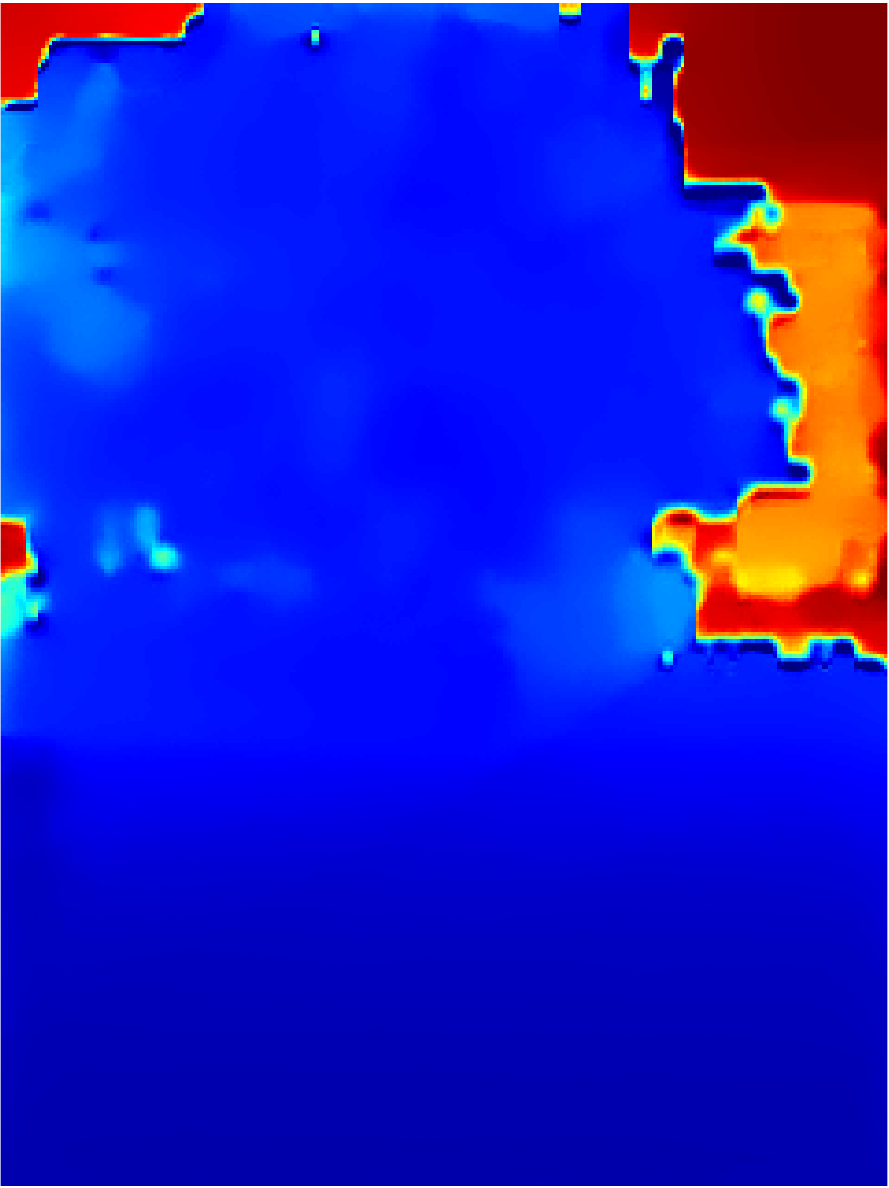}\\
\multicolumn{9}{c}{\hspace{-0.0cm}\includegraphics[width=0.8\linewidth]{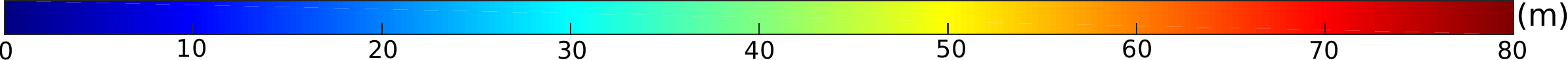}}\\
\raisebox{0.0em}{\rotatebox{90}{{ Noisy Semantic}}}&\hspace{-0.0cm}\includegraphics[width=0.118\linewidth]{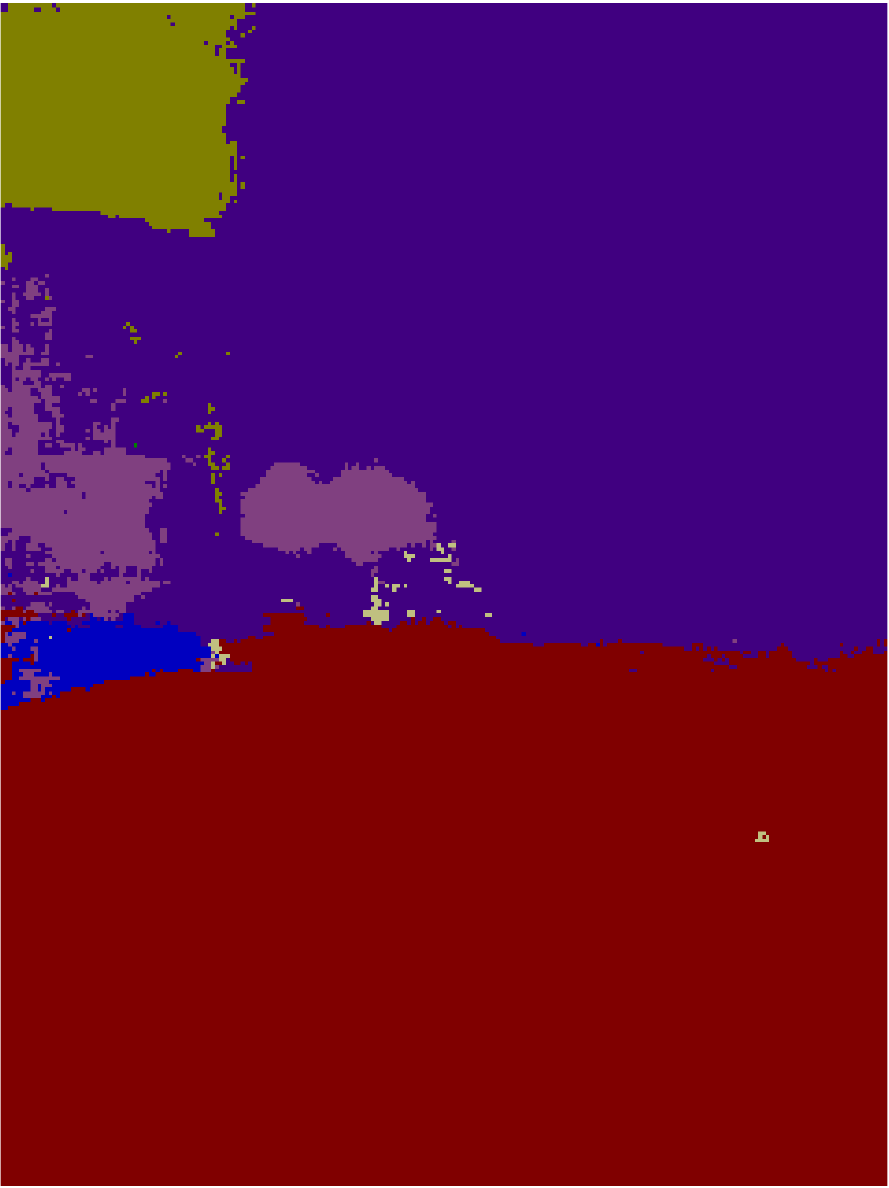} & 
\hspace{-0.0cm}\includegraphics[width=0.118\linewidth]{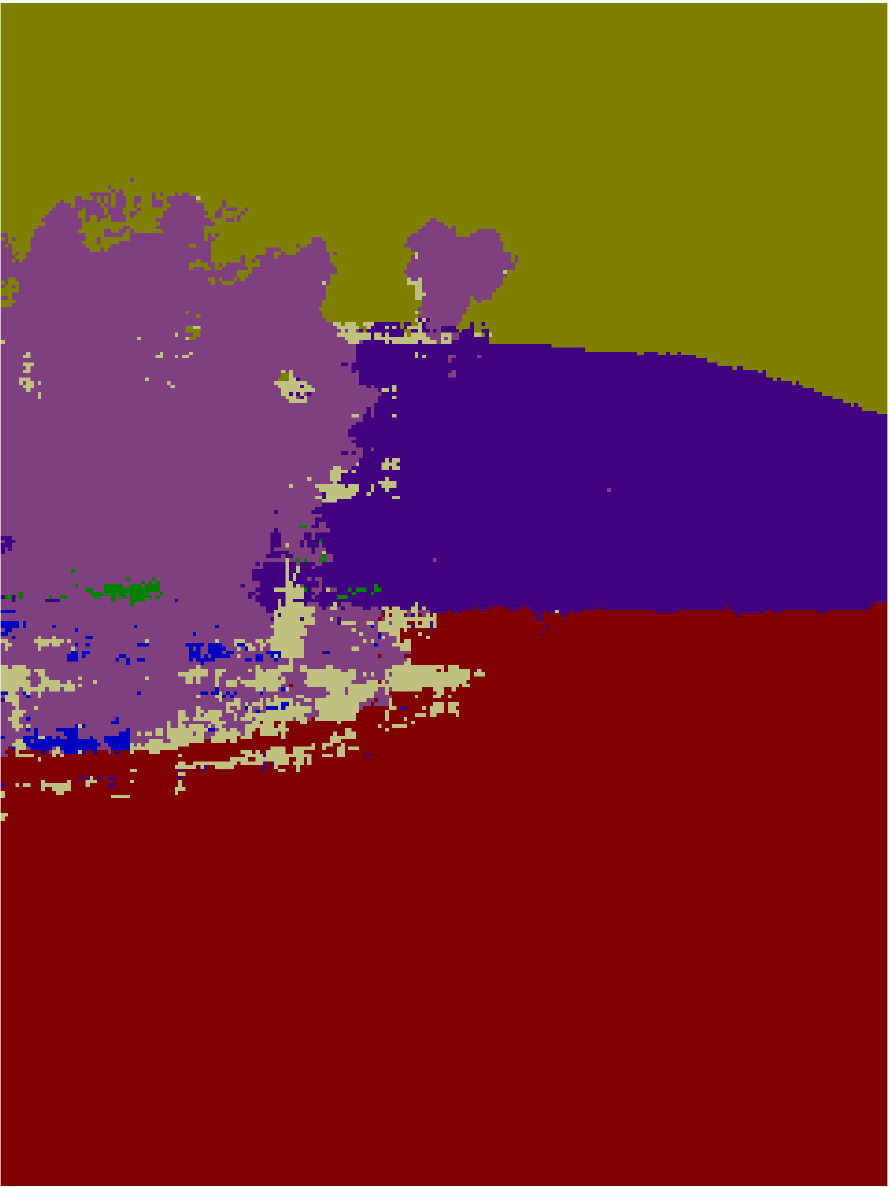}&
\hspace{-0.0cm}\includegraphics[width=0.118\linewidth]{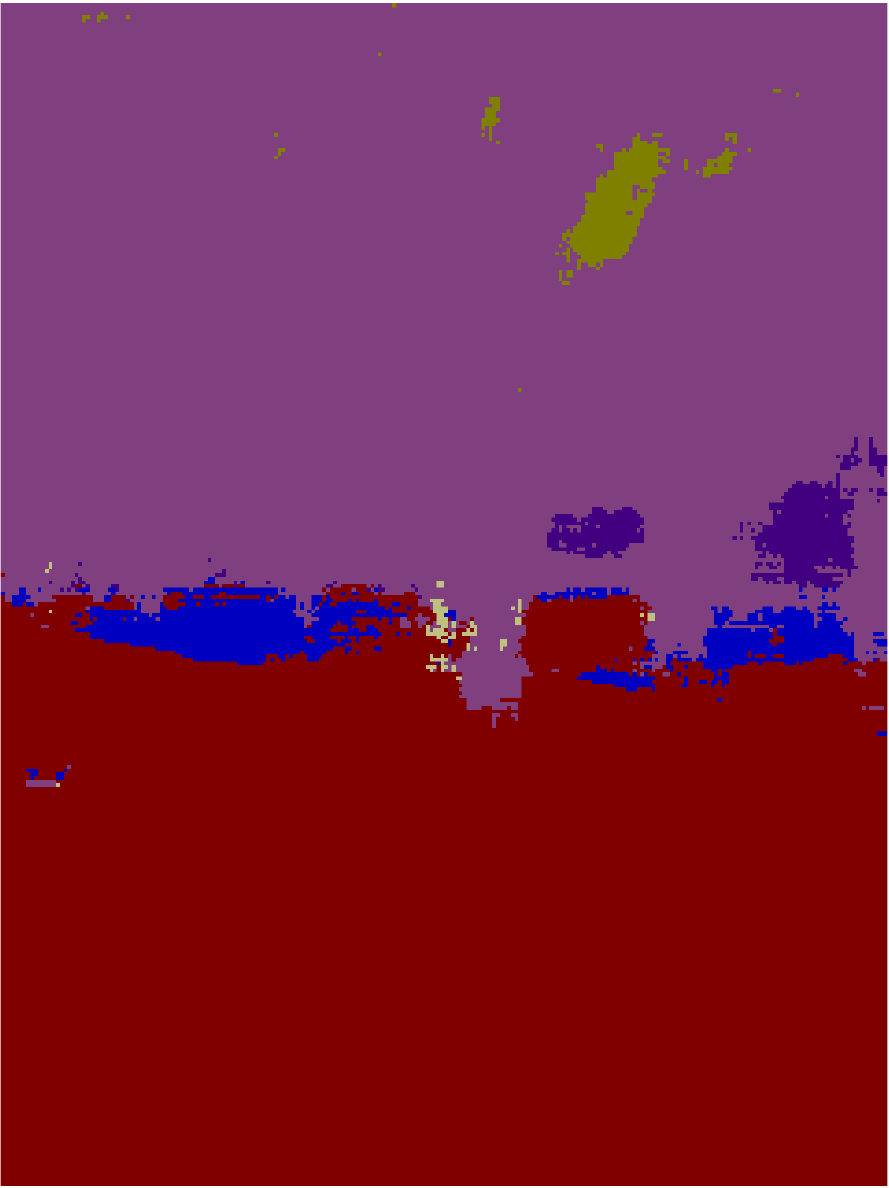} & 
\hspace{-0.0cm}\includegraphics[width=0.118\linewidth]{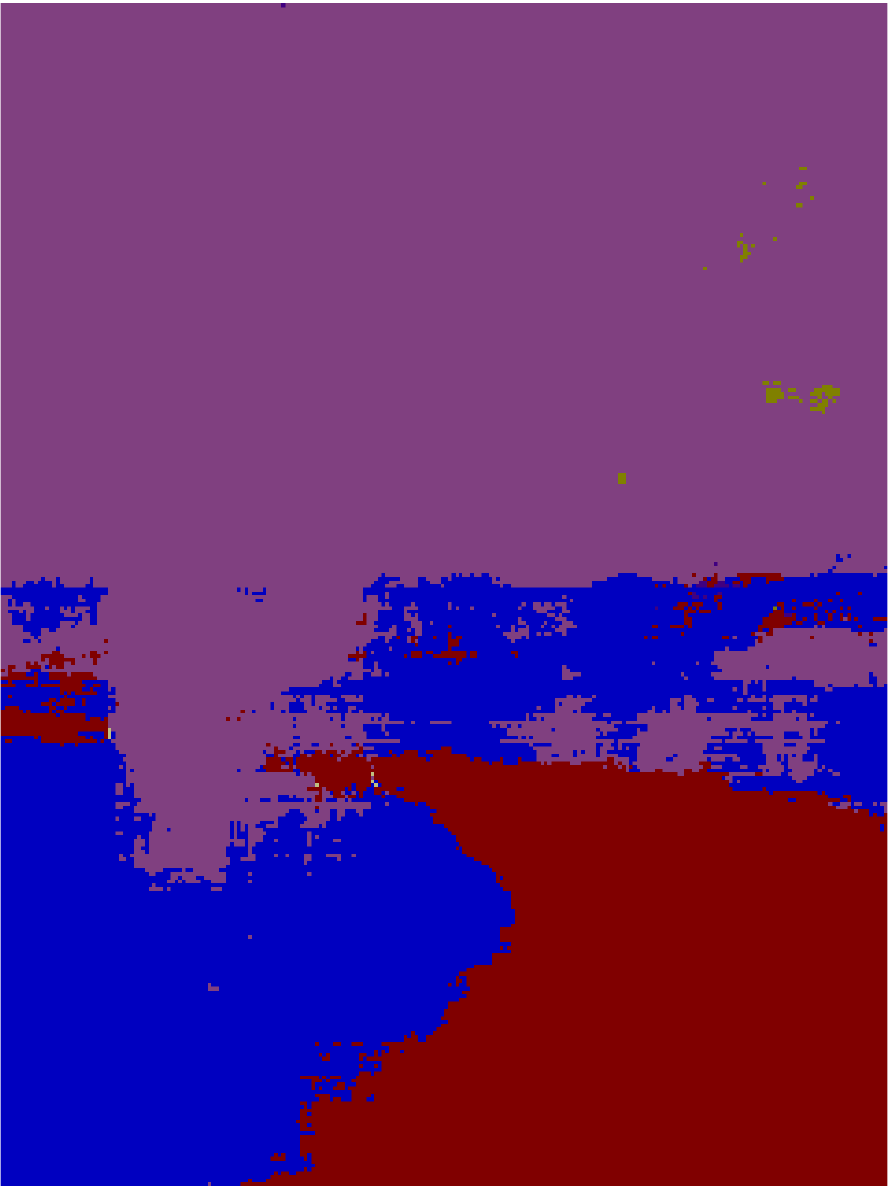}&
\hspace{-0.0cm}\includegraphics[width=0.118\linewidth]{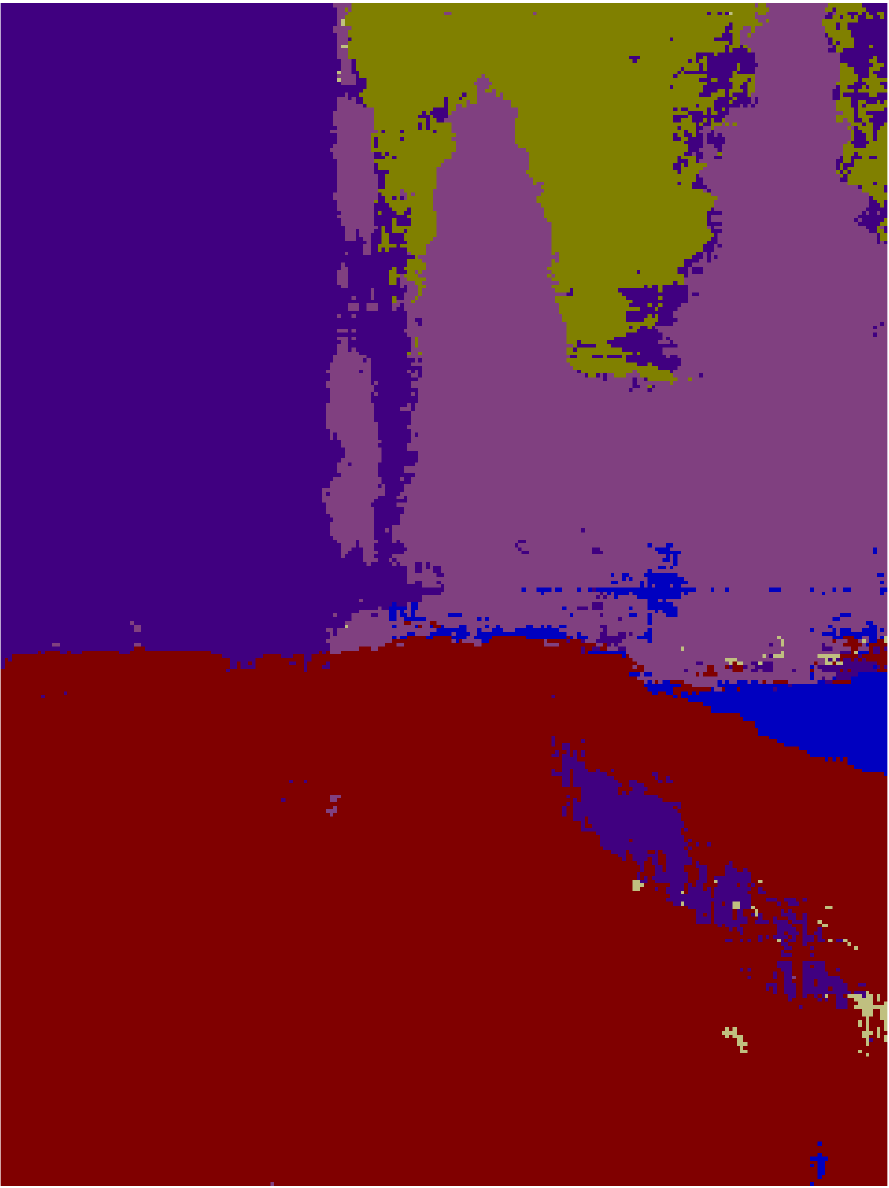} & 
\hspace{-0.0cm}\includegraphics[width=0.118\linewidth]{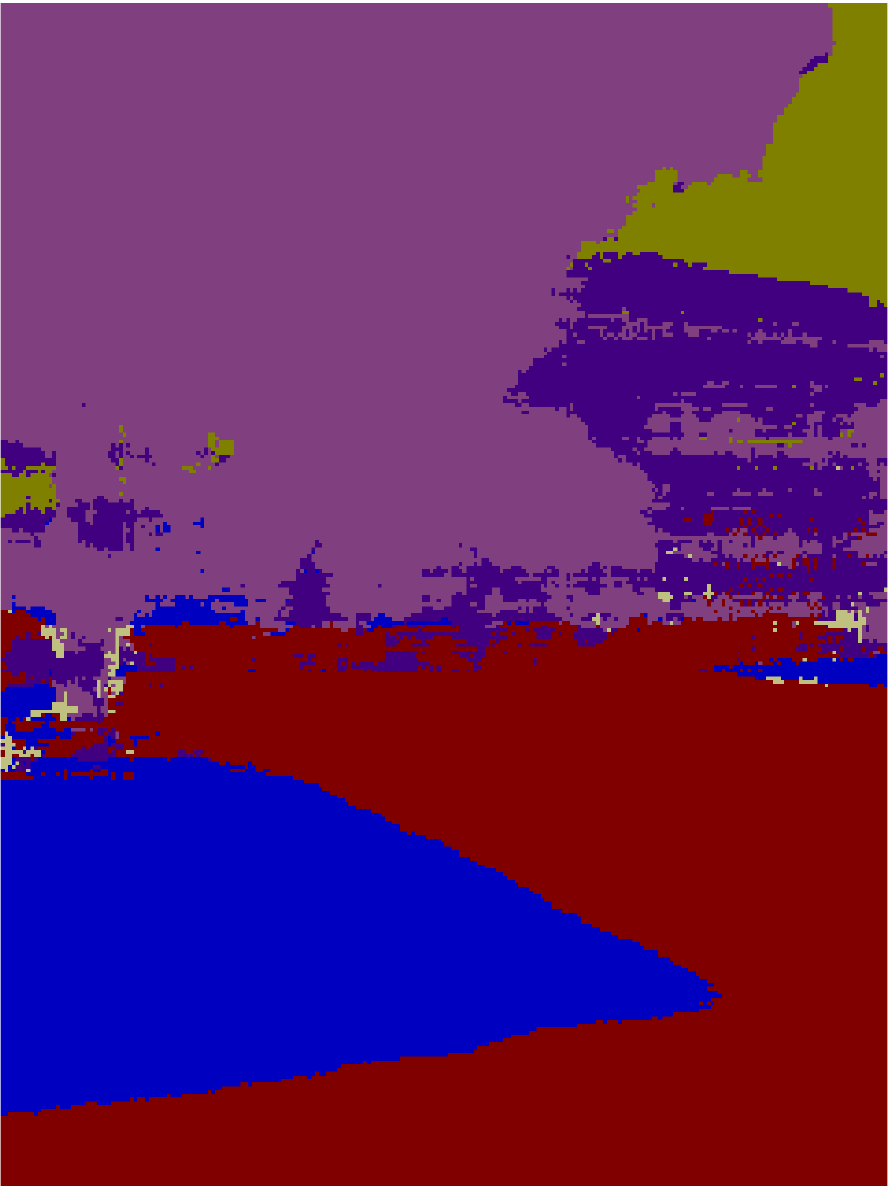} & 
\hspace{-0.0cm}\includegraphics[width=0.118\linewidth]{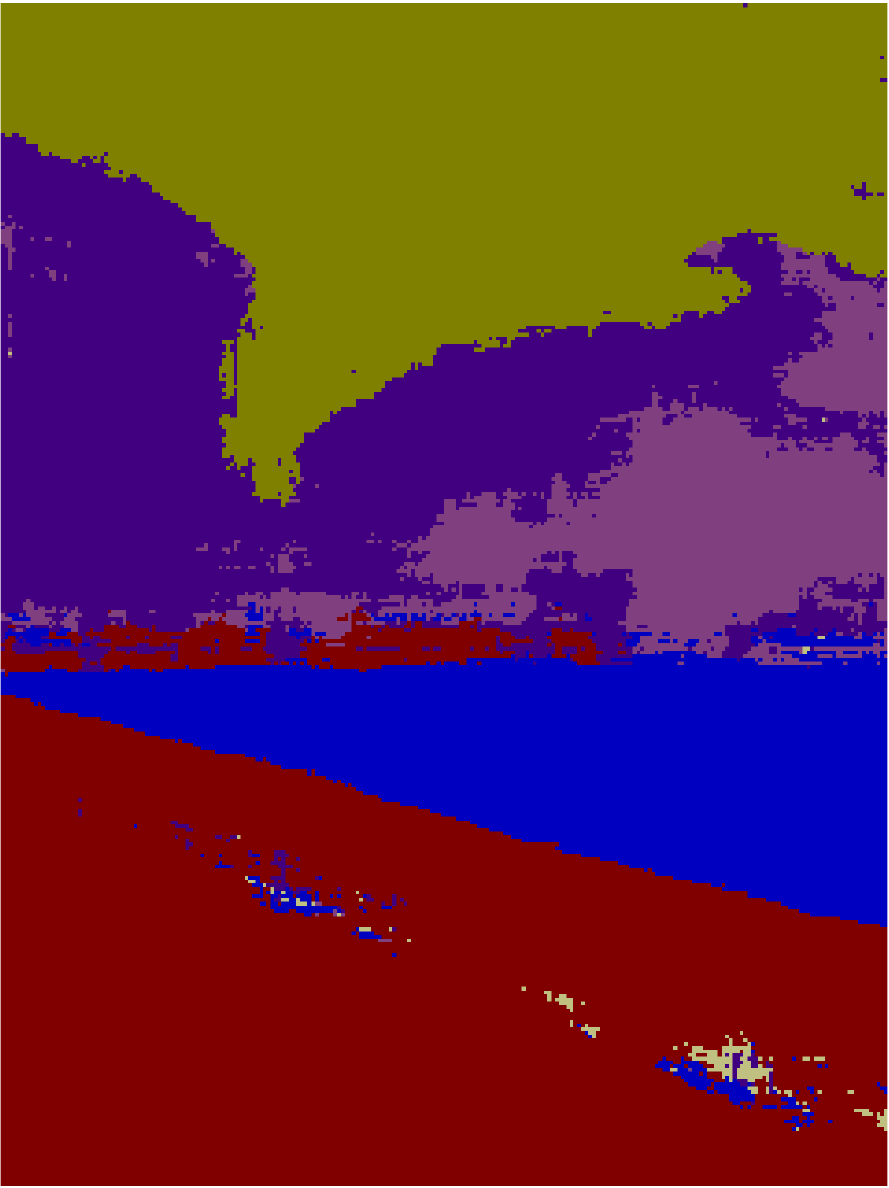} & 
\hspace{-0.0cm}\includegraphics[width=0.118\linewidth]{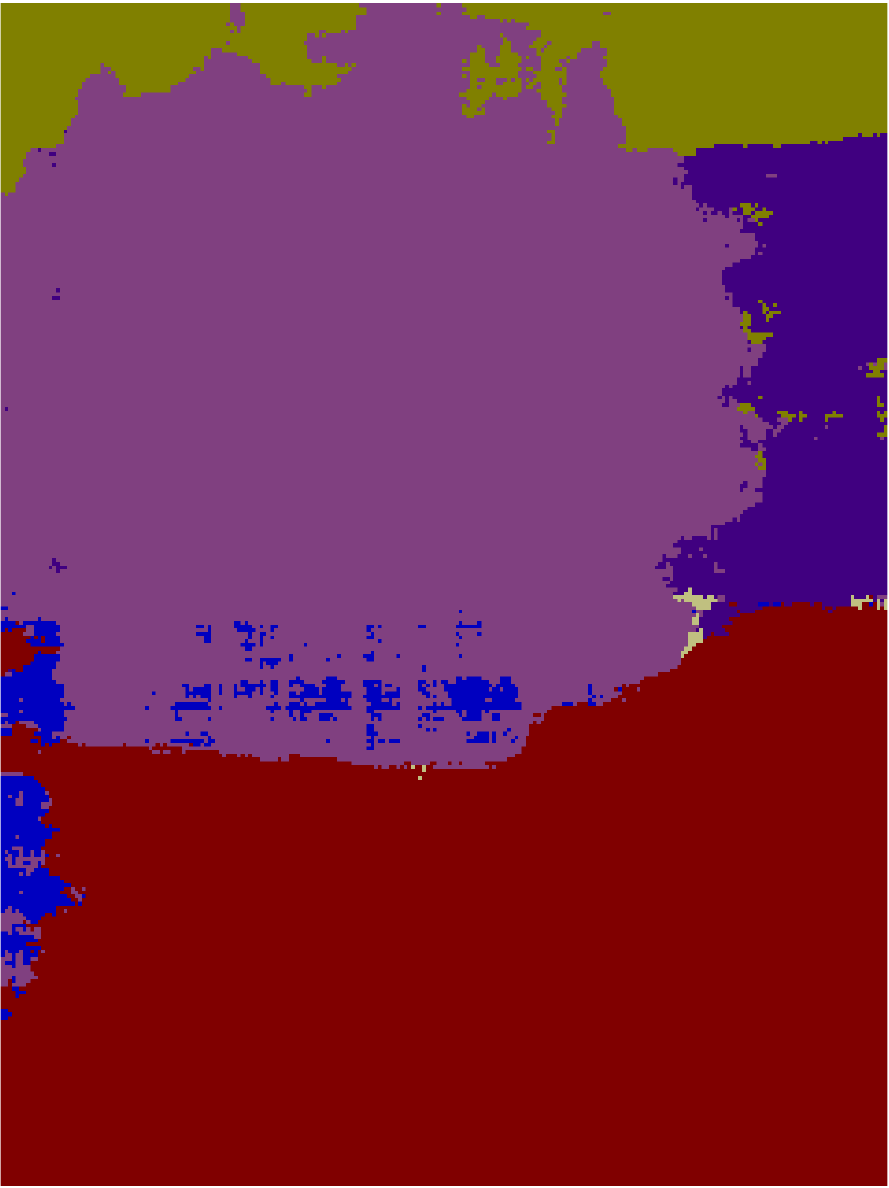}\\
\raisebox{0.0em}{\rotatebox{90}{{ Our Semantic}}}&\hspace{-0.0cm}\includegraphics[width=0.118\linewidth]{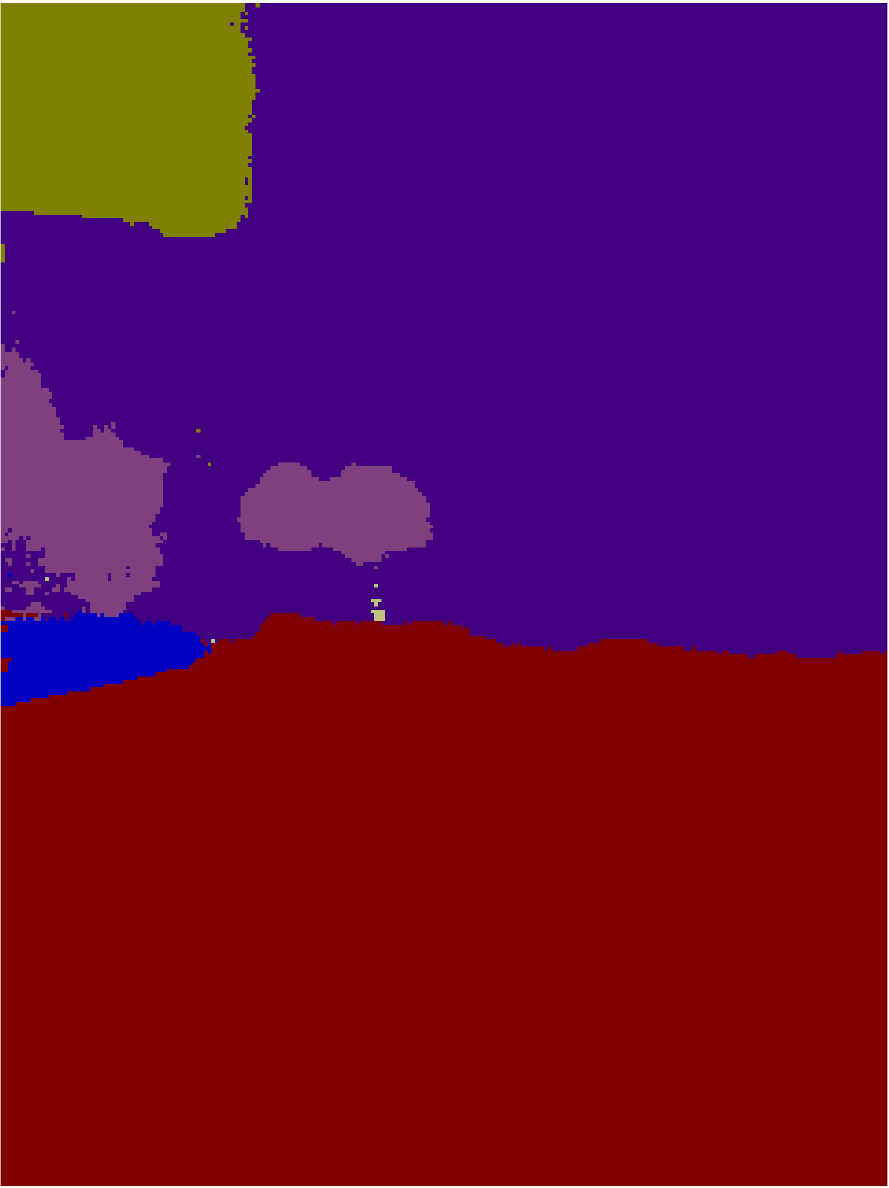} & 
\hspace{-0.0cm}\includegraphics[width=0.118\linewidth]{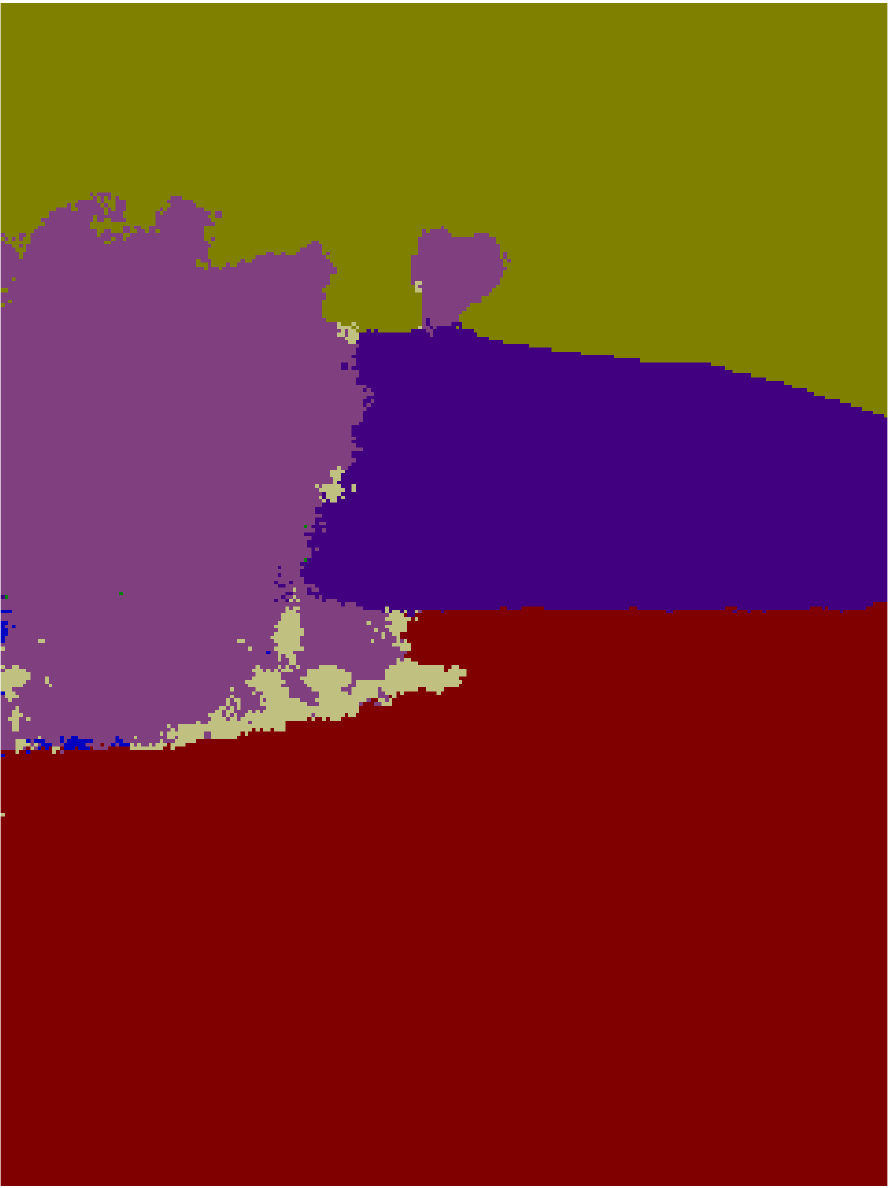}&
\hspace{-0.0cm}\includegraphics[width=0.118\linewidth]{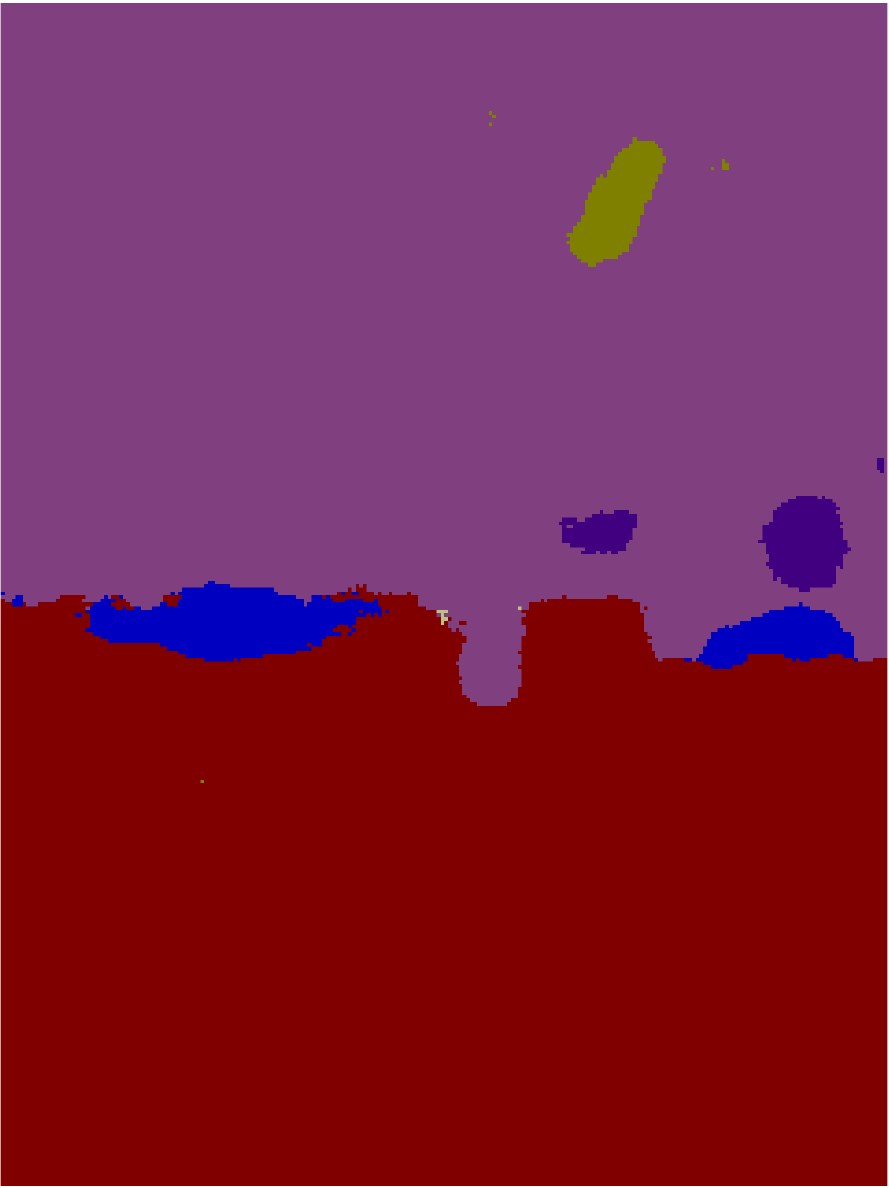} & 
\hspace{-0.0cm}\includegraphics[width=0.118\linewidth]{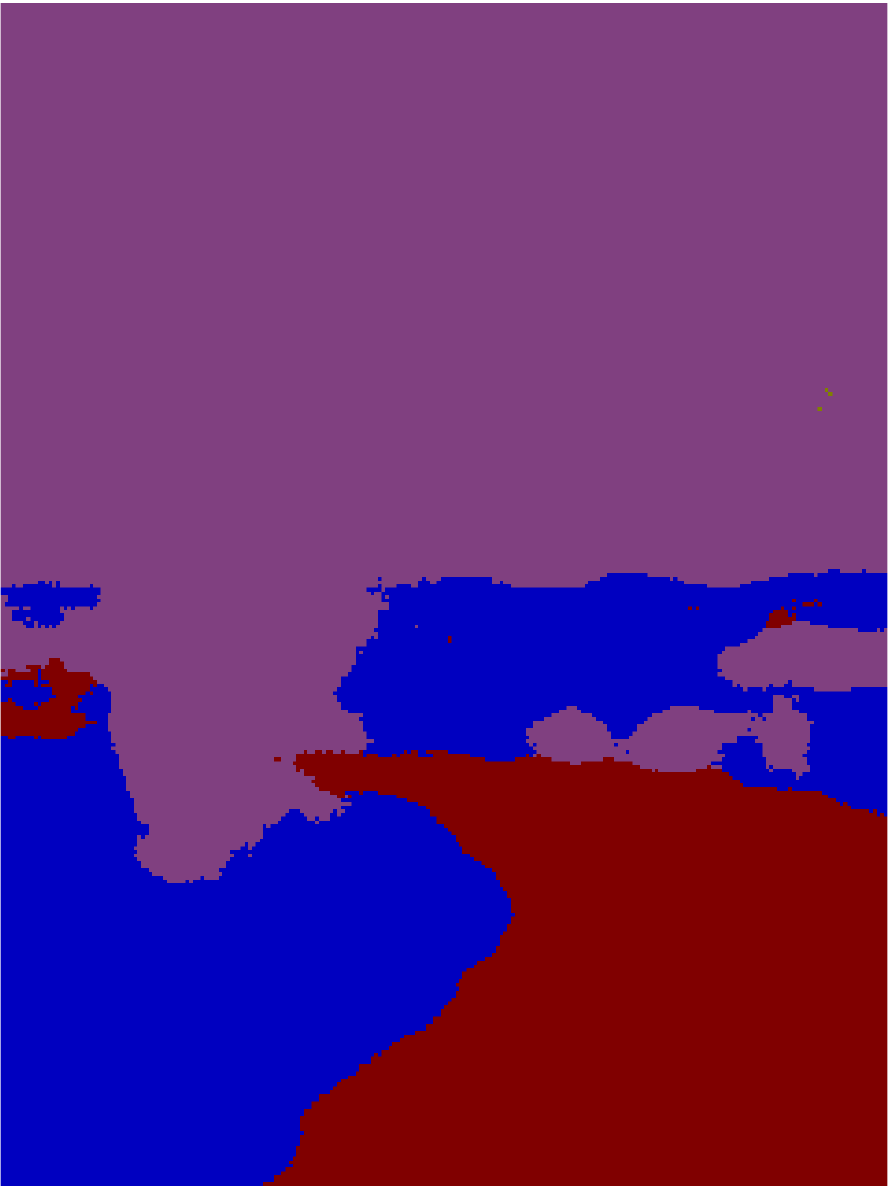}&
\hspace{-0.0cm}\includegraphics[width=0.118\linewidth]{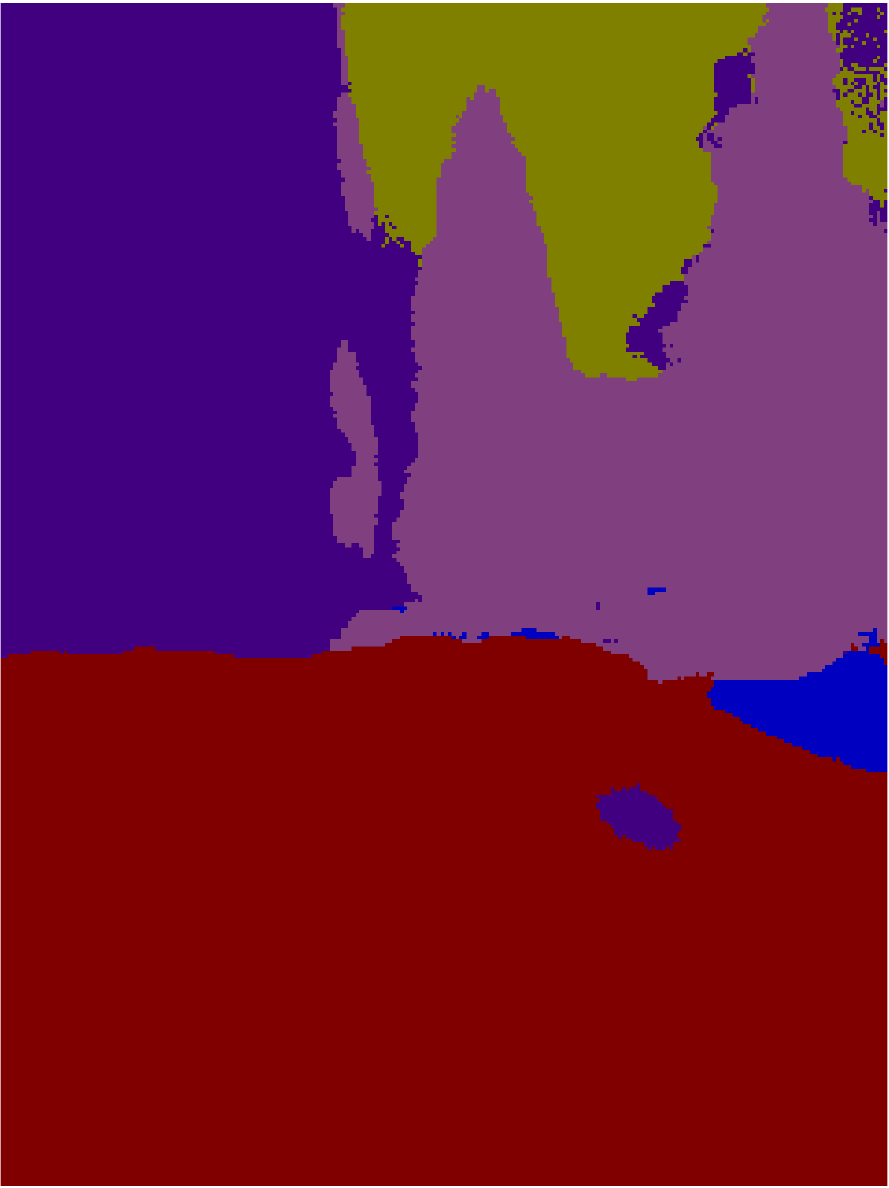} & 
\hspace{-0.0cm}\includegraphics[width=0.118\linewidth]{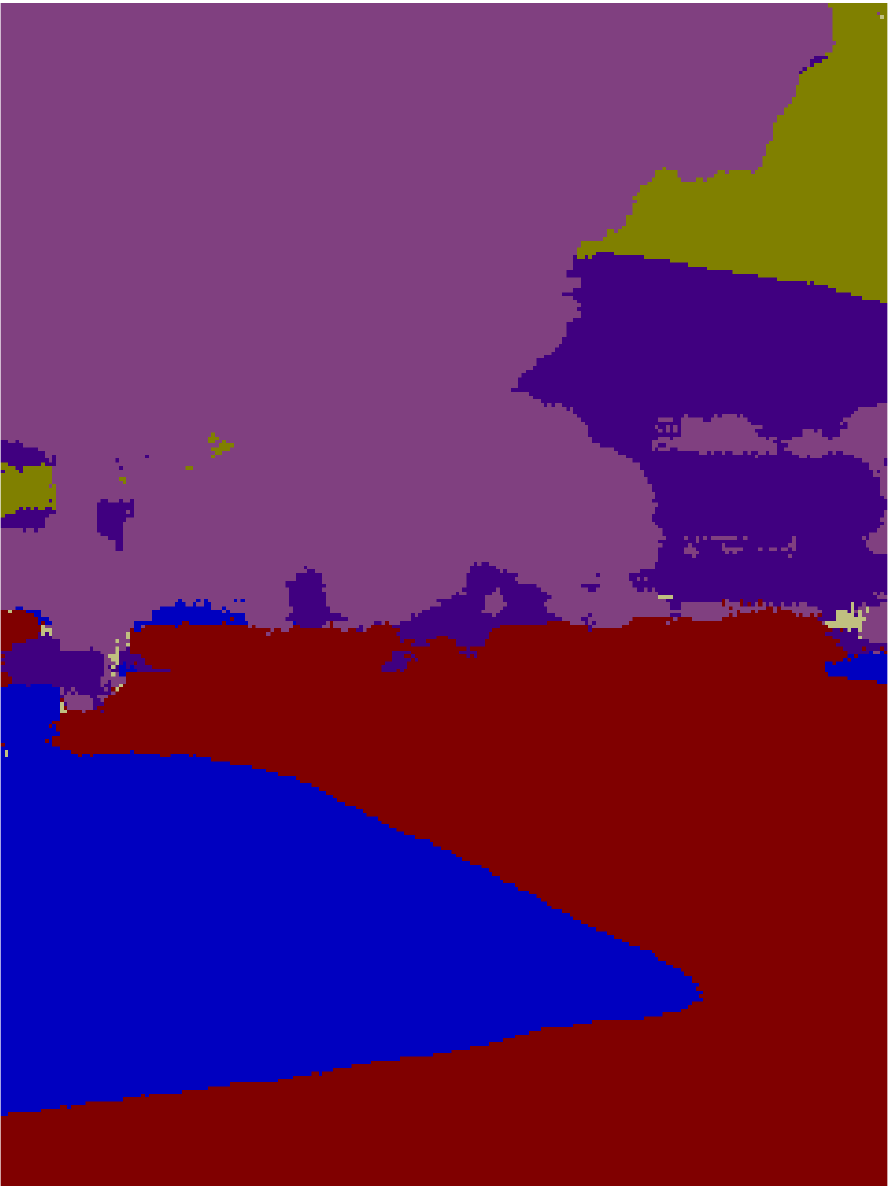} & 
\hspace{-0.0cm}\includegraphics[width=0.118\linewidth]{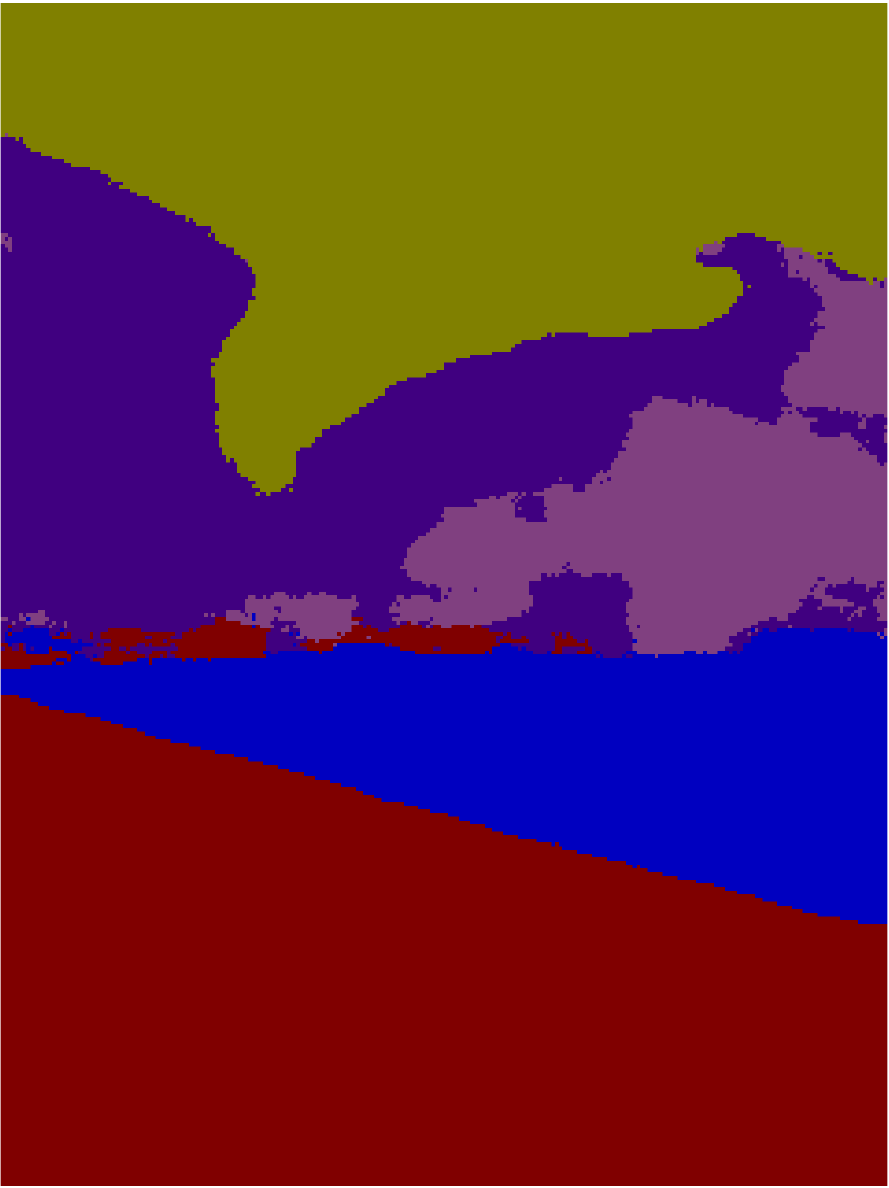} & 
\hspace{-0.0cm}\includegraphics[width=0.118\linewidth]{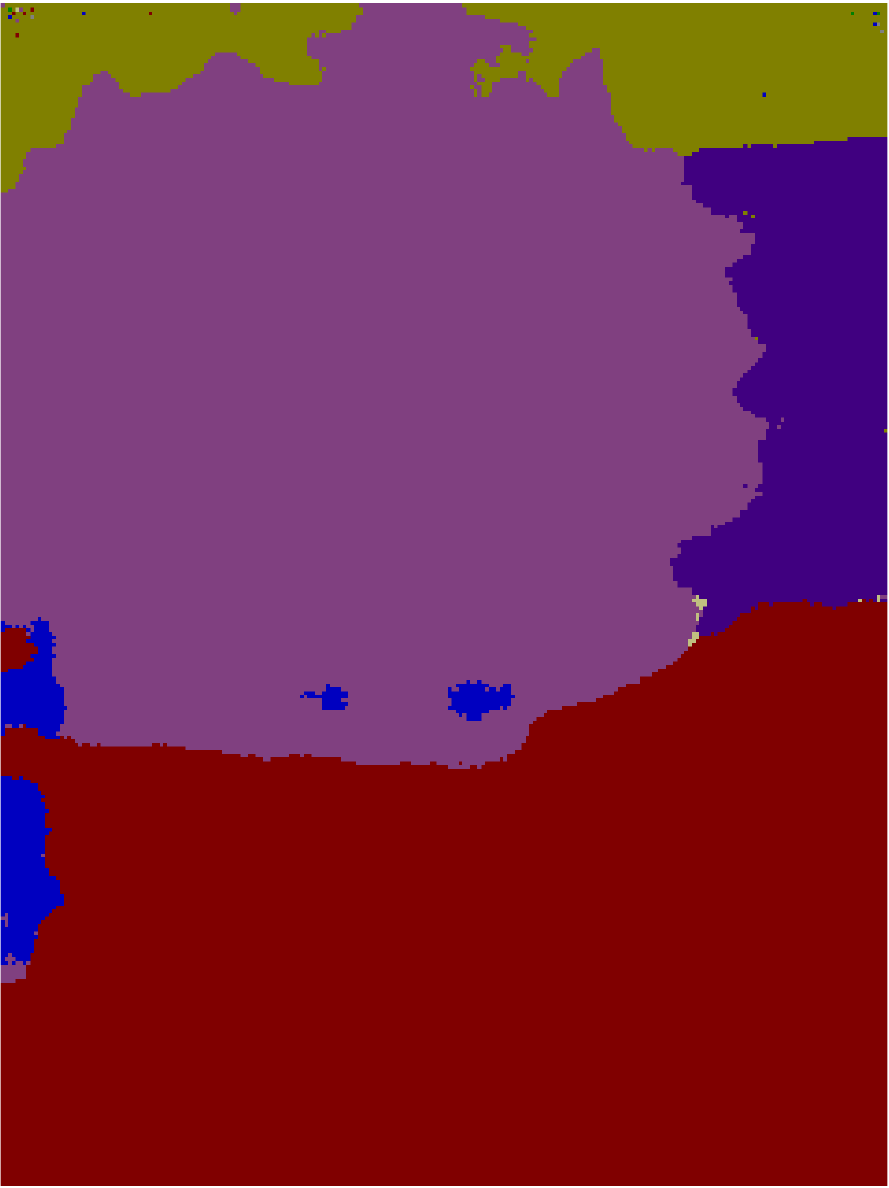}\\
\multicolumn{9}{c}{\hspace{-0.0cm}\includegraphics[width=0.8\linewidth]{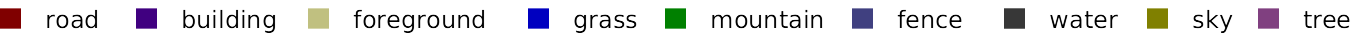}}
%\hspace{-0.3cm}\includegraphics[width=0.110\linewidth]{./figures/make3d/m3dDep_op1-p-343t000-crop.pdf} &
%\hspace{-0.3cm}\includegraphics[width=0.110\linewidth]{./figures/make3d/m3dUnary_op1-p-343t000-crop.pdf} &
%\hspace{-0.3cm}\includegraphics[width=0.110\linewidth]{./figures/make3d/m3destSem_op1-p-343t000-crop.pdf} &
%\hspace{-0.3cm}\includegraphics[width=0.110\linewidth]{./figures/make3d/m3destDeph_op1-p-343t000-crop.pdf} 
\end{tabular}
\end{small}
\vspace{0.05cm}
\caption{{\bf Qualitative results on the Make3D dataset.}  {\bf From top to bottom:} RGB image, ground-truth semantics, observed sparse depth map, estimated high-resolution depth map, noisy semantics obtained from~\cite{Gould2012JMLR} and our improved semantic labels. Best viewed in color.
}
\label{fig:make3dres}
\vspace{-0.3cm}
\end{figure*}

\subsection{Make3d Dataset}
\vspace{-0.07cm}
We further evaluated our model on the Make3d dataset, another challenging outdoor dataset with sparse depth measurements. The ground-truth semantic labels for this dataset were provided by Liu et al.~\cite{Liu10}. We made use of the training and test splits provided with their data, \ie, 400 training images and 134 test images. We estimated depth maps that match the size of the semantic images, \ie, $320\times 240$. Due to the lack of information about camera calibration, we obtained the observation mask by approximately mapping the low-resolution depth maps to the image grid, and excluding the pixels whose depth was greater than $78$. This yields about $15\%$ of observations on average. We  then generated the noisy semantics using~\cite{Gould2012JMLR}, which gave average per pixel and per class accuracies of 87.89\% and 73.75\%.

As before, we extracted square patches of size $5\times 5$ on the training images, depth maps and semantic maps. This dataset contains $L=8$ classes. With our $1.2$ redundancy factor, this yields ${\bf \Omega}_I \in \mathbb{R}^{240\times 25}$, ${\bf \Omega}_D \in \mathbb{R}^{240\times 25}$, and ${\bf \Omega}_S \in \mathbb{R}^{240\times 200}$. We validated the parameters of our approach and obtained $\nu_S = 300$, $\nu_I = \nu_D = 3$. We used the same optimization strategy as before, consisting of iteratively restarting the conjugate gradient descent, and started with $\eta = 300$ decreased to a final value of $\eta = 50$. Our method improves the semantic labeling accuracy to 88.67\%~and~73.81\%. Since no ground-truth depth maps are available for this dataset, we can only perform a qualitative evaluation of our high-resolution depth maps. Some examples of these depth maps are provided in Fig.~\ref{fig:make3dres}. Note that they look realistic and respect the object boundaries.
\vspace{-0.05cm}

\section{Conclusion}
\vspace{-0.2cm}
In this paper, we have presented a novel approach to depth super-resolution in the challenging outdoor setting, where intensity images have large variations due to illumination changes and shadows, and where high-resolution depth maps are difficult to acquire for training. In particular, we have proposed to incorporate semantic information into the super-resolution process, and have shown how to exploit low-resolution training depth maps. Our empirical evaluation on two outdoor datasets has demonstrated the effectiveness of our approach at predicting accurate high-resolution depth maps and semantic labelings. Furthermore, by outperforming state-of-the-art techniques, we have evidenced the benefits of exploiting semantics for depth super-resolution. In our current implementation, partially exploiting a GPU, reconstructing a high-resolution depth map takes 5 minutes for 500 iterations. In the future, we plan to speed this up by making better use of the GPU power. Furthermore, we intend to develop more effective methods to learn analysis operators.
\label{sec:conclusion}

%===========================================================
\bibliographystyle{splncs}
\bibliography{references}

%this would normally be the end of your paper, but you may also have an appendix
%within the given limit of number of pages
\end{document}